\documentclass[a4paper]{article}

\usepackage{bm}
\usepackage{amsmath}
\usepackage{amsthm}
\usepackage{amssymb}
\usepackage{hyperref}
\usepackage{cleveref}
\usepackage{xparse}
\usepackage{dsfont}
\usepackage{tocbibind} 
\usepackage{enumitem}
\usepackage{caption}

\hypersetup{bookmarksnumbered}
\numberwithin{equation}{section}

\usepackage[lmargin=3cm,rmargin=3cm,tmargin=3cm,bmargin=3.5cm]{geometry}

\usepackage{lastpage}
\usepackage{fancyhdr}
\usepackage[ddmmyyyy,hhmmss]{datetime}
\newdateformat{daymonthyeardate}{%
  \THEDAY.~\monthname[\THEMONTH] \THEYEAR}
\fancypagestyle{plain}{
\fancyhf{}
\rfoot{}
\cfoot{\footnotesize Page \thepage{} of \pageref{LastPage}}
}
\newtheorem{theorem}{Theorem}[section]

\newtheorem{lemma}[theorem]{Lemma}
\newtheorem{proposition}[theorem]{Proposition}

\theoremstyle{remark}
\newtheorem{remark}[theorem]{Remark}
\newtheorem*{rem*}{Remark}

\usepackage{todonotes}

\newcommand{\Lip}{\operatorname{Lip}}
\newcommand{\LebesgueMeasure}{\boldsymbol{\lambda}}
\newcommand{\FirstN}[1]{\underline{#1}}

\renewcommand{\emptyset}{\varnothing}

\newcommand{\LayerFunc}{\boldsymbol{\ell}}
\newcommand{\CoeffFunc}{\boldsymbol{c}}
\newcommand{\Architecture}{\boldsymbol{a}}
\NewDocumentCommand\PreNorm{O{p}}{\Gamma_{\alpha,#1}}
\NewDocumentCommand\NNSigma{O{n}D<>{\LayerFunc}O{\CoeffFunc}}{\Sigma_{#1}^{#2, #3}}
\NewDocumentCommand\ApproxSpace{O{p}D<>{\CoeffFunc}O{\alpha}}{A^{#3,#1}_{\LayerFunc,#2}}

\NewDocumentCommand\UnitBall{O{p}D<>{\CoeffFunc}O{\alpha}}{U^{#3,#1}_{\LayerFunc,#2}}
\NewDocumentCommand\ClosedUnitBall{O{p}D<>{\CoeffFunc}O{\alpha}}{\overline{U}^{#3,#1}_{\LayerFunc,#2}}

\newcommand{\WeightSize}[1]{\| #1 \|_{\mathcal{NN}}}

\newcommand{\CompressedDots}{\makebox[1em][c]{.\hfil.\hfil.}}

\newcommand{\N}{\mathbb{N}}

\newcommand{\R}{\mathbb{R}}

\newcommand{\EE}{\mathbb{E}}
\newcommand{\PP}{\mathbb{P}}
\newcommand{\indicator}{\mathds{1}}
\newcommand{\Indicator}{\mathds{1}}

\newcommand{\CalP}{\mathcal{P}}

\newcommand{\SpecialSet}{\CalP_k (\FirstN{2m})}

\newcommand{\deterministic}{\mathrm{det}}
\newcommand{\MonteCarlo}{\mathrm{MC}}

\newcommand{\ClosedBall}{\overline{B}}

\DeclareMathOperator*{\argmin}{argmin}

\newcommand{\eps}{\varepsilon}

\newcommand{\boldnu}{\boldsymbol{\nu}}
\newcommand{\supp}{\operatorname{supp}}
\newcommand{\Alg}{\operatorname{Alg}}

\newcommand{\Covering}{\mathrm{Cov}}

\makeatletter
\newcommand{\oset}[3][0ex]{%
  \mathrel{\mathop{#3}\limits^{
    \vbox to#1{\kern-2\ex@
    \hbox{$\scriptstyle#2$}\vss}}}}
\makeatother

\newcommand{\avsum}{\mathop{\mathpalette\avsuminner\relax}\displaylimits}

\makeatletter
\newcommand\avsuminner[2]{%
  {\sbox0{$\m@th#1\sum$}%
   \vphantom{\usebox0}%
   \ooalign{%
     \hidewidth
     \smash{\vrule height\dimexpr\ht0+1pt\relax depth\dimexpr\dp0+1pt\relax}%
     \hidewidth\cr
     $\m@th#1\sum$\cr
   }%
  }%
}
\makeatother

\newcommand{\A}{\boldsymbol{A}}
\newcommand{\x}{\boldsymbol{x}}
\newcommand{\y}{\boldsymbol{y}}
\newcommand{\z}{\boldsymbol{z}}
\newcommand{\m}{\boldsymbol{m}}

\newcommand{\din}{d_{\mathrm{in}}}
\newcommand{\dout}{d_{\mathrm{out}}}
\newcommand{\VC}{\operatorname{VC}}

\newcommand{\CalE}{\mathcal{E}}
\newcommand{\CalF}{\mathcal{F}}
\newcommand{\CalG}{\mathcal{G}}
\newcommand{\CalH}{\mathcal{H}}
\newcommand{\CalO}{\mathcal{O}}

\newcommand{\CalNN}{\mathcal{NN}}

\newcommand{\nrow}[1]{%
  \relax
  \vcenter to 0pt{%
    \vss
    \kern-1.5ex
    \rlap{$\left.\vphantom{\begin{matrix}0\\\vdots\\0\end{matrix}}\kern3em\right\}#1$}%
    \vss
  }%
}

\newcommand\bovermat[2]{%
 \makebox[0pt][l]{$\smash{\overbrace{\phantom{%
        \begin{matrix}#2\end{matrix}}}^{\text{#1}}}$}#2}

\makeatletter
\newcommand{\subjclass}[2][1991]{%
  \let\@oldtitle\@title%
  \gdef\@title{\@oldtitle\footnotetext{#1 \emph{Mathematics subject classification.} #2}}%
}
\newcommand{\keywords}[1]{%
  \let\@@oldtitle\@title%
  \gdef\@title{\@@oldtitle\footnotetext{\emph{Key words and phrases.} #1.}}%
}
\makeatother

\title{Proof of the Theory-to-Practice Gap in Deep Learning\\
via Sampling Complexity bounds\\
for Neural Network Approximation Spaces}
\date{}

\author{Philipp Grohs%
\thanks{Faculty of Mathematics,
University of Vienna,
Oskar-Morgenstern-Platz~1,
A-1090 Vienna, Austria}
\thanks{Research Platform Data Science @ Uni Vienna,
Währinger Straße 29/S6,
A-1090 Vienna, Austria}
\thanks{Johann Radon Institute,
Altenberger Straße 69,
A-4040 Linz,
Austria}
\thanks{Both authors contributed equally to this work.}
\, and Felix Voigtlaender\footnotemark[1]\,\,\footnotemark[4]}

\begin{document}
\maketitle

\begin{abstract}
We study the computational complexity of (deterministic or randomized) algorithms
based on point samples for approximating or integrating functions
that can be well approximated by neural networks.
Such algorithms (most prominently stochastic gradient descent and its variants)
are used extensively in the field of deep learning.
One of the most important problems in this field concerns the question of whether it is possible
to realize theoretically provable neural network approximation rates by such algorithms.
We answer this question in the negative by proving hardness results
for the problems of approximation and integration on a novel class
of neural network approximation spaces.
In particular, our results confirm a conjectured
and empirically observed theory-to-practice gap in deep learning.
We complement our hardness results by showing that approximation rates
of a comparable order of convergence are (at least theoretically) achievable.
\end{abstract}


\section{Introduction}
\label{sec:Introduction}


The use of data driven classification and regression algorithms based on deep neural networks---
coined \emph{deep learning}---has made a big impact in the areas of artificial intelligence,
machine learning, and data analysis and has led to a number of breakthroughs
in diverse areas of artificial intelligence, including image classification
\cite{lecun1998gradient, krizhevsky2017imagenet, szegedy2015going, he2016deep},
natural language processing \cite{young2018recent},
game playing \mbox{\cite{silver2016mastering,silver2017mastering,vinyals2019grandmaster,mnih2013playing}},
and symbolic mathematics \cite{saxton2018analysing,lample2019deep}. 

More recently, these methods have been applied to problems from the natural sciences
where data driven approaches are combined with physical models.
Example applications in this field---called \emph{scientific machine learning}---%
include the development of drugs \cite{ma2015deep}, molecular dynamics \cite{faber2017prediction},
high-energy physics \cite{baldi2014searching}, protein folding \cite{senior2020improved},
or numerically solving inverse problems and partial differential equations (PDEs)
\cite{arridge2019solving,raissi2019physics,weinan2018deep,hermann2020deep,pfau2020ab}.  

For this wide variety of different application areas,
one can summarize the underlying computational problem as approximating a function $f$
(or a quantity of interest depending on $f$) based on possibly noisy
and random samples $(f(x_i))_{i=1}^m$.
In deep learning this is being done by fitting a neural network 
to these samples using stochastic optimization algorithms.
While there is still no convincingly comprehensive explanation
for the empirically observed success (or failure) of this methodology,
its success critically hinges on the properties
\begin{enumerate}[label=\Alph*.]
	\item \label{itm:approx}
        that $f$  can be well approximated by neural networks, and 
	\item \label{itm:algo}
        that $f$ (or a quantity of interest depending on $f$) can be efficiently
        and accurately reconstructed from a relatively small number of samples $(f(x_i))_{i=1}^m$.
\end{enumerate}
In other words, the validity of both \ref{itm:approx} and \ref{itm:algo}
constitutes a \emph{necessary} condition for a deep learning approach to be efficient.
This is especially true in applications related to scientific machine learning
where often a guaranteed high accuracy is required and where obtaining samples
is computationally expensive.

To date most theoretical contributions focused on property \ref{itm:approx},
namely studying which functions can be well approximated by neural networks.
It is now well understood that neural networks are superior approximators
compared to virtually all classical approximation methods, including polynomials,
finite elements, wavelets, or low rank representations;
see \cite{devore2020neural,grohs2019deep} for two recent surveys.
Beyond that it was recently shown that neural networks can approximate solutions
of high dimensional PDEs without suffering from the curse of dimensionality
\cite{grohs2018proof,hutzenthaler2020proof,kutyniok2019theoretical}.
In light of these results it becomes clear that neural networks are a highly expressive
and versatile function class whose theoretical approximation capabilities vastly outperform
classical numerical function representations.

On the other hand, the question of whether property \ref{itm:algo} holds,
namely to which extent these superior approximation properties can be harnessed
by an efficient algorithm based on point samples,
remains one of the most relevant open questions in the field of deep learning.
At present, almost no theoretical results exist in this direction.
On the empirical side, Adcock and Dexter \cite{adcock2020gap} recently performed a careful study
finding that the theoretical approximation rates are in general not attained by common algorithms,
meaning that the convergence rate of these algorithms does not match
the theoretically postulated approximation rates.
In \cite{adcock2020gap} this empirically observed phenomenon is coined
the \emph{theory-to-practice gap} of deep learning.
In this paper we prove the existence of this gap.

\subsection{Description of Results}

To provide an appropriate mathematical framework for understanding properties \ref{itm:approx}
and \ref{itm:algo} we introduce neural network spaces $\ApproxSpace([0,1]^d)$
which classify functions $f:[0,1]^d \to \R$ according to how rapidly the error
of approximation by neural networks with $n$ weights decays as $n\to \infty$.
Specifically we consider neural networks using the rectified linear unit (ReLU)
activation function, i.e., functions of the form  
$T_L \circ (\varrho \circ T_{L-1}) \circ \cdots \circ (\varrho \circ T_1)$,
where $T_\ell \, x = A_\ell \, x + b_\ell $ are affine mappings
and $\varrho \bigl( (x_1,\dots,x_n)\bigr) = \bigl(\max\{x_1,0\},\dots,\max\{x_n,0\}\bigr)$.
The spaces $\ApproxSpace([0,1]^d)$ also take into account
various common constraints on the magnitude
of the individual network weights (encoded by the function $\CoeffFunc:\N\to \N\cup\{\infty\}$)
and the maximal depth of the network
(encoded by the function $\LayerFunc : \N \to \N_{\geq 2} \cup \{\infty\}$).

Roughly speaking, a function $f : [0,1]^d \to \R$ is an element of the unit ball
\[
  \UnitBall[p]([0,1]^d)
  := \big\{
       f \in \ApproxSpace[p]([0,1]^d)
       \,\,\colon\,\,
       \| f \|_{\ApproxSpace[p]([0,1]^d)}
       \leq 1
     \big\}
\]
of $\ApproxSpace([0,1]^d)$ if for every $n\in \N$ there exists a neural network
with at most $n$ nonzero weights of magnitude at most $\CoeffFunc(n)$
and $L \leq \LayerFunc(n)$ many layers approximating $f$
to within accuracy $\le n^{-\alpha}$ in the $L^p([0,1]^d)$ norm;
see \Cref{sub:NeuralNetworkApproxSpaces} for more details.
Membership of $f$ in such a space for large $\alpha$ simply means
that Property~\ref{itm:approx} is satisfied. 

For the mathematical formalization of Property~\ref{itm:algo}
we employ the formalism of \emph{Information Based Complexity},
as for example presented in \cite{HeinrichRandomApproximation}.
This theory provides a general framework for studying the complexity
of approximating a given solution mapping $S : U \to Y$,
with $U \subset C([0,1]^d)$ bounded,
and $Y$ a Banach space, under the constraint that the approximating algorithm
is only allowed to access \emph{point samples} of the functions $f \in U$.
Formally, a (deterministic) algorithm using $m$ point samples is determined
by a set of sample points $\x = (x_1,\dots,x_m) \in ([0,1]^d)^m$  and a map
$Q : \R^m \to Y$ such that
\[
  A (f) = Q\bigl(f(x_1),\dots,f(x_m)\bigr)
  \qquad \forall \, f \in U .
\]
The set of all such algorithms is denoted $\Alg_m (U,Y)$ and we define the optimal order
for (deterministically) approximating $S : U \to Y$ using point samples
as the best possible convergence rate with respect to the number of samples:
\[
  \beta^{\deterministic}_{\ast} (U,S)
  := \sup
     \Big\{
       \beta \geq 0
       \,\,\colon\,\,
       \exists \, C > 0 \,\,
         \forall \, m \in \N: \quad
           \inf_{A \in \Alg_m (U,Y)} \sup_{f \in U} \| A(f) - S(f) \|_Y \leq C \cdot m^{-\beta}
     \Big\} .
\]
In a similar way one can define randomized (Monte Carlo) algorithms
and consider the optimal order $\beta_\ast^{\MonteCarlo} (U, S)$
for approximating $S$ using randomized algorithms based on point samples;
see Section~\ref{sub:MonteCarloAlgorithms} below.
We emphasize that \emph{all currently used deep learning algorithms,
such as stochastic gradient descent (SGD) \cite{shalev2014understanding}
and its variants (such as ADAM \cite{kingma2014adam}) are of this form.}

In this paper we derive bounds for the optimal orders $\beta_\ast^{\deterministic} (U, S)$
and $\beta_\ast^{\MonteCarlo} (U, S)$ for the unit ball
$U=\UnitBall[\infty]([0,1]^d) \subset \ApproxSpace[\infty]([0,1]^d)$
and the following solution mappings:
\begin{enumerate}
  \item the embedding into $C([0,1]^d)$, i.e., $S = \iota_{\infty}$
        for $\iota_\infty : U \to C([0,1]^d), f \mapsto f$,
  \item the embedding into $L^2([0,1]^d)$, i.e., $S = \iota_2$ for
        $\iota_2 : U \to L^2([0,1]^d), f \mapsto f$, and
  \item the definite integral, i.e., $S = T_{\int}$ for
        $T_{\int} : U \to \R, f \mapsto \int_{[0,1]^d} f(x) \, d x$.
\end{enumerate}

\subsubsection{Approximation with respect to the uniform norm}

We first consider the solution mapping $S = \iota_{\infty}$
operating on $U=\UnitBall[\infty]([0,1]^d)$, i.e., the problem of approximation
with respect to the uniform norm.
Then the property $\beta_\ast^{\MonteCarlo} (U, \iota_\infty)=\alpha$
would imply that the theoretical approximation rate $\alpha$
with respect to the uniform norm can in principle be realized by a (randomized) algorithm
such as SGD and its variants.
On the other hand, if $\beta_\ast^{\MonteCarlo} (U, \iota_\infty)<\alpha$,
then there cannot exist any (randomized) algorithm based on point samples
that realizes the theoretical approximation rate $\alpha$
with respect to the uniform norm---that is, there exists a theory-to-practice gap. 
We now present (a slightly simplified version of) our first main result
establishing such a gap for $\iota_\infty$.
\begin{theorem}[special case of Theorems~\ref{thm:ErrorBoundUniformApproximation} and \ref{thm:UniformApproximationHardness}]
  \label{thm:ErrorBoundUniformApproximationSketch}
  Let $\CoeffFunc : \N \to \N \cup \{ \infty \}$
  be of the form $\CoeffFunc(n) \asymp n^\theta \cdot (\ln (2 n))^{\kappa}$
  for certain $\theta \geq 0$ and $\kappa \in \R$, and let 
  $\LayerFunc^\ast := \sup_{n\in\N} \LayerFunc(n) \in \N \cup \{ \infty \}$.
  Then, if $\LayerFunc^\ast\geq 3$ we have
  \[
    \beta_\ast^{\MonteCarlo}  \bigl(\UnitBall[\infty]([0,1]^d), \iota_\infty\bigr)
    = \beta_\ast^{\deterministic} \bigl(\UnitBall[\infty]([0,1]^d), \iota_\infty\bigr)
    = \begin{cases}
        \frac{1}{d}
        \cdot \frac{\alpha}{\theta\cdot \LayerFunc^\ast +\lfloor\LayerFunc^\ast/2\rfloor  + \alpha}
        & \text{if } \LayerFunc^\ast< \infty \\
        0
        & \text{else}
      \end{cases}
    \in \bigl[0,\tfrac{1}{d} \bigr].
  \]
\end{theorem}

Probably the term ``gap'' is a vast understatement for the difference
between the theoretical approximation rate $\alpha$
and the rate $\beta_\ast\le \min\{ \frac1d,\frac \alpha d\}$
that can actually be realized by a numerical algorithm.
A particular consequence of Theorem~\ref{thm:ErrorBoundUniformApproximationSketch} is that
if all one knows is that a function $f$ is well approximated by neural networks---%
no matter how rapid the approximation error decays---any conceivable numerical algorithm
based on function samples (such as SGD and its variants)
requires at least $\Theta(\varepsilon^{-d})$ many samples
to guarantee an error $\varepsilon>0$ with respect to the uniform norm.
Since evaluating $f$ takes a certain minimum amount of time,
\emph{any conceivable numerical algorithm based on function samples (such as SGD and its variants)
must have a worst-case runtime scaling at least as $\Theta(\varepsilon^{-d})$
to guarantee an error $\varepsilon>0$ with respect to the uniform norm---%
irrespective of how well $f$ can be theoretically approximated by neural networks.}
In particular:
\begin{itemize}
  \item Any conceivable numerical algorithm based on function samples
        (such as SGD and its variants) suffers from the curse of dimensionality%
        ---even if neural network approximations exist that do not.

  \item On the class of all functions well approximable by neural networks
        it is impossible to realize these high convergence rates
        for uniform approximation with any conceivable numerical algorithm
        based on function samples (such as SGD and its variants).

  \item If the number of layers is unbounded (i.e., $\LayerFunc^\ast =\infty$)
        it is impossible to realize \emph{any} positive convergence rate
        on the class of all functions well approximable by neural networks
        for the problem of uniform approximation with any conceivable numerical algorithm
        based on function samples (such as SGD and its variants).
\end{itemize}
Our findings disqualify deep learning based methods for problems
where high uniform accuracy is desired, at least if the only available information
is that the function of interest is well approximated by neural networks.

\subsubsection{Approximation with respect to the \texorpdfstring{$L^2$}{L²} norm}

Next we consider the solution mapping $S = \iota_{2}$ operating on $U=\UnitBall[\infty]([0,1]^d)$,
i.e., the problem of approximation with respect to the $L^2$ norm.
Also in this case we establish a considerable theory-to-practice gap,
albeit not as severe as in the case of $S = \iota_{\infty}$.
A slightly simplified version of our main result is as follows.

\begin{theorem}[special case of Theorems~\ref{thm:L2ErrorBound} and \ref{thm:L2HardnessResult}]
  \label{thm:L2ErrorBoundSketch}
  Let $\CoeffFunc : \N \to \N \cup \{ \infty \}$ be of the form
  $\CoeffFunc(n) \asymp n^\theta \cdot (\ln (2 n))^{\kappa}$
  for certain $\theta \geq 0$ and $\kappa \in \R$,
  assume $\LayerFunc(n)\lesssim  ( \ln (2 n))^\nu $ for certain  $\nu\in [0,\infty)$,
  and let $\LayerFunc^\ast := \sup_{n\in\N} \LayerFunc(n) \in \N \cup \{ \infty \}$.
  Then we have
  \begin{align*}
    \beta_\ast^{\MonteCarlo}  \bigl(\UnitBall[\infty]([0,1]^d), \iota_2\bigr) ,
    \beta_\ast^{\deterministic} \bigl(\UnitBall[\infty]([0,1]^d), \iota_2\bigr)
    &\in \begin{cases}
           \left[
             \frac{1}{2+2/\alpha},
             \frac{1}{2}
             + \frac{\alpha}{\theta\cdot \LayerFunc^\ast +\lfloor\LayerFunc^\ast/2\rfloor + \alpha}
           \right]
           & \text{if } \LayerFunc^\ast< \infty \\[0.2cm]
           \left[ \frac{1}{2+2/\alpha},\min\{\frac{1}{2},\alpha\}\right]
           & \text{else}.
         \end{cases}
  \end{align*}
  In particular, if $\LayerFunc^\ast = \infty$ it holds that 
  \begin{align*}
  	\lim_{\alpha \to \infty}
      \beta_\ast^{\MonteCarlo}  \bigl(\UnitBall[\infty]([0,1]^d), \iota_2\bigr)
    = \lim_{\alpha \to \infty}
        \beta_\ast^{\deterministic} \bigl(\UnitBall[\infty]([0,1]^d), \iota_2\bigr)
     = \frac{1}{2}.
  \end{align*}
\end{theorem}

We see again that it is impossible to realize a high convergence rate
with any conceivable algorithm based on point samples,
no matter how high the theoretically possible approximation rate $\alpha$ may be.
Indeed, the theorem easily implies
\(
  \beta_\ast^{\MonteCarlo}
  \bigl(
    \UnitBall[\infty]([0,1]^d), \iota_2
  \bigr),
  \beta_\ast^{\deterministic}
  \bigl(
    \UnitBall[\infty]([0,1]^d), \iota_2
  \bigr)
  \leq \frac{3}{2},
\)
irrespective of $\alpha$. 
This means that \emph{any conceivable (possibly randomized) numerical algorithm
based on function samples (such as SGD and its variants) must have a worst-case runtime
scaling at least as $\Theta(\varepsilon^{-2/3})$ to guarantee an $L^2$ error $\varepsilon>0$---%
irrespective of how well the function of interest can be theoretically approximated by neural networks.}
On the positive side, there is a uniform lower bound of $\frac{1}{2+\frac{2}{\alpha}}$
for the optimal rate, which means that there exist algorithms (in the sense defined above)
that almost realize an error bound of $\mathcal{O}(m^{-1/2})$, given $m$ point samples,
for $\alpha$ sufficiently large.
Note however that the existence of such an algorithm by no means implies
the existence of an \emph{efficient} algorithm, say,
with runtime scaling linearly or even polynomially in $m$. 

Our findings disqualify deep learning based methods for problems
where a high convergence rate of the $L^2$ error is desired,
at least if the only available information is that the function of interest
is well approximated by neural networks.
On the other hand, deep learning based methods may be a viable option for problems where a low%
---but dimension independent---convergence rate of the $L^2$ error is sufficient.  

\subsubsection{Integration}

Finally we consider the solution mapping $S = T_{\int}$
operating on $U=\UnitBall[\infty]([0,1]^d)$.
The question of estimating $\beta_\ast^{\MonteCarlo}\bigl(\UnitBall[\infty]([0,1]^d,T_{\int}\bigr)$
and $\beta_\ast^{\deterministic} \bigl(\UnitBall[\infty]([0,1]^d), T_{\int}\bigr)$
can be equivalently stated as the question of determining the optimal order
of (Monte Carlo or deterministic) quadrature on neural network approximation spaces.
Again we exhibit a significant theory-to-practice gap
that we summarize in the following simplified version of our main result.

\begin{theorem}[special case of Theorems~\ref{thm:QuadratureDeterministicHardness}, \ref{thm:QuadratureMonteCarloHardness}, \ref{thm:DeterministicIntegrationErrorBound} and \ref{thm:MonteCarloIntegrationErrorBound}]
\label{thm:IntegrationErrorBoundSketch}
  Let $\CoeffFunc : \N \to \N \cup \{ \infty \}$ be of the form
  $\CoeffFunc(n) \asymp n^\theta \cdot (\ln (2 n))^{\kappa}$
  for certain $\theta \geq 0$ and $\kappa \in \R$,
  assume $\LayerFunc(n)\lesssim  ( \ln (2 n))^\nu $ for certain  $\nu\in [0,\infty)$,
  and let $\LayerFunc^\ast := \sup_{n\in\N} \LayerFunc(n) \in \N \cup \{ \infty \}$.
  Then we have
  \begin{align*}
    \beta_\ast^{\deterministic}  \bigl(\UnitBall[\infty]([0,1]^d), T_{\int}\bigr)
    & \in \begin{cases}
             \left[
               \frac{1}{2+1/\alpha},
               1 + \frac{\alpha}{\theta\cdot \LayerFunc^\ast +\lfloor\LayerFunc^\ast/2\rfloor + \alpha  }
             \right]
             & \text{if } \LayerFunc^\ast< \infty \\[0.2cm]
            \left[ \frac{1}{2+1/\alpha}, \min\{1,\alpha\}\right]
             & \text{else},
          \end{cases} \\
     \beta_\ast^{\MonteCarlo}  \bigl(\UnitBall[\infty]([0,1]^d), T_{\int}\bigr)
    & \in \begin{cases}
            \left[
              \frac{1}{2} + \frac{1}{2+2/\alpha},
              1 + \frac{\alpha}{\theta\cdot \LayerFunc^\ast +\lfloor\LayerFunc^\ast/2\rfloor + \alpha}
            \right]
            & \text{if } \LayerFunc^\ast< \infty \\[0.2cm]
            \left[ \frac{1}{2} + \frac{1}{2+2/\alpha},\min\{1,\frac{1}{2} + \alpha\}\right]
            & \text{else}.
    \end{cases}
  \end{align*}
  In particular, if $\LayerFunc^\ast = \infty$ it holds that 
  \[
  	\lim_{\alpha \to \infty}
      \beta_\ast^{\MonteCarlo}  \bigl(\UnitBall[\infty]([0,1]^d), T_{\int}\bigr)
     = 1.
  \]
\end{theorem}

We see in particular that \emph{there are no (deterministic or Monte Carlo) quadrature schemes
achieving a convergence order greater than 2.
Further, if the number of layers is unbounded,
there are no (deterministic or Monte Carlo) quadrature schemes
achieving a convergence order greater than 1.}
On the other hand there exist Monte Carlo algorithms that almost realize a rate $1$
for $\alpha$ sufficiently large.
This again does not imply the existence of an \emph{efficient} algorithm
with this convergence rate;
but it is well-known that the error bound $\CalO(m^{-1/2})$ can be \emph{efficiently} realized
by standard Monte Carlo integration,
since each $f\in \UnitBall[\infty]$ satisfies $\|f\|_{L^\infty}\leq 1$.
Theorem~\ref{thm:IntegrationErrorBoundSketch} implies that there is not much room for improvement.

\subsubsection{General Comments}

We close with the following general comments.

\begin{itemize}
  \item Our results for the first time shed light on the question
        of which problem classes can be efficiently tackled by deep learning methods
        and which problem classes might be better handled
        using classical methods such as finite elements.
        These findings enable informed choices regarding the use of these methods.
        Concretely, we find that it is not advisable to use deep learning methods
        for problems where a high convergence rate and/or uniform accuracy is needed.
        In particular, \emph{no high order (approximation or quadrature) algorithms exist},
        provided that the only available information is that the function of interest
        is well approximated by neural networks.

  \item As another contribution, we exhibit the exact impact of the choice of the architecture,
        i.e., the number of layers, encoded by $\LayerFunc$, and magnitude of the coefficients,
        encoded by $\CoeffFunc$.
        For example, if $\CoeffFunc(n) \asymp n^\theta \cdot (\ln (2 n))^{\kappa}$
        and $\LayerFunc^\ast := \sup_{n\in\N} \LayerFunc(n) \in \N \cup \{ \infty \}$,
        the effect of the architecture on the algorithmic performance is encoded by the critical quantity 
        $\gamma:=\theta\cdot \LayerFunc^\ast +\lfloor\LayerFunc^\ast/2\rfloor$.
        Particularly, we show that allowing the number of layers to be unbounded
        adversely affects the optimal rate $\beta_\ast$.

  \item Our hardness results hold universally across virtually all choices of network architectures.
        Concretely, all hardness results of Theorems~\ref{thm:ErrorBoundUniformApproximationSketch},
        \ref{thm:L2ErrorBoundSketch} and \ref{thm:IntegrationErrorBoundSketch}
        hold true whenever at least $\LayerFunc^\ast \geq 3$ layers are used.
        This means that \emph{limiting the number of layers will not help.}
        In this context we also note that it is known that at least
        $\LayerFunc^\ast \geq \lfloor \alpha/2d \rfloor$ layers are needed
        for ReLU neural networks to achieve the (essentially) optimal approximation rate
        $\frac{\alpha}{d}$ for all $f\in C^{\alpha}([0,1]^d)$;
        see \cite[Theorem~C.6]{petersen2018optimal}.

  \item Our hardness results hold universally across all size constraints
        on the magnitudes of the approximating network weights, as encoded by $\CoeffFunc$.
        Furthermore, a careful analysis of our proofs reveals that our hardness results
        qualitatively remain true if analogous constraints are put on the $\ell^2$
        norms of the weights of the approximating networks.
        Such constraints constitute a common regularization strategy,
        termed \emph{weight decay} \cite{gupta1998weight}.
        This means that \emph{applying standard regularization strategies---such as weight decay---%
        will not help.} 
\end{itemize}

\subsection{Related work}%
\label{sub:RelatedWork}

To put our results in perspective we discuss related work.

\subsubsection{Information Based Complexity and Classical Function Spaces}

The study of optimal rates $\beta_\ast$ for approximating a given solution map
based on point samples or general linear samples has a long tradition in approximation theory,
function space theory, spectral theory and information based complexity.
It is closely  related to so-called \emph{Gelfand numbers} of linear operators%
---a classical and well studied concept in function space theory
and spectral theory \cite{pietsch1986eigenvalues,pinkus2012n}.
It is instructive to compare our findings to these classical results,
for example for $U$ the unit ball in a Sobolev spaces $W_\infty^\alpha([0,1]^d)$ and $S=\iota_\infty$.
These Sobolev spaces can be (not quite but almost,
see for example \cite[Theorem~5.3.2]{timan2014theory} and \cite[Theorem~12.1.1]{ditzian2012moduli})
characterized by the property that its elements can be approximated by polynomials
of degree $\leq n$ to within $L^\infty$ accuracy $\mathcal{O}(n^{-\alpha})$.
Since the set of polynomials of degree $\leq n$ in dimension $d$
possesses $\asymp n^d$ degrees of freedom,
this approximation rate can be fully harnessed by a deterministic,
resp.~Monte Carlo algorithm based on point samples
if $\beta_\ast^{\deterministic}  \bigl(U, S\bigr) = \alpha/d$,
resp.~$\beta_\ast^{\MonteCarlo}  \bigl(U, S\bigr) = \alpha/d$.
It is a classical result that this is indeed the case, see
\cite[Theorem~6.1]{HeinrichRandomApproximation}.
This fact implies that there is no theory-to-practice gap
in polynomial approximation and can be considered the basis of any high order
(approximation or quadrature) algorithm in numerical analysis.

In the case of classical function spaces it is the generic behavior
that the optimal rate $\beta_\ast$ increases (linearly)
with the underlying smoothness $\alpha$, at least for fixed dimension $d$.
On the other hand, our results show that neural network approximation spaces
have the peculiar property that the optimal rate $\beta_\ast$ is always uniformly bounded,
regardless of the underlying ``smoothness'' $\alpha$. 

To put our results in a somewhat more abstract context
we can compare the optimal rate $\beta_\ast$ to
other complexity measures of a function space.
A well studied example is the metric entropy related to the covering numbers
$\Covering(V,\varepsilon)$ of sets $V \subset C[0,1]^d$.
The associated entropy exponent is
\[
  s_{\ast} (U)
  := \sup
     \big\{
       \lambda \geq 0
       \,\,\colon\,\,
       \exists \, C > 0 \,\,
         \forall \,\varepsilon \in (0,1): \quad
           \Covering (U,\varepsilon) \leq \exp\bigl( C \cdot \varepsilon^{-1/\lambda} \bigr)
     \big\} ,
\]
which, roughly speaking, determines the theoretically optimal rate $\mathcal{O}(m^{-s_\ast})$
at which an arbitrary element of $U$ can be approximated from a representation using at most $m$ bits.
On the other hand, $\beta_\ast$ determines the optimal rate $\mathcal{O}(m^{-\beta_\ast})$
that can actually be realized by an algorithm using $m$ point samples
of the input function $f\in U$.
For a solution mapping $S$ to be efficiently computable from point samples,
one would therefore expect that $\beta_\ast = s_\ast$
or at least that $\beta_\ast$ grows linearly with $s_\ast$.
For example, for $U$ the unit ball in a Sobolev spaces $W_\infty^\alpha([0,1]^d)$
and $S=\iota_\infty$ we have
\({
  s_{\ast}(U)
  = \beta_\ast^{\deterministic} ( U, \iota_\infty )
  = \beta_\ast^{\MonteCarlo} ( U, \iota_\infty )
  =\frac{\alpha}{d}
  .
}\)
In contrast, $\UnitBall[\infty] = \UnitBall[\infty]([0,1]^d)$ satisfies
$s_{\ast}\bigl(\UnitBall[\infty]\bigr) \geq \alpha$
according to Lemma~\ref{lem:ApproxSpaceCoveringBounds},
while Theorem~\ref{thm:ErrorBoundUniformApproximationSketch} shows
\(
  \beta_{\ast}^{\deterministic} \bigl( \UnitBall[\infty], \iota_\infty \bigr),
  \beta_{\ast}^{\MonteCarlo} \bigl( \UnitBall[\infty], \iota_\infty \bigr)
  \leq \frac{1}{d}
\)
independent of $\alpha$, and even
\(
  \beta_{\ast}^{\deterministic} \bigl( \UnitBall[\infty], \iota_\infty \bigr)
  =\beta_{\ast}^{\MonteCarlo} \bigl( \UnitBall[\infty], \iota_\infty \bigr)
  = 0
\)
if the number of layers is unbounded.
This is yet another manifestation of the wide theory-to-practice gap in neural network approximation.

\subsubsection{Other Hardness Results for Deep Learning}

While we are not aware of any work addressing the optimal sampling complexity
on neural network spaces, there exist a number of different approaches
to establishing various ``hardness'' results for deep learning.
We comment on some of them. 

A prominent and classical research direction considers the computational complexity
of fitting a neural network of a fixed architecture to given (training) samples.
It is known that this can be an NP complete problem for certain specific architectures
and samples; see \cite{blum1989training} for the first result in this direction
that has inspired a large body of follow-up work.
This line of work does however not consider the full scope of the problem,
namely the relation between theoretically possible approximation rates
and algorithmically realizable rates.
In our results we do not take into account the computational efficiency
of algorithms at all.
Our results are stronger in the sense that they show that
\emph{even if there was an efficient algorithm for fitting a neural network to samples,
one would need to access too many samples to achieve efficient runtimes.}

Another research direction considers the existence of convergent algorithms
that only have access to inexact information about the samples,
as is commonly the case when computing in floating point arithmetic.
Specifically, \cite{antun2021can} identifies various problems in sparse approximation
that cannot be algorithmically solved based on inputs with finite precision
using neural networks.
The deeper underlying reason is that these problems cannot be solved
by \emph{any} algorithm based on inexact measurements.
Thus, the results of \cite{antun2021can} are not really specific to neural networks.
In contrast, \emph{our hardness results are highly specific to the structure
of neural networks and do not occur for most other computational approaches.} 

A different kind of hardness results appears in the neural network approximation theory literature.
There, typically lower bounds are provided for the number of network weights and/or number of layers
that a neural network needs to have in order to reach a desired accuracy
in the approximation of functions from various classical smoothness spaces
\cite{yarotsky2018optimal,petersen2018optimal,boelcskeiNeural,telgarsky2016benefits}.
Yet, these bounds exclusively concern theoretical approximation rates
for classical smoothness spaces while
\emph{our results provide bounds for the algorithmic realizability of these rates.}

\subsubsection{Other Work on Neural Network Approximation Spaces}

Our definition of neural network approximation spaces is inspired
by \cite{NNApproximationSpaces} where such spaces were first introduced
and some structural properties, such as embedding theorems into classical function spaces,
are investigated.
The neural network spaces $\ApproxSpace([0,1]^d)$ introduced in the present work
differ from those spaces in the sense that we also allow to take the size
of the network weights into account.
This is important, as such bounds on the weights
are often enforced in applications through regularization procedures.
Another construction of neural network approximation spaces can be found in
\cite{JentzenNNApproximationSpaces} for the purpose of providing a calculus
on functions that can be approximated by neural networks without curse of dimensionality.
While all these works focus on aspects related to theoretical approximability of functions,
our main focus concerns the algorithmic realization of such approximations.

\subsection{Notation}%
\label{sub:Notation}

For $n \in \N$, we write $\FirstN{n} := \{ 1,2,\dots,n \}$.
For any finite set $I \neq \emptyset$ and any sequence $(a_i)_{i \in I} \subset \R$,
we define $\avsum_{i \in I} a_i := \frac{1}{|I|} \sum_{i \in I} a_i$.
The expectation of a random variable $X$ will be denoted by $\EE[X]$.

For a subset $M \subset X$ of a metric space $X$,
we write $\overline{M}$ for the closure of $M$ and $M^\circ$ for the interior of $M$.
In particular, this notation applies to subsets of $\R^d$.

\subsection{Structure of the paper}%
\label{sub:Structure}

\Cref{sec:ApproximationSpacesAndSamplingComplexity} formally introduces the
neural network approximation spaces $\ApproxSpace$ and furthermore provides
a review of the most important notions and definitions from information based complexity.
The basis for all our hardness results is developed in \Cref{sec:HatFunctionConstruction},
where we show that the unit ball $\UnitBall[\infty]([0,1]^d)$
in the approximation space $\ApproxSpace[\infty]([0,1]^d)$ contains a large family
of ``hat functions'', depending on the precise properties of the functions $\LayerFunc,\CoeffFunc$
and on $\alpha > 0$.

The remaining sections develop error bounds and hardness results for the problems of
uniform approximation (\Cref{sec:UniformApproximationErrorBounds,sec:UniformApproximationHardness}),
approximation in $L^2$ (\Cref{sec:L2ApproximationErrorBounds,sec:L2ApproximationHardness}),
and numerical integration (\Cref{sec:IntegrationErrorBounds,sec:IntegrationHardness}).
Several technical proofs and results are deferred to \Cref{sec:TechnicalResults}.

\section{The notion of sampling complexity on neural network approximation spaces}%
\label{sec:ApproximationSpacesAndSamplingComplexity}


In this section, we first formally introduce the neural network approximation spaces $\ApproxSpace$
and then review the framework of information based complexity,
including the notion of randomized (Monte Carlo) algorithms
and the concept of the optimal order of convergence based on point samples.

\subsection{The mathematical formalization of neural networks}%
\label{sub:NeuralNetworks}

In our analysis, it will be helpful to distinguish between a neural network $\Phi$
as a set of weights and the associated function $R_\varrho \Phi$ computed by the network.
Thus, we say that a \emph{neural network} is a tuple
${\Phi = \big( (A_1,b_1), \dots, (A_L,b_L) \big)}$,
with $A_\ell \in \R^{N_\ell \times N_{\ell-1}}$ and $b_\ell \in \R^{N_\ell}$.
We then say that ${\Architecture(\Phi) := (N_0,\dots,N_L) \in \N^{L+1}}$
is the \emph{architecture} of $\Phi$, $L(\Phi) := L$ is the \emph{number of layers}%
\footnote{Note that the number of \emph{hidden} layers is given by $H = L-1$.}
of $\Phi$, and ${W(\Phi) := \sum_{j=1}^L (\| A_j \|_{\ell^0} + \| b_j \|_{\ell^0})}$
denotes the \emph{number of (non-zero) weights} of $\Phi$.
The notation $\| A \|_{\ell^0}$ used here denotes the number of non-zero entries
of a matrix (or vector) $A$.
Finally, we write $\din(\Phi) := N_0$ and $\dout(\Phi) := N_L$
for the \emph{input and output dimension} of $\Phi$, and we set
$\WeightSize{\Phi} := \max_{j = 1,\dots,L} \max \{ \| A_j \|_{\infty}, \| b_j \|_{\infty} \}$,
where ${\| A \|_{\infty} := \max_{i,j} |A_{i,j}|}$.

To define the function $R_\varrho \Phi$ computed by $\Phi$, we need to specify an
\emph{activation function}.
In this paper, we will only consider the so-called \emph{rectified linear unit (ReLU)}
${\varrho : \R \to \R, x \mapsto \max \{ 0, x \}}$,
which we understand to act componentwise on $\R^n$, i.e.,
$\varrho \bigl( (x_1,\dots,x_n)\bigr) = \bigl(\varrho(x_1),\dots,\varrho(x_n)\bigr)$.
The function $R_\varrho \Phi : \R^{N_0} \to \R^{N_L}$ computed by the network $\Phi$
(its \emph{realization}) is then given by
\[
  R_\varrho \Phi
  := T_L \circ (\varrho \circ T_{L-1}) \circ \cdots \circ (\varrho \circ T_1)
  \quad \text{where} \quad
  T_\ell \, x = A_\ell \, x + b_\ell .
\]

\subsection{Neural network approximation spaces}%
\label{sub:NeuralNetworkApproxSpaces}

\emph{Approximation spaces} \cite{DeVoreConstructiveApproximation} classify functions
according to how well they can be approximated by a family
$\boldsymbol{\Sigma} = (\Sigma_n)_{n \in \N}$ of certain ``simple functions''
of increasing complexity $n$, as $n \to \infty$.
Common examples consider the case where $\Sigma_n$ is the set of polynomials of degree $n$,
or the set of all linear combinations of $n$ wavelets.
The notion of \emph{neural network approximation spaces} was originally introduced in
\cite{NNApproximationSpaces}, where $\Sigma_n$ was taken to be a family of neural networks
of increasing complexity.
However, \emph{\cite{NNApproximationSpaces} does not impose any restrictions on the size
of the individual network weights}, which plays an important role in practice and%
---as we shall see---also influences the possible performance of algorithms based on point samples.

For this reason, we introduce a modified notion of neural network approximation spaces
that also takes the size of the individual network weights into account.
Precisely, given an input dimension $d \in \N$ (which we will keep fixed throughout this paper) and
non-decreasing functions ${\LayerFunc : \N \to \N_{\geq 2} \cup \{ \infty \}}$
and $\CoeffFunc : \N \to \N \cup \{ \infty \}$ (called the \textbf{depth-growth function}
and the \textbf{coefficient growth function}, respectively), we define
\[
  \NNSigma
  := \Big\{
       R_\varrho \Phi
       \,\, \colon
       \begin{array}{l}
         \Phi \text{ NN with }
         d_{\mathrm{in}}(\Phi) = d,
         d_{\mathrm{out}}(\Phi) = 1, \\
         W(\Phi) \leq n,
         L(\Phi) \leq \LayerFunc(n),
         \WeightSize{\Phi} \leq \CoeffFunc(n)
       \end{array}
     \Big\} .
\]
Then, given a measurable subset $\Omega \subset \R^d$, $p \in [1,\infty]$,
and $\alpha \in (0,\infty)$, for each measurable $f : \Omega \to \R$, we define
\[
  \PreNorm (f)
  := \max
     \Big\{
       \| f \|_{L^p(\Omega)} , \quad
       \sup_{n \in \N}
       \big[
         n^{\alpha}
         \cdot d_{p} \bigl(f, \NNSigma\bigr)
       \big]
     \Big\}
  \in [0,\infty] ,
\]
where $d_{p}(f, \Sigma) := \inf_{g \in \Sigma} \| f - g \|_{L^p(\Omega)}$.

The remaining issue is that since the set $\NNSigma$ is in general neither
closed under addition nor under multiplication with scalars, $\PreNorm$
is \emph{not a (quasi)-norm}.
To resolve this issue,
taking inspiration from the theory of Orlicz spaces
(see e.g.~\cite[Theorem~3 in Section~3.2]{RaoRenOrliczSpaces}),
we define the \emph{neural network approximation space quasi-norm}
$\| \cdot \|_{\ApproxSpace}$ as
\[
  \| f \|_{\ApproxSpace}
  := \inf \bigl\{ \theta > 0 \,\,\colon\,\, \PreNorm(f / \theta) \leq 1 \bigr\}
  \in [0,\infty],
\]
giving rise to the \emph{approximation space}
\[
  \ApproxSpace
  := \ApproxSpace (\Omega)
  := \bigl\{
       f \in L^p(\Omega)
       \,\,\colon\,\,
       \| f \|_{\ApproxSpace} < \infty
     \bigr\}
  .
\]
The following lemma summarizes the main elementary properties of these spaces.

\begin{lemma}\label{lem:ApproximationSpaceProperties}
  Let $\emptyset \neq \Omega \subset \R^d$ be measurable,
  let $p \in [1,\infty]$ and $\alpha \in (0,\infty)$.
  Then, $\ApproxSpace := \ApproxSpace (\Omega)$ satisfies the following properties:
  \begin{enumerate}
    \item $(\ApproxSpace, \| \cdot \|_{\ApproxSpace})$ is a quasi-normed space.
          Precisely, given arbitrary measurable functions $f,g : \Omega \to \R$, it holds that
          $\| f + g \|_{\ApproxSpace} \leq C \cdot (\| f \|_{\ApproxSpace} + \| g \|_{\ApproxSpace})$
          for $C := 17^\alpha$.

    \item We have $\PreNorm (c f) \leq |c| \, \PreNorm(f)$ for $c \in [-1,1]$.

    \item $\PreNorm(f) \leq 1$ if and only if $\| f \|_{\ApproxSpace} \leq 1$.

    \item $\PreNorm(f) < \infty$ if and only if $\| f \|_{\ApproxSpace} < \infty$.

    \item $\ApproxSpace (\Omega) \hookrightarrow L^p(\Omega)$.
          Furthermore, if $\Omega \subset \overline{\Omega^\circ}$,
          then $\ApproxSpace[\infty](\Omega) \hookrightarrow C_b(\Omega)$,
          where $C_b (\Omega)$ denotes the Banach space of continuous functions
          that are bounded and extend continuously to the closure $\overline{\Omega}$
          of $\Omega$.
  \end{enumerate}
\end{lemma}

\begin{proof}
  See \Cref{sub:NNApproximationSpacesProofs}.
\end{proof}

\subsection{Quantities characterizing the complexity of the network architecture}%
\label{sub:GammaDefinition}

To conveniently summarize those aspects of the growth behavior of the functions
$\LayerFunc$ and $\CoeffFunc$ most relevant to us, we introduce three quantities
that will turn out to be crucial for characterizing the sample complexity
of the neural network approximation spaces.
First, we set
\begin{equation}
  \LayerFunc^\ast
  := \sup_{n\in\N} \LayerFunc(n)
  \in \N \cup \{ \infty \} .
  \label{eq:MaximalDepth}
\end{equation}
Furthermore, we define
\begin{equation}
  \begin{split}
    \gamma^{\flat} (\LayerFunc,\CoeffFunc)
    & := \sup
         \Big\{
           \gamma \in [0,\infty)
           \colon
           \exists \, L \in \N_{\leq \LayerFunc^\ast} \text{ and } C > 0
             \quad \forall \, n \in \N:
               n^\gamma \leq C \cdot (\CoeffFunc(n))^L \cdot n^{\lfloor L/2 \rfloor}
         \Big\} , \\
    \gamma^{\sharp} (\LayerFunc, \CoeffFunc)
    & := \inf
         \Big\{
           \gamma \in [0,\infty)
           \colon
           \exists \, C > 0
             \quad \forall \, n \in \N , L \in \N_{\leq \LayerFunc^\ast} :
               (\CoeffFunc(n))^L \cdot n^{\lfloor L/2 \rfloor} \leq C \cdot n^\gamma
         \Big\} .   
  \end{split}
  \label{eq:GammaDefinition}
\end{equation}

\begin{remark}\label{rem:GammaRemark}
  Clearly, $\gamma^{\flat}(\LayerFunc,\CoeffFunc) \leq \gamma^{\sharp}(\LayerFunc,\CoeffFunc)$.
  Furthermore, since we will only consider settings in which $\LayerFunc^\ast \geq 2$,
  we always have $\gamma^{\sharp}(\LayerFunc,\CoeffFunc) \geq \gamma^{\flat}(\LayerFunc,\CoeffFunc) \geq 1$.
  Next, note that if $\LayerFunc^\ast = \infty$ (i.e., if $\LayerFunc$ is unbounded), then
  $\gamma^{\flat}(\LayerFunc,\CoeffFunc) = \gamma^{\sharp}(\LayerFunc,\CoeffFunc) = \infty$.
  Finally, we remark that if $\LayerFunc^\ast < \infty$ and if $\CoeffFunc$ satisfies
  the natural growth condition $\CoeffFunc(n) \asymp n^\theta \cdot (\ln (2 n))^{\kappa}$
  for certain $\theta \geq 0$ and $\kappa \in \R$, then
  \(
    \gamma^{\flat}(\LayerFunc,\CoeffFunc)
    = \gamma^{\sharp}(\LayerFunc,\CoeffFunc)
    = \theta \cdot \LayerFunc^\ast + \lfloor \LayerFunc^\ast / 2 \rfloor .
  \)
  Thus, in most natural cases---but not always---$\gamma^{\flat}$ and $\gamma^{\sharp}$ agree.

  An explicit example where $\gamma^{\flat}$ is not identical to $\gamma^{\sharp}$ is as follows:
  Define $c_1 := c_2 := c_3 := 1$ and for $n,m \in \N$ with $2^{2^m} \leq n < 2^{2^{m+1}}$, define
  $c_n := 2^{2^m}$.
  Then, assume that $\gamma_1,\gamma_2 \in [0,\infty)$ and $\kappa_1,\kappa_2 > 0$
  satisfy $\kappa_1 \, n^{\gamma_1} \leq c_n \leq \kappa_2 \, n^{\gamma_2}$ for all $n \in \N$.
  Applying the upper estimate for arbitrary $m \in \N$ and $n = n_m = 2^{2^m}$,
  we see $n = c_n \leq \kappa_2 \, n^{\gamma_2}$; since $n_m = 2^{2^m} \to \infty$ as $m \to \infty$,
  this is only possible if $\gamma_2 \geq 1$.
  On the other hand, if we applying the lower estimate for arbitrary $m \in \N$ and
  $n = n_m = 2^{2^{m+1}} - 1$, we see because of
  $c_n = 2^{2^m} = 2^{2^{m+1} / 2} = \sqrt{2^{2^{m+1}}} = \sqrt{n+1}$ that
  \(
    \kappa_1 \, n^{\gamma_1}
    \leq c_n
    =    \sqrt{n+1} .
  \)
  Again, since $n_m = 2^{2^{m+1}} - 1$ to $\infty$ as $m \to \infty$,
  this is only possible if $\gamma_1 \leq \frac{1}{2}$.

  Given these considerations, it is easy to see for $\ell \equiv L \in \N_{\geq 2}$ that
  $\gamma^{\flat}(\LayerFunc,\CoeffFunc) \leq \frac{L}{2} + \lfloor \frac{L}{2} \rfloor$,
  while $\gamma^{\sharp}(\LayerFunc,\CoeffFunc) \geq L + \lfloor \frac{L}{2} \rfloor$.
  In particular, $\gamma^{\flat}(\LayerFunc,\CoeffFunc) < \gamma^{\sharp}(\LayerFunc,\CoeffFunc)$.
  \hfill$\triangle$
\end{remark}

\subsection{The framework of sampling complexity}%


Let $d \in \N$, let $\emptyset \neq U \subset C([0,1]^d)$ be bounded,
and let $Y$ be a Banach space.
We are interested in numerically approximating a given \textbf{solution mapping}
$S : U \to Y$, where the numerical procedure is only allowed to access \emph{point samples}
of the functions $f \in X_0$.
The procedure can be either deterministic or probabilistic (Monte Carlo).
In this short section, we discuss the mathematical formalization of this problem,
based on the setup of \emph{numerical complexity theory}, as for instance
outlined in \cite[Section~2]{HeinrichRandomApproximation}.

The reader should keep in mind that we are mostly interested in the setting where $U$
is the unit ball in the neural network approximation space $\ApproxSpace[\infty]([0,1]^d)$, i.e.,
\begin{equation}
  U
  = \UnitBall[\infty]([0,1]^d)
  := \big\{
       f \in \ApproxSpace[\infty]([0,1]^d)
       \,\,\colon\,\,
       \| f \|_{\ApproxSpace[\infty]}
       \leq 1
     \big\}
  ,
  \label{eq:UnitBallDefinition}
\end{equation}
and where the solution mapping is one of the following:
\begin{enumerate}
  \item the embedding into $C([0,1]^d)$, i.e., $S = \iota_{\infty}$
        for $\iota_\infty : U \to C([0,1]^d), f \mapsto f$,
  \item the embedding into $L^2([0,1]^d)$, i.e., $S = \iota_2$ for
        $\iota_2 : U \to L^2([0,1]^d), f \mapsto f$, or
  \item the definite integral, i.e., $S = T_{\int}$ for
        $T_{\int} : U \to \R, f \mapsto \int_{[0,1]^d} f(x) \, d x$.
\end{enumerate}

\subsubsection{The deterministic setting}%
\label{sub:DeterministicAlgorithms}

A (potentially non-linear) map $A : U \to Y$ is called
a \textbf{deterministic method of order $m \in \N$} (written $A \in \Alg_m (U,Y)$)
if there exists $\x = (x_1,\dots,x_m) \in ([0,1]^d)^m$ and a map $Q : \R^m \to Y$ such that
\[
  A (f) = Q\bigl(f(x_1),\dots,f(x_m)\bigr)
  \qquad \forall \, f \in U .
\]
Given a (solution) mapping $S : U \to Y$, we define the error of $A$ in approximating $S$ as
\[
  e(A,U,S)
  := \sup_{f \in U} \| A(f) - S(f) \|_Y .
\]
The \textbf{optimal error for (deterministically) approximating $S : U \to Y$ using $m$ point samples}
is then
\[
  e^{\deterministic}_m (U,S)
  := \inf_{A \in \Alg_m (U,Y)} e(A,S) .
\]
Finally, the \textbf{optimal order for (deterministically) approximating $S : U \to Y$ using point samples}
is
\begin{equation}
  \beta^{\deterministic}_{\ast} (U,S)
  := \sup
     \big\{
       \beta \geq 0
       \,\,\colon\,\,
       \exists \, C > 0 \,\,
         \forall \, m \in \N: \quad
           e^{\deterministic}_m (U,S) \leq C \cdot m^{-\beta}
     \big\} .
  \label{eq:OptimalOrderDeterministic}
\end{equation}

\subsubsection{The Monte Carlo setting}%
\label{sub:MonteCarloAlgorithms}

A \textbf{Monte Carlo method using $m \in \N$ point measurements (in expectation)}
is a tuple $(\A,\m)$ consisting of a family $\A = (A_\omega)_{\omega \in \Omega}$
of (potentially non-linear) map $A_\omega : U \to Y$
indexed by a probability space $(\Omega,\CalF,\PP)$
and a measurable function $\m : \Omega \to \N$ with the following properties:
\begin{enumerate}
  \item \label{enu:MonteCarloMeasurable}
        for each $f \in U$, the map $\Omega \to Y, \omega \mapsto A_\omega(f)$
        is measurable (with respect to the Borel $\sigma$-algebra on $Y$),
  \item for each $\omega \in \Omega$, we have $A_\omega \in \Alg_{\m(\omega)}(U,Y)$,
  \item \label{enu:MonteCarloNumberMeasurements}
        $\EE_{\omega} [\m(\omega)] \leq m$.
\end{enumerate}
We write $(\A, \m) \in \Alg^{\MonteCarlo}_m (U,Y)$ if these conditions are satisfied.
We say that $(\A,\m)$ is \emph{strongly measurable} if the map
$\Omega \times U \to Y, (\omega,f) \mapsto A_\omega(f)$ is measurable,
where $U \subset C([0,1]^d)$ is equipped with the Borel $\sigma$-algebra induced by $C([0,1]^d)$.

\begin{rem*}
  In most of the literature (see e.g.~\cite[Section~2]{HeinrichRandomApproximation}),
  Monte Carlo algorithms are always assumed to be strongly measurable.
  All Monte Carlo algorithms that we construct will have this property.
  On the other hand, all our hardness results apply to arbitrary Monte Carlo algorithms
  satisfying Properties~\ref{enu:MonteCarloMeasurable}--\ref{enu:MonteCarloNumberMeasurements}
  from above.
  Thus, using the terminology just introduced we obtain stronger results than we would get
  using the usual definition.
\end{rem*}

The \textbf{expected error} of a Monte Carlo algorithm $(\A,\m)$ for approximating
a (solution) mapping $S : U \to Y$ is defined as
\[
  e\bigl( (\A,\m), U, S\bigr)
  := \sup_{f \in U}
       \EE_\omega \bigl[\| S(f) - A_\omega (f) \|_Y\bigr] .
\]
The \textbf{optimal Monte Carlo error for approximating $S : U \to Y$ using $m$ point samples
(in expectation)} is
\[
  e_m^{\MonteCarlo} (U, S)
  := \inf_{(\A,\m) \in \Alg^{\MonteCarlo}_m (U, Y)}
       e\bigl( (\A,\m), U, S\bigr) .
\]
Finally, the \textbf{optimal Monte Carlo order for approximating $S : U \to Y$ using point samples}
is
\[
  \beta_\ast^{\MonteCarlo} (U, S)
  := \sup
     \big\{
       \beta \geq 0
       \,\,\colon\,\,
       \exists \, C > 0 \,\,
         \forall \, m \in \N: \quad
           e_m^{\MonteCarlo} (U,S)
           \leq  C \cdot m^{-\beta}
     \big\} .
\]

\medskip{}

The remainder of this paper is concerned with deriving upper and lower bounds for
the exponents $\beta_\ast^{\deterministic}(U,S)$ and $\beta_\ast^{\MonteCarlo}(U,S)$,
where $U = \UnitBall[\infty]$ is the unit ball in $\ApproxSpace[\infty]$,
and $S$ is either the embedding of $\ApproxSpace[\infty]$ into $C([0,1]^d)$,
the embedding into $L^2([0,1]^d)$, or the definite integral $S f = \int_{[0,1]^d} f(t) \, d t$.

For deriving upper bounds (i.e., hardness bounds) for Monte Carlo algorithms,
we will frequently use the following lemma, which is a slight adaptation of
\cite[Proposition~4.1]{HeinrichRandomApproximation}.
In a nutshell, the lemma shows that if one can establish a hardness result
that holds for deterministic algorithms \emph{in the average case},
then this implies a hardness result for Monte Carlo algorithms.

\begin{lemma}\label{lem:MonteCarloHardnessThroughAverageCase}
  Let $\emptyset \neq U \subset C([0,1]^d)$ be bounded, let $Y$ be a Banach space,
  and let $S : U \to Y$.
  Assume that there exist $\lambda \in [0,\infty)$, $\kappa > 0$, and $m_0 \in \N$
  such that for every $m \in \N_{\geq m_0}$ there exists a finite set $\Gamma_m \neq \emptyset$
  and a family of functions $(f_{\gamma})_{\gamma \in \Gamma_m} \subset U$ satisfying
  \begin{equation}
    \avsum_{\gamma \in \Gamma_m}
      \| S(f_\gamma) - A(f_\gamma) \|_Y
    \geq \kappa \cdot m^{-\lambda}
    \qquad \forall \, A \in \Alg_m (U, Y) .
    \label{eq:AverageCaseHardness}
  \end{equation}
  Then $\beta_\ast^{\deterministic}(U,S),\beta_\ast^{\MonteCarlo}(U,S) \leq \lambda$.
\end{lemma}

\begin{proof}
  \textbf{Step~1 (proving $\beta_\ast^{\deterministic}(U,S) \leq \lambda$):}
  For every $A \in \Alg_m(U,Y)$, \Cref{eq:AverageCaseHardness} implies because of $f_{\gamma} \in U$
  that
  \[
    e(A,U,S)
    = \sup_{f \in U}
        \| A(f) - S(f) \|_{Y}
    \geq \avsum_{\gamma \in \Gamma_m}
           \| S(f_\gamma) - A(f_\gamma) \|_{Y}
    \geq \kappa \, m^{-\lambda} .
  \]
  Since this holds for every $m \in \N_{\geq m_0}$ and every $A \in \Alg_m(U,Y)$,
  with $\kappa$ independent of $A,m$, this easily implies
  $e_{m}^{\deterministic}(U,S) \geq \kappa \, m^{-\lambda}$ for all $m \in \N_{\geq m_0}$,
  and then $\beta_{\ast}^{\deterministic}(U,S) \leq \lambda$.

  \medskip{}

  \textbf{Step~2 (proving $\beta_\ast^{\MonteCarlo}(U,S) \leq \lambda$):}
  Let $m \in \N_{\geq m_0}$ and let $(\A,\m) \in \Alg_{m}^{\MonteCarlo}(U,Y)$ be arbitrary,
  with $\A = (A_\omega)_{\omega \in \Omega}$ for a probability space $(\Omega,\CalF,\PP)$.
  Define $\Omega_0 := \{ \omega \in \Omega \colon \m(\omega) \leq 2 m \}$ and note
  $m \geq \EE_\omega [\m(\omega)] \geq 2 m \cdot \PP(\Omega_0^c)$, which shows
  $\PP(\Omega_0^c) \leq \frac{1}{2}$ and hence $\PP(\Omega_0) \geq \frac{1}{2}$.

  Note that $A_\omega \in \Alg_{2m} (U, Y)$ for each $\omega \in \Omega_0$,
  so that \Cref{eq:AverageCaseHardness} (with $2m$ instead of $m$) shows
  \(
    \avsum_{\gamma \in \Gamma_{2m}}
      \big\| f_{\gamma} - A_\omega (f_{\gamma}) \big\|_{Y}
    \geq \kappa \cdot (2 m)^{-\lambda}
    \geq \widetilde{\kappa} \cdot m^{-\lambda}
  \)
  for a constant $\widetilde{\kappa} = \widetilde{\kappa}(\kappa,\lambda) > 0$.
  Therefore,
  \begin{equation}
    \begin{split}
      e \big( (\A,\m), U, Y \big)
      = \sup_{f \in U}
          \EE_\omega \| f - A_\omega (f) \|_{Y}
      & \geq \avsum_{\gamma \in \Gamma_{2m}}
               \EE_\omega \big\| f_{\gamma} - A_\omega (f_{\gamma}) \big\|_{Y} \\
      & \geq \EE_\omega
             \bigg[
               \Indicator_{\Omega_0}(\omega)
               \avsum_{\gamma \in \Gamma_{2m}}
                 \big\| f_{\gamma} - A_\omega (f_{\gamma}) \big\|_{Y}
             \bigg] \\
      & \geq \PP(\Omega_0)
             \cdot \widetilde{\kappa}
             \cdot m^{-\lambda}
        \geq \frac{\widetilde{\kappa}}{2}
             \cdot m^{-\lambda} ,
    \end{split}
    \label{eq:AverageHardnessImpliesMCHardness}
  \end{equation}
  and hence
  \(
    e_m^{\MonteCarlo} \big( U, S \big)
    \geq \frac{\widetilde{\kappa}}{2} \cdot m^{-\lambda} ,
  \)
  since \Cref{eq:AverageHardnessImpliesMCHardness} holds for any Monte Carlo algorithm
  $(\A,\m) \in \Alg_m^{\MonteCarlo} (U, Y)$.
  Finally, since $m \in \N_{\geq m_0}$ can be chosen arbitrarily, we see as claimed that
  \(
    \beta_\ast^{\MonteCarlo}(U, S)
    \leq \lambda
    .
  \)
\end{proof}


\section{Richness of the unit ball in \texorpdfstring{the spaces $\ApproxSpace[\infty]$}{neural network approximation spaces}}%
\label{sec:HatFunctionConstruction}

\begin{figure}[h]
  \begin{center}
    \includegraphics[width=0.35\textwidth]{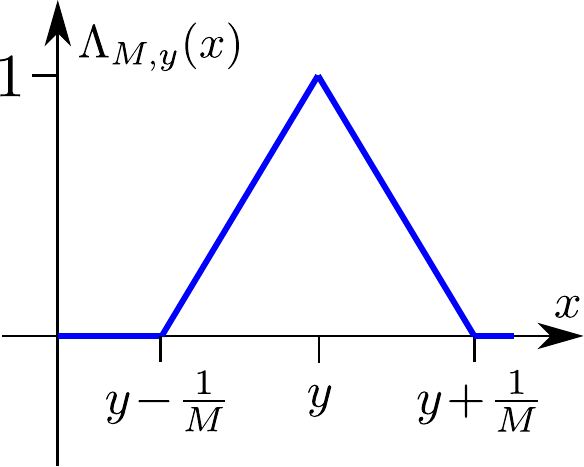}
  \end{center}
  \caption{\label{fig:LambdaPlot}A plot of the ``hat-function'' $\Lambda_{M,y}$
  formally defined in \Cref{eq:LambdaDefinition}.}
\end{figure}

In this section, we show that ReLU networks with a limited number of neurons
and bounded weights can well approximate several different functions of ``hat-function type,''
as shown in \Cref{fig:LambdaPlot}.
The fact that this is possible implies that the unit ball
$\UnitBall[\infty] \subset \ApproxSpace[\infty]$ is quite rich;
this will be the basis of all of our hardness results.

We begin by considering the most basic ``hat function'' ${\Lambda_{M,y} : \R \to [0,1]}$,
defined for $M > 0$ and $y \in \R$ by
\begin{equation}
  \Lambda_{M,y}(x)
  = \begin{cases}
       0,                         & \text{if }                 x \leq y - M^{-1} , \\
        M \cdot (x - y + M^{-1}), & \text{if } y - M^{-1} \leq x \leq y , \\
       -M \cdot (x - y - M^{-1}), & \text{if } y          \leq x \leq y + M^{-1} , \\
       0,                         & \text{if } y + M^{-1} \leq x . \\
     \end{cases}
  \label{eq:LambdaDefinition}
\end{equation}
For later use, we note that $\int_{\R} \Lambda_{M,y}(x) \, d x = M^{-1}$.
Furthermore, we ``lift'' $\Lambda_{M,y}$ to a function on $\R^d$ by setting
$\Lambda_{M,y}^\ast : \R^d \to \R, x = (x_1,\dots,x_d) \mapsto \Lambda_{M,y}(x_1)$.
The following lemma gives a bound on how economically sums of the functions $\Lambda_{M,y}$
can be implemented by ReLU networks.

\begin{lemma}\label{lem:OneDimensionalHatSumImplementation}
  Let $\LayerFunc : \N \to \N_{\geq 2} \cup \{ \infty \}$
  and $\CoeffFunc : \N \to \N \cup \{ \infty \}$ be non-decreasing.
  Let $M \geq 1$, $n \in \N$, and $0 < C \leq \CoeffFunc(n)$,
  as well as $L \in \N_{\geq 2}$ with $L \leq \LayerFunc(n)$.
  Then
  \[
    \frac{C^L \cdot n^{\lfloor L/2 \rfloor}}{4 M n} \sum_{i=1}^n \eps_i \Lambda_{M,y_i}^\ast
    \in \NNSigma[(2 L + 8) n] .
    \qquad \forall \, \eps_1,\dots,\eps_n \in [-1,1] \text{ and } y_1,\dots,y_n \in [0,1].
  \]
\end{lemma}

\begin{proof}
  Let $\eps_1,\dots,\eps_n \in [-1,1]$ and $y_1,\dots,y_n \in [0,1]$.
  Let $e_1 := (1,0,\dots,0) \in \R^{1 \times d}$ and define
  \vspace*{0.1cm}
  \par
  \begin{minipage}[c]{0.25\textwidth}
    \[
      A_1
      := \frac{C}{2}
         \left(
           \begin{matrix}
             e_1 \\ \vdots \\ e_1
           \end{matrix}
         \right)
      \!\in\! \R^{3 n \times d} ,
      \vspace*{0.2cm}
    \]
  \end{minipage}
  \begin{minipage}[c]{0.75\textwidth}
    \begin{alignat*}{3}
      A_2^{(0)}
      & := \frac{C}{2} \!\cdot\!
           \left(
             \begin{array}{ccccccc}
               \!\!
               \eps_1 & -2 \eps_1 & \eps_1 & \cdots & \eps_n & -2 \eps_n & \eps_n
               \!\!
             \end{array}
           \right)
      \!\in \R^{1 \times 3 n} , \\
      A_2
      & := \left(
             \begin{matrix}
               A_2^{(0)} \\
               - A_2^{(0)}
             \end{matrix}
           \right)
      \in \R^{2 \times 3 n} ,
    \end{alignat*}
  \end{minipage}
  as well as
  \[
    b_1
    := \frac{C}{2} \!\cdot\!
       \Big(
         \begin{array}{c|c|c|c|c|c|c}
           \!\!
           -y_1 + M^{-1} \vphantom{\displaystyle\sum}
           & -y_1
           & -y_1 - M^{-1}
           & \cdots
           & -y_n + M^{-1}
           & -y_n
           & -y_n - M^{-1}\!\!
         \end{array}
       \Big)^T
    \!\!\in \R^{3 n} .
  \]
  Finally, set $E := (C \mid -C) \in \R^{1 \times 2}$ and
  \par
  \begin{minipage}[c]{0.3\textwidth}
    \[
      A
      := C \cdot
         \left(
           \begin{array}{cc}
             1 & 0 \\
             \nrow{n} \vdots & \vdots \\
             1 & 0 \\
             \hline
             \\\\[-3.5\medskipamount]
             0 & 1 \\
             \nrow{n} \vdots & \vdots \\
             0 & 1
           \end{array}
         \right)
      \quad \in \R^{2 n \times 2} ,
    \]
  \end{minipage}
  \begin{minipage}[c]{0.7\textwidth}
    \begin{alignat*}{3}
      B
      & := C \cdot
           \left(
             \begin{array}{ccc|ccc}
               \bovermat{$n$}{1 & \cdots & 1}
               & \bovermat{$n$}{\hphantom{-} 0 & \cdots & \hphantom{-} 0} \\
               0  & \cdots & 0 & \hphantom{-} 1 & \cdots & \hphantom{-} 1 \\
             \end{array}
           \right)
      && \in \R^{2 \times 2 n} , \\
      D
      & := C \cdot \;
           \left(
             \begin{array}{ccc|ccc}
               1 & \dots & 1 & -1 & \dots & -1
             \end{array}
           \right)
      && \in \R^{1 \times 2 n} .
    \end{alignat*}
  \end{minipage}
  Note that
  \(
    \| A \|_{\infty},
    \| B \|_{\infty},
    \| D \|_{\infty},
    \| E \|_{\infty},
    \| A_1 \|_{\infty},
    \| A_2 \|_{\infty},
    \| A_2^{(0)} \|_{\infty}
    \leq C .
  \)
  Furthermore, since $y_j \in [0,1]$ and $M \geq 1$, we also see $\| b_1 \|_{\infty} \leq C$.
  Next, note that $\| A_1 \|_{\ell^0}, \| A_2^{(0)} \|_{\ell^0}, \| b_1 \|_{\ell^0} \leq 3n$,
  $\| A_2 \|_{\ell^0} \leq 6 n$, $\| A \|_{\ell^0}, \| B \|_{\ell^0}, \| D \|_{\ell^0} \leq 2n$,
  and $\| E \|_{\ell^0} \leq 2 \leq 2 n$.

  \smallskip{}

  For brevity, set $\gamma := \frac{C^L \, n^{\lfloor L/2 \rfloor}}{4 n M}$
  and $\Xi := \sum_{i=1}^n \eps_i \Lambda_{M,y_i}^\ast$, so that $\Xi : \R^d \to \R$.
  Before we describe how to construct a network $\Phi$ implementing
  $\gamma \cdot \Xi$, we collect a few auxiliary observations.
  First, a direct computation shows that
  \[
    \tfrac{C}{2 M} \Lambda_{M,y} (x)
    = \varrho \big( \tfrac{C}{2}(x - y + \tfrac{1}{M}) \big)
      - 2 \varrho \bigl(\tfrac{C}{2} (x - y)\bigr)
      + \varrho \big( \tfrac{C}{2} (x - y - \tfrac{1}{M}) \big) .
  \]
  Based on this, it is easy to see
  \begin{align}
    A_2^{(0)} \big[ \varrho (A_1 x + b_1) \big]
    & = \frac{C}{2}
        \sum_{j=1}^n
        \bigg[
          \eps_j \cdot
          \Big(
            \varrho \big( \tfrac{C}{2} (x_1 - y_j + \tfrac{1}{M}) \big)
            - 2 \varrho \big( \tfrac{C}{2} (x_1 - y_j) \big)
            + \varrho \big( \tfrac{C}{2} (x - y_j - \tfrac{1}{M}) \big)
          \Big)
        \bigg]
      \nonumber \\
    & = \frac{C}{2} \frac{C}{2 M}
        \sum_{j=1}^n
          \eps_j \, \Lambda_{M,y_j} (x)
      = \frac{C^2}{4 M} \Xi (x) .
  \label{eq:HatImplementation1DMainIdentity}
  \end{align}
  By definition of $A_2$, this shows
  $F(x) = \frac{C^2}{4 M} \bigl(\varrho(\Xi(x)) , \varrho(-\Xi(x))\bigr)^T$
  for all $x \in \R^d$, for the function
  $F := \varrho \circ A_2 \circ \varrho \circ (A_1 \bullet + b_1) : \R^d \to \R$.

  A further direct computation shows for $x,y \in \R$ that
  \begin{equation}
    \begin{split}
      & \Big[ B \varrho\bigl(A (\begin{smallmatrix} x \\ y \end{smallmatrix})\bigr) \Big]_1
        = C \sum_{j=1}^n
               \varrho \Big( \big( A (\begin{smallmatrix} x \\ y \end{smallmatrix}) \big)_j \Big)
        = C \sum_{j=1}^n \varrho (C x)
        = C^2 n \, \varrho(x) \\
      \text{ and similarly } \quad
      & \Big[ B \varrho\bigl(A (\begin{smallmatrix} x \\ y \end{smallmatrix})\bigr) \Big]_2
        = C^2 n \, \varrho(y) .
    \end{split}
    \label{eq:AmplificationTrick}
  \end{equation}
  Thus, setting $G := \varrho \circ B \circ A : \R^2 \to \R^2$,
  we see $G(x,y) = C^2 n \bigl(\varrho(x), \varrho(y)\bigr)^T$.
  Therefore, denoting by $G^j := G \circ \cdots \circ G$ the $j$-fold composition of $G$ with itself,
  we see $G^j (x,y) = (C^2 n)^j \cdot \bigl(\varrho(x), \varrho(y)\bigr)^T$ for $j \in \N$, and hence
  \begin{equation}
    G^j \bigl(F(x)\bigr)
    = \frac{C^{2 j + 2} \, n^j}{4 M}
      \cdot \big( \varrho(\Xi(x)), \quad \varrho(-\Xi(x)) \big)^T
    \qquad \forall \, j \in \N_0 \text{ and } x \in \R^d ,
    \label{eq:GjFExplicit}
  \end{equation}
  where the case $j = 0$ is easy to verify separately.

  In a similar way, we see for $H := D \circ \varrho \circ A : \R^2 \to \R$ that
  \begin{equation}
    H (x,y)
    = D \Big[ \varrho \bigl(A (\begin{smallmatrix} x \\ y \end{smallmatrix}) \bigr) \Big]
    = C \cdot
      \bigg(
        \sum_{j=1}^n
          \varrho (C x)
        - \sum_{j=1}^n
            \varrho (C y)
      \bigg)
    = C^2 n \bigl(\varrho(x) - \varrho(y)\bigr)
    \qquad \forall \, x,y \in \R .
    \label{eq:HExplicit}
  \end{equation}

  \smallskip{}

  Now, we prove the claim of the lemma, distinguishing three cases regarding $L \in \N_{\geq 2}$.
  \smallskip{}

  \textbf{Case~1} ($L = 2$):
  Define $\Phi := \big( (A_1, b_1), (A_2^{(0)}, 0) \big)$.
  Then \Cref{eq:HatImplementation1DMainIdentity} shows $R_\varrho \Phi = \frac{C^2}{4 M} \Xi$.
  Because of $\frac{C^L \, n^{\lfloor L/2 \rfloor}}{4 n M} = \frac{C^2}{4 M}$ for $L = 2$,
  this implies the claim, once we note that
  \[
    L(\Phi)
    = L
    \leq \LayerFunc(n)
    \leq \LayerFunc ( (2L + 8) n)
    \quad \text{ and } \quad
    \WeightSize{\Phi} \leq C \leq \CoeffFunc(n) \leq \CoeffFunc( (2 L + 8) n) ,
  \]
  as well as $W(\Phi) \leq 9 n \leq (2 L + 8) n$, since $L = 2$.

  \medskip{}

  \textbf{Case~2} ($L \geq 4$ is even):
  In this case, define
  \[
    \Phi
    := \Big(
         (A_1, b_1),
         (A_2, 0),
         \underbrace{
           (A,0), (B,0), \dots, (A,0), (B,0)
         }_{\frac{L-4}{2} \text{ copies of } (A,0), (B,0)},
         (A,0),
         (D,0)
       \Big)
  \]
  and note for $j := \frac{L - 4}{2}$ that $j+1 = \frac{L-2}{2} = \lfloor L/2 \rfloor - 1$,
  so that a combination of \Cref{eq:HExplicit,eq:GjFExplicit} shows
  \[
    R_\varrho \Phi (x)
    = (H \circ G^j \circ F)(x)
    = C^2 n
      \cdot \frac{C^{2 j + 2} \, n^j}{4 M}
      \cdot \bigl(\varrho(\Xi(x)) - \varrho(- \Xi(x))\bigr)
    = \frac{C^L \, n^{\lfloor L/2 \rfloor}}{4 M n} \cdot \Xi (x) ,
  \]
  since $\varrho (\varrho(z)) = \varrho(z)$ and $\varrho(z) - \varrho(-z) = z$ for all $z \in \R$.
  Finally, we note as in the previous case that $L(\Phi) = L \leq \LayerFunc( (2 L + 8) n)$
  and $\WeightSize{\Phi} \leq C \leq \CoeffFunc( (2L + 8) n)$, and furthermore that
  \[
    W(\Phi)
    \leq 3 n + 3 n + 6 n + \frac{L-4}{2} \big( 2 n + 2 n \big) + 4 n
    =    16 n + (2 L - 8) n
    =    (8 + 2 L) n
    .
  \]
  Overall, we see also in this case that $\gamma \cdot \Xi \in \NNSigma[(2 L + 8) n]$, as claimed.

  \medskip{}

  \textbf{Case~3} ($L \geq 3$ is odd):
  In this case, define
  \[
    \Phi
    := \Big(
         (A_1, b_1),
         (A_2, 0) ,
         \underbrace{
           (A,0), (B,0), \dots, (A,0),(B,0)
         }_{\frac{L-3}{2} \text{ copies of } (A,0),(B,0)} ,
         (E,0)
       \Big) .
  \]
  Then, setting $j := \frac{L-3}{2}$ and noting $j = \lfloor L/2 \rfloor - 1$,
  we see thanks to \Cref{eq:GjFExplicit} and because of $E = (C \mid -C)$ that
  \[
    R_\varrho \Phi (x)
    = E \Big( G^j \bigl(F(x)\bigr) \Big)
    = C \cdot
      \frac{C^{2 j + 2} \, n^j}{4 M} \cdot
      \big( \varrho(\Xi(x)) - \varrho(- \Xi(x)) \big)
    = \frac{C^L \, n^{\lfloor L/2 \rfloor}}{4 M n} \cdot \Xi(x) .
  \]
  It remains to note as before that $L(\Phi) = L \leq \LayerFunc( (2L + 8) n)$
  and $\WeightSize{\Phi} \leq C \leq \CoeffFunc( (2L + 8) n)$, and finally that
  \(
    W(\Phi)
    \leq 3 n + 3 n + 6 n + \frac{L-3}{2} (2 n + 2 n) + 2
    =    2 + 6n + 2 L n
    \leq (8 + 2 L) n,
  \)
  so that indeed $\gamma \cdot \Xi \in \NNSigma[(8 + 2 L) n]$ also in this case.
\end{proof}

As an application of \Cref{lem:OneDimensionalHatSumImplementation},
we now describe a large class of functions contained in the unit ball of
the approximation space $\ApproxSpace[\infty]([0,1]^d)$.

\begin{lemma}\label{lem:HatSumsInUnitBall}
  Let $\alpha > 0$ and let $\LayerFunc, \CoeffFunc : \N \to \N \cup \{ \infty \}$
  be non-decreasing with $\LayerFunc^\ast \geq 2$.
  Let $\sigma \geq 2$, $0 < \gamma < \gamma^{\flat}(\LayerFunc,\CoeffFunc)$,
  $\theta \in (0,\infty)$ and $\lambda \in [0,1]$ with $\theta \lambda \leq 1$
  be arbitrary and define
  \[
    \omega
    := \min
       \big\{
         -\theta \alpha, \quad
         \theta \cdot (\gamma - \lambda) - 1
       \big\}
    \in (-\infty,0) .
  \]
  Then there exists a constant
  $\kappa = \kappa(\alpha,\theta,\lambda,\gamma,\sigma,\LayerFunc,\CoeffFunc) > 0$
  such that for every $m \in \N$, the following holds:

  Setting $M := 4 m$ and $z_j := \frac{1}{4 m} + \frac{j-1}{2 m}$ for $j \in \FirstN{2m}$,
  the functions $\bigl(\Lambda_{M,z_j}^\ast\bigr)_{j \in \FirstN{2m}}$
  are supported in $[0,1]^d$ and have disjoint supports, up to a null-set.
  Furthermore, for any $\boldnu = (\nu_j)_{j \in \FirstN{2m}} \in [-1,1]^{2m}$
  and $J \subset \FirstN{2 m}$ satisfying $|J| \leq \sigma \cdot m^{\theta \lambda}$, we have
  \[
    f_{\boldnu,J}
    := \kappa \cdot m^\omega \cdot \sum_{j \in J} \nu_j \, \Lambda_{M,z_j}^\ast
    \in \ApproxSpace[\infty]([0,1]^d)
    \qquad \text{and} \qquad
    \big\| f_{\boldnu,J} \big\|_{\ApproxSpace[\infty]([0,1]^d)} \leq 1 .
  \]
\end{lemma}

\begin{proof}
  Since $\gamma < \gamma^{\flat}(\LayerFunc,\CoeffFunc)$, we see by definition of $\gamma^{\flat}$
  that there exist $L = L(\gamma,\LayerFunc,\CoeffFunc) \in \N_{\leq \LayerFunc^\ast}$
  and $C_1 = C_1(\gamma,\LayerFunc,\CoeffFunc) > 0$ such that
  $n^\gamma \leq C_1 \cdot (\CoeffFunc(n))^L \cdot n^{\lfloor L/2 \rfloor}$ for all $n \in \N$.
  Because of $\LayerFunc^\ast$, we can assume without loss of generality that $L \geq 2$.
  Furthermore, since $L \leq \LayerFunc^\ast$, we can choose
  $n_0 = n_0(\gamma,\LayerFunc,\CoeffFunc) \in \N$ satisfying $L \leq \LayerFunc(n_0)$.

  Let $m \in \N$ and let $\boldnu$ and $J$ be as in the statement of the lemma.
  For brevity, define ${f_{\boldnu,J}^{(0)} := \sum_{j \in J} \nu_j \Lambda_{M,z_j}^\ast}$.
  We note that $\Lambda_{M,z_j}^\ast$ is continuous with $0 \leq \Lambda_{M,z_j}^\ast \leq 1$ and
  \[
    \supp \Lambda_{M,z_j}^\ast
    \subset \bigl\{ x \in \R^d \colon x_1 \in y_j + [\tfrac{1}{M}, \tfrac{1}{M}] \bigr\}
    \subset \bigl\{ x \in \R^d \colon x_1 \in \tfrac{j-1}{2 m} + [0, \tfrac{1}{2 m}] \bigr\} .
  \]
  This shows that the supports of the functions $\Lambda_{M,z_j}^\ast$ are
  contained in $[0,1]^d$ and are pairwise disjoint (up to null-sets),
  which then implies $\big\| f_{\boldnu,J}^{(0)} \big\|_{L^\infty} \leq 1$.

  Next, since $\theta \lambda \leq 1$, we have
  $\lceil m^{\theta \lambda} \rceil \leq \lceil m \rceil = m \leq 2 m$.
  Thus, by possibly enlarging the set $J \subset \FirstN{2m}$ and setting $\nu_j := 0$
  for the added elements, we can without loss of generality assume that
  $|J| \geq \lceil m^{\theta \lambda} \rceil \geq 1$.
  Note that the extended set still satisfies $|J| \leq \sigma \cdot m^{\theta \lambda}$
  since $\lceil m^{\theta \lambda} \rceil \leq 2 m^{\theta \lambda}$ and $\sigma \geq 2$.

  Now, define $N := n_0 \cdot \big\lceil m^{(1-\lambda) \theta} \big\rceil$ and
  $n := N \cdot |J|$, noting that $n \geq n_0$.
  Furthermore, writing ${J = \{ i_1,\dots,i_{|J|} \}}$, define
  \[
    (\eps_1,\CompressedDots,\eps_n)
    := \Bigl(
         \underbrace{
           \nu_{i_1}, \CompressedDots, \nu_{i_1}
         }_{N \text{ times}},
         \CompressedDots,
         \underbrace{
           \nu_{i_{|J|}}, \CompressedDots, \nu_{i_{|J|}}
         }_{N \text{ times}}
       \Bigr)
    \quad \text{and} \quad
    (y_1,\CompressedDots,y_n)
    := \Big(
         \underbrace{
           z_{i_1}, \CompressedDots, z_{i_1}
         }_{N \text{ times}},
         \CompressedDots,
         \underbrace{
           z_{i_{|J|}}, \CompressedDots, z_{i_{|J|}}
         }_{N \text{ times}}
       \Big)
    .
  \]
  By choice of $C_1$, we have $n^\gamma \leq C_1 \cdot (\CoeffFunc(n))^L \cdot n^{\lfloor L/2 \rfloor}$,
  so that we can choose $0 < C \leq \CoeffFunc(n)$ satisfying
  $n^\gamma \leq C_1 \cdot C^L \cdot n^{\lfloor L/2 \rfloor}$.
  Since we also have $L \geq 2$ and $L \leq \LayerFunc(n_0) \leq \LayerFunc(n)$,
  \Cref{lem:OneDimensionalHatSumImplementation} shows that
  \[
    \NNSigma[(2L+8) n]
    \ni \frac{C^L \, n^{\lfloor L/2 \rfloor}}{4 M n}
        \sum_{i=1}^n \eps_i \Lambda_{M,y_i}^\ast
    = \frac{C^L \, n^{\lfloor L/2 \rfloor} N}{4 M n}
      \cdot f_{\boldnu,J}^{(0)} ;
  \]
  here the final equality comes from our choice of $\eps_1,\dots,\eps_n$ and $z_1,\dots,z_n$.

  To complete the proof, we first collect a few auxiliary estimates.
  First, we see because of ${|J| \geq m^{\theta \lambda}}$ that
  $n \geq n_0 \, m^{(1-\lambda) \theta } \, m^{\theta \lambda} \geq m^\theta$.
  Thus, setting $C_2 := 16 \sigma C_1$ and recalling that
  ${\omega \leq \theta \cdot (\gamma - \lambda) - 1}$ by choice of $\omega$,
  we see for any $0 < \kappa \leq C_2^{-1}$ that
  \[
    \kappa \cdot m^\omega
    \leq \frac{m^{\theta \gamma - \theta \lambda - 1}}
              {16 \sigma C_1}
    \leq \frac{C_1^{-1} n^\gamma \cdot \sigma^{-1} m^{-\theta \lambda}}{4 \cdot 4 m}
    \leq \frac{C^L n^{\lfloor L/2 \rfloor} \cdot \sigma^{-1} m^{-\theta \lambda}}{4 M}
    \leq \frac{C^L n^{\lfloor L/2 \rfloor} N}{4 M n} .
  \]
  Here, we used in the last step that $|J| \leq \sigma \, m^{\theta \lambda}$, which implies
  \(
    \frac{N}{n}
    = |J|^{-1}
    \geq \sigma^{-1} m^{-\theta \lambda} .
  \)
  Thus, noting that $c \NNSigma[t] \subset \NNSigma$ for $c \in [-1,1]$,
  we see $\kappa \, m^\omega \, f_{\boldnu,J}^{(0)} \in \NNSigma[(2L+8)n]$
  as long as $0 < \kappa \leq C_2^{-1}$.

  Finally, set $C_3 := \max \bigl\{ 1, \,\, C_2, \,\, (2 L + 8)^\alpha \, (2 n_0 \sigma)^\alpha \bigr\}$.
  We claim that $\PreNorm[\infty]\bigl(\kappa \, m^\omega \, f_{\boldnu,J}^{(0)}\bigr) \leq 1$
  for $\kappa := C_3^{-1}$.
  Once this is shown, \Cref{lem:ApproximationSpaceProperties} will show that
  $\big\| \kappa \, m^\omega \, f_{\boldnu,J}^{(0)} \big\|_{\ApproxSpace[\infty]} \leq 1$ as well.
  To see $\PreNorm[\infty]\bigl(\kappa \, m^\omega \, f_{\boldnu,J}^{(0)}\bigr) \leq 1$,
  first note that
  \(
    \big\| \kappa \, m^\omega \, f_{\boldnu,J}^{(0)} \big\|_{L^\infty}
    \leq \| f_{\boldnu,J}^{(0)} \|
    \leq 1 ,
  \)
  since $\omega < 0$ and $\kappa = C_3^{-1} \leq 1$.
  Furthermore, for $t \in \N$ there are two cases:
  For $t \geq (2 L + 8) n$ we have shown above that
  $\kappa \, m^\omega \, f_{\boldnu,J}^{(0)} \in \NNSigma[(2L+8)n] \subset \NNSigma[t]$
  and hence $t^\alpha \, d_\infty(\kappa \, m^\omega \, f_{\boldnu,J}^{(0)}) = 0 \leq 1$.
  On the other hand, if $t \leq (2 L + 8) n$ then we see because of
  \(
    \big\lceil m^{(1-\lambda) \theta} \big\rceil
    \leq 1 + m^{(1-\lambda) \theta}
    \leq 2 \cdot m^{(1-\lambda) \theta}
  \)
  and $|J| \leq \sigma \, m^{\theta \lambda}$ that $n \leq 2 n_0 \sigma \, m^\theta$.
  Since we also have $\omega \leq - \theta \alpha$, this implies
  \[
    t^\alpha \, d_\infty \bigl(\kappa \, m^\omega \, f_{\boldnu,J}^{(0)}\bigr)
    \leq (2 L + 8)^\alpha \,
         n^\alpha \,
         \kappa \,
         m^\omega \,
         \big\| f_{\boldnu,J}^{(0)} \big\|_{L^\infty}
    \leq (2L+8)^\alpha (2 n_0 \sigma)^\alpha \, \kappa \, m^{\theta \alpha} m^{-\theta \alpha}
    \leq 1 .
  \]
  All in all, this shows $\PreNorm[\infty]\bigl(\kappa \, m^\omega \, f_{\boldnu,J}^{(0)}\bigr) \leq 1$.
  As seen above, this completes the proof.
\end{proof}

For later use, we also collect the following technical result which shows
how to select a large number of ``hat functions'' as in \Cref{lem:HatSumsInUnitBall}
that are annihilated by a given set of sampling points.

\begin{lemma}\label{lem:AvoidingSamplingPoints}
  Let $m \in \N$ and let $M = 4 m$ and $z_j = \frac{1}{4 m} + \frac{j-1}{2 m}$
  as in \Cref{lem:HatSumsInUnitBall}.
  Given arbitrary points $\x = (x_1,\dots,x_m) \in ([0,1]^d)^m$, define
  \[
    I_{\x}
    := \big\{
         i \in \FirstN{2 m}
         \,\,\,\colon\,\,\,
         \forall \, n \in \FirstN{m}:
           \Lambda_{M,z_i}^\ast (x_n) = 0
       \big\}.
  \]
  Then $|I_{\x}| \geq m$.
\end{lemma}

\begin{proof}
  Let $I_{\x}^c := \FirstN{2m} \setminus I_{\x}$.
  For each $i \in I_{\x}^c$, there exists $n_i \in \FirstN{m}$ satisfying
  $\Lambda_{M,z_i}^\ast (x_{n_i}) \neq 0$.
  The map $I_{\x}^c \to \FirstN{m}, i \mapsto n_i$ is injective, since
  $\Lambda_{M,z_i}^\ast \Lambda_{M,z_\ell}^\ast \equiv 0$ for $i \neq \ell$
  (see \Cref{lem:HatSumsInUnitBall}).
  Therefore, $|I_{\x}^c| \leq m$ and hence $|I_{\x}| = 2m - |I_{\x}^c| \geq m$.
\end{proof}

The function $\Lambda_{M,y}^\ast : \R^d \to \R$ has a controlled support with respect to the
first coordinate of $x$, but unbounded support with respect to the remaining variables.
For proving more refined hardness bounds, we shall therefore use the following modified construction
of a function of ``hat-type'' with controlled support.
As we will see in \Cref{lem:MultidimensionalHatImplementation} below,
this function can also be well implemented by ReLU networks,
provided one can use networks \emph{with at least two hidden layers}.

\begin{lemma}\label{lem:MultidimensionalHatProperties}
  Given $d \in \N$, $M > 0$ and $y \in \R^d$, define
  \[
    \begin{alignedat}{5}
      && \theta : \quad
      & \R \to [0,1], \quad
      && x \mapsto \varrho(x) - \varrho(x-1) , \\
      && \Delta_{M,y} : \quad
      & \R^d \to \R, \quad
      && x \mapsto \bigg[ \sum_{j=1}^d \Lambda_{M,y_j}(x_j) \bigg] - (d - 1) , \\
      \quad \text{ and } \quad
      && \vartheta_{M,y} : \quad
      & \R^d \to [0,1], \quad
      && x \mapsto \theta\bigl(\Delta_{M,y}(x)\bigr) .
    \end{alignedat}
  \]
  Then the function $\vartheta_{M,y}$ has the following properties:
  \begin{enumerate}[label=\alph*)]
    \item $\vartheta_{M,y}(x) = 0$ for all $x \in \R^d \setminus \bigl(y + M^{-1} (-1,1)^d\bigr)$;
    \item $\| \vartheta_{M,y} \|_{L^p (\R^d)} \leq (2 / M)^{d/p}$ for arbitrary $p \in (0,\infty]$;
    \item For any $p \in (0,\infty]$ there is a constant $C = C(d,p) > 0$ satisfying
          \[
            \| \vartheta_{M,y} \|_{L^p([0,1]^d)}
            \geq C \cdot M^{-d/p},
            \qquad
            \forall \, y \in [0,1]^d \text{ and } M \geq \tfrac{1}{2d} .
          \]
  \end{enumerate}
\end{lemma}

\begin{figure}[h]
  \begin{center}
    \includegraphics[width=0.35\textwidth]{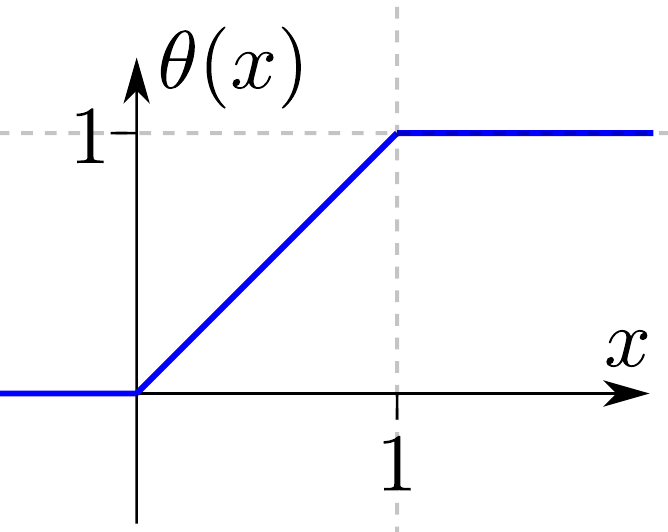}
  \end{center}
  \caption{\label{fig:ThetaPlot}A plot of the function $\theta$
  appearing in \Cref{lem:MultidimensionalHatProperties}.
  Note that $\theta$ is non-decreasing and satisfies $\theta(x) = 0$ for $x \leq 0$
  as well as $\theta(x) = 1$ for $x \geq 1$.}
\end{figure}

\begin{proof}[Proof of \Cref{lem:MultidimensionalHatProperties}]
  \textbf{Ad a)}
  For $x \in \R^d \setminus \big( y + M^{-1} (-1,1)^d \big)$, there exists $\ell \in \FirstN{d}$
  with $|x_\ell - y_\ell| \geq M^{-1}$ and hence $\Lambda_{M,y_\ell}(x_\ell) = 0$;
  see \Cref{fig:LambdaPlot}.
  Because of $0 \leq \Lambda_{M,y_j} \leq 1$, this implies
  \[
    \Delta_{M,y}(x)
    = \sum_{j \in \FirstN{d} \setminus \{ \ell \}}
        \Lambda_{M,y_j} (x_j)
      - (d - 1)
    \leq d-1 - (d-1)
    =    0 .
  \]
  By elementary properties of the function $\theta$ (see \Cref{fig:ThetaPlot}), this shows
  $\vartheta_{M,y}(x) = \theta(\Delta_{M,y}(x)) = 0$.

  %

  \medskip{}

  \textbf{Ad b)} Since $0 \leq \theta \leq 1$, we also have $0 \leq \vartheta_{M,y} \leq 1$.
  Combined with Part~a), this implies
  \(
    \| \vartheta_{M,y} \|_{L^p}
    \leq \bigl[\LebesgueMeasure(y + M^{-1}(-1,1)^d)\bigr]^{1/p}
    =    (2/M)^{d/p} ,
  \)
  as claimed.

  \medskip{}

  \textbf{Ad c)} Set $T := \frac{1}{2 d M} \in (0,1]$ and $P := y + [-T,T]^d$.
  For $x \in P$ and arbitrary $j \in \FirstN{d}$, we have $|x_j - y_j| \leq \frac{1}{2 d M}$.
  Since $\Lambda_{M,y_j}$ is Lipschitz with $\Lip(\Lambda_{M,y_j}) \leq M$
  (see \Cref{fig:LambdaPlot}) and $\Lambda_{M,y_j}(y_j) = 1$, this implies
  \[
    \Lambda_{M,y_j}(x_j)
    \geq \Lambda_{M,y_j}(y_j) - \bigl|\Lambda_{M,y_j}(y_j) - \Lambda_{M,y_j}(x_j)\bigr|
    \geq 1 - M \cdot \frac{1}{2 d M}
    =    1 - \frac{1}{2 d} .
  \]
  Since this holds for all $j \in \FirstN{d}$, we see
  \({
    \Delta_{M,y}(x)
    = \sum_{j=1}^d \Lambda_{M,y_j}(x_j) - (d \!-\! 1)
    \geq d \!\cdot\! (1 \!-\! \frac{1}{2 d}) - (d \!-\! 1)
    = \frac{1}{2}
    ,
  }\)
  and hence $\vartheta_{M,y}(x) = \theta(\Delta_{M,y}(x)) \geq \theta(\frac{1}{2}) = \frac{1}{2}$,
  since $\theta$ is non-decreasing.

  Finally, \Cref{lem:CubeIntersectionMeasure} shows that
  $\LebesgueMeasure(Q \cap P) \geq 2^{-d} T^d \geq C_1 \cdot M^{-d}$ with $C_1 = C_1(d) > 0$.
  Hence,
  \(
    \| \vartheta_{M,y} \|_{L^p([0,1]^d)}
    \geq \frac{1}{2} [\LebesgueMeasure(Q \cap P)]^{1/p}
    \geq C_1^{1/p} M^{-d/p} ,
  \)
  which easily yields the claim.
\end{proof}

The next lemma shows how well the function $\vartheta_{M,y}$ can be implemented by ReLU networks.
We emphasize that the lemma requires using networks with $L \geq 3$,
i.e., with at least two hidden layers.

\begin{lemma}\label{lem:MultidimensionalHatImplementation}
  Let $\LayerFunc : \N \to \N_{\geq 2} \cup \{ \infty \}$
  and $\CoeffFunc : \N \to \N \cup \{ \infty \}$ be non-decreasing.
  Let $M \geq 1$, $n \in \N$ and $0 < C \leq \CoeffFunc(n)$, as well as $L \in \N_{\geq 3}$
  with $L \leq \LayerFunc(n)$.
  Then
  \[
    \frac{C^L \cdot n^{\lfloor L/2 \rfloor}}{4 M}
    \cdot \vartheta_{M,y}
    \in \NNSigma[15 (d + L) n]
    \qquad \forall \, y \in [0,1]^d .
  \]
\end{lemma}

\begin{proof}
  Let $y \in [0,1]^d$ be fixed.
  For $j \in \FirstN{d}$, denote by $e_j \in \R^{d \times 1}$ the $j$-th standard basis vector.
  Define $A_1 \in \R^{4 n d \times d}$ and $b_1 \in \R^{4 n d}$ by
  \begin{align*}
    A_1^T
    & := \frac{C}{2}
         \cdot \Big(
                 \underbrace{e_1 \big| \dots \big| e_1}_{3n \text{ times}}, \,\,
                 \underbrace{0 \,\big| \dots \big| \, 0}_{n \text{ times}},
                 \quad
                 \underbrace{e_2 \big| \dots \big| e_2}_{3n \text{ times}}, \,\,
                 \underbrace{0 \,\big| \dots \big| \, 0}_{n \text{ times}},
                 \quad
                 \dots,
                 \quad
                 \underbrace{e_d \big| \dots \big| e_d}_{3n \text{ times}}, \,\,
                 \underbrace{0 \,\big| \dots \big| \, 0}_{n \text{ times}}
               \Big), \\
    b_1
    & := -\frac{C}{2}
         \cdot \Big(
                 \underbrace{y_1 - \tfrac{1}{M} , \dots , y_1 - \tfrac{1}{M}}_{n \text{ times}}, \,\,
                 \underbrace{y_1 , \dots , y_1}_{n \text{ times}}, \,\,
                 \underbrace{y_1 + \tfrac{1}{M} , \dots , y_1 + \tfrac{1}{M}}_{n \text{ times}},
                 \underbrace{-1, \dots , -1}_{n \text{ times}}, \\
    &            \quad\qquad\qquad
                 \underbrace{y_2 - \tfrac{1}{M} , \dots , y_2 - \tfrac{1}{M}}_{n \text{ times}}, \,\,
                 \underbrace{y_2 , \dots , y_2}_{n \text{ times}}, \,\,
                 \underbrace{y_2 + \tfrac{1}{M} , \dots , y_2 + \tfrac{1}{M}}_{n \text{ times}},
                 \underbrace{-1, \dots , -1}_{n \text{ times}}, \\
    &            \quad\qquad\qquad \dots, \\
    &            \quad\qquad\qquad
                 \underbrace{y_d - \tfrac{1}{M} , \dots , y_d - \tfrac{1}{M}}_{n \text{ times}},
                 \underbrace{y_d , \dots , y_d}_{n \text{ times}},
                 \underbrace{y_d + \tfrac{1}{M} , \dots , y_d + \tfrac{1}{M}}_{n \text{ times}}
                 \underbrace{-1, \dots , -1}_{n \text{ times}}
               \Big)^T
  \end{align*}
  Furthermore, set $b_2 := 0 \in \R^2$ and $b_3 := 0 \in \R^n$,
  let $\zeta := -\frac{1}{M}\frac{d-1}{d}$ and $\xi := -\frac{1}{M}$,
  and define $A_2 \in \R^{2 \times 4 n d}$ and $A_3 \in \R^{n \times 2}$ by
  \begin{align*}
    A_2
    & := \frac{C}{2}
         \bigg(\,
           \begin{matrix}
             \overbrace{1,\dots,1}^{n \text{ times}},
             \overbrace{-2,\dots,-2}^{n \text{ times}},
             \overbrace{1,\dots,1}^{n \text{ times}},
             \overbrace{\zeta,\dots,\zeta}^{n \text{ times}},
             \dots,
             \overbrace{1,\dots,1}^{n \text{ times}},
             \overbrace{-2,\dots,-2}^{n \text{ times}},
             \overbrace{1,\dots,1}^{n \text{ times}},
             \overbrace{\zeta,\dots,\zeta}^{n \text{ times}}
             \\
             \underbrace{1,\dots,1}_{n \text{ times}},
             \underbrace{-2,\dots,-2}_{n \text{ times}},
             \underbrace{1,\dots,1}_{n \text{ times}},
             \underbrace{\xi,\dots,\xi}_{n \text{ times}},
             \dots,
             \underbrace{1,\dots,1}_{n \text{ times}},
             \underbrace{-2,\dots,-2}_{n \text{ times}},
             \underbrace{1,\dots,1}_{n \text{ times}},
             \underbrace{\xi,\dots,\xi}_{n \text{ times}}
           \end{matrix}
         \,\bigg), \\
    A_3^T
    & := C
         \bigg(
           \begin{matrix}
             1, \dots, 1 \\
             -1,\dots,-1
           \end{matrix}
         \bigg)
      \in \R^{2 \times n} .
  \end{align*}
  Finally, set $A := C \cdot (1,\dots,1) \in \R^{1 \times n}$,
  $B := C \cdot (1,\dots,1)^T \in \R^{n \times 1}$, and $D := C \cdot (1 , -1) \in \R^{1 \times 2}$,
  as well as $E := (C) \in \R^{1 \times 1}$.
  Note that
  \(
    \| A_1 \|_{\infty},
    \| A_2 \|_{\infty},
    \| A_3 \|_{\infty},
    \| A \|_{\infty},
    \| B \|_{\infty},
    \| D \|_{\infty},
    \| E \|_{\infty}
    \leq C
  \)
  and $\| b_1 \|_{\infty}, \| b_2 \|_{\infty} \leq C$, since $M \geq 1$ and $y \in [0,1]^d$.
  Furthermore, note $\| A_1 \|_{\ell^0} \leq 3 d n$, $\| A_2 \|_{\ell^0} \leq 8 d n$,
  $\| A_3 \|_{\ell^0} \leq 2 n$, $\| A \|_{\ell^0}, \| B \|_{\ell^0} \leq n$,
  $\| D \|_{\ell^0} \leq 2$, and finally $\| b_1 \|_{\ell^0} \leq 4 d n$ and $\| b_2 \|_{\ell^0} = 0$.
  Furthermore, note $C \leq \CoeffFunc(n) \leq \CoeffFunc(15 (d + L) n)$
  and likewise $L \leq \LayerFunc(n) \leq \LayerFunc(15 (d+L) n)$ thanks to the monotonicity
  of $\CoeffFunc,\LayerFunc$.

  A direct computation shows that
  \[
    \tfrac{C/2}{M} \Lambda_{M,y}(x)
    = \varrho\bigl(\tfrac{C}{2} (x - y + \tfrac{1}{M})\bigr)
      - 2 \varrho\bigl(\tfrac{C}{2} (x - y)\bigr)
      + \varrho\bigl(\tfrac{C}{2} (x - y - \tfrac{1}{M})\bigr)
    .
  \]
  Combined with the positive homogeneity of the ReLU
  (i.e., $\varrho(t x) = t \varrho(x)$ for $t \geq 0$), this shows
  \begin{align*}
    & \bigl(A_2 \, \varrho(A_1 x + b_1) + b_2\bigr)_1 \\
    & \!
      = \frac{C}{2}
        \sum_{j=1}^d
          \sum_{\ell=1}^n
          \Big[
            \varrho\bigl(\tfrac{C}{2} (\langle x,e_j \rangle - (y_j \!-\! \tfrac{1}{M}))\bigr)
            -2 \varrho \bigl( \tfrac{C}{2} (\langle x, e_j \rangle - y_j) \bigr)
            + \varrho \bigl( \tfrac{C}{2} (\langle x, e_j \rangle - (y_j \!+\! \tfrac{1}{M})) \bigr)
            + \zeta \, \varrho(\tfrac{C}{2})
          \Big] \\
    & \!
      = \frac{C^2 n}{4 M}
        \sum_{j=1}^d
        \Big[
          \Lambda_{M,y_j} (x_j)
          - \frac{d-1}{d}
        \Big]
      = \frac{C^2 n}{4 M} \Delta_{M,y} (x) .
  \end{align*}
  In the same way, it follows that
  $\bigl(A_2 \, \varrho(A_1 x + b_1) + b_2\bigr)_2 = \frac{C^2 n}{4 M} \cdot (\Delta_{M,y}(x) - 1)$.
  We now distinguish three cases:

  \textbf{Case 1:} $L=3$.
  In this case, set $\Phi := \big( (A_1,b_1), (A_2,b_2), (D,0) \big)$.
  Then the calculation from above, combined with the positive homogeneity of the ReLU shows
  \[
    R_\varrho \Phi (x)
    = C \cdot
      \Big(
        \varrho\bigl(\tfrac{C^2 n}{4 M} \Delta_{M,y}(x)\bigr)
        - \varrho\bigl(\tfrac{C^2 n}{4 M} (\Delta_{M,y}(x) - 1)\bigr)
      \Big)
    = \tfrac{C^3 n}{4 M} \theta(\Delta_{M,y}(x))
    = \tfrac{C^3 n}{4 M} \vartheta_{M,y}(x) .
  \]
  Furthermore, it is straightforward to see
  $W(\Phi) \leq 3 d n + 4 d n + 8 d n + 2 \leq 2 + 15 d n \leq 15 (L + d) n$.
  Combined with our observations from above, and noting $\lfloor \frac{L}{2} \rfloor = 1$,
  we thus see as claimed that
  $\frac{C^L \cdot n^{\lfloor L/2 \rfloor}}{4 M} \vartheta_{M,y} \in \NNSigma[15 (L + d) n]$.

  \medskip{}
  \textbf{Case 2:} $L \geq 4$ is even.
  In this case, define
  \[
    \Phi
    = \Big(
        (A_1,b_1),
        (A_2,b_2),
        (A_3,b_3),
        (A,0),
        \underbrace{
          (B,0),(A,0),\dots,(B,0),(A,0)
        }_{(L-4)/2 \text{ copies of ``} (B,0),(A,0)\text{''}}
      \Big)
  \]
  Similar arguments as in Case~1 show that
  \(
    \bigl(
      A_3 \, \varrho \bigl( A_2 \, \varrho(A_1 x + b_1) + b_2 \bigr)
      + b_3
    \bigr)_j
    = \frac{C^3 n}{4 M} \, \vartheta_{M,y}(x)
  \)
  for all $j \in \FirstN{n}$, and hence
  \(
    A \circ \varrho \circ A_3 \circ \varrho \circ A_2 \circ (A_1 \bullet + b_1)
    = \frac{C^4 n^2}{4 M} \vartheta_{M,y}
    .
  \)
  Furthermore, using similar arguments as in \Cref{eq:AmplificationTrick},
  we see for $z \in [0,\infty)$ that $A (\varrho(B z)) = C^2 n z$.
  Combining all these observations, we see
  \[
    R_\varrho \Phi (x)
    = (C^2 n)^{(L-4)/2} \cdot \frac{C^4 n^2}{4 M} \vartheta_{M,y}(x)
    = \frac{C^L \cdot n^{\lfloor L/2 \rfloor}}{4 M} \cdot \vartheta_{M,y} (x) .
  \]
  Since also
  $W(\Phi) \leq 3 d n + 4 d n + 8 d n + 2 n + n + \frac{L-4}{2} \cdot 2 n \leq 15 (d + L) n$,
  we see overall as claimed that
  $\frac{C^L \cdot n^{\lfloor L/2 \rfloor}}{4 M} \vartheta_{M,y} \in \NNSigma[15 (d + L) n]$.

  \medskip{}
  \textbf{Case 3:} $L \geq 5$ is odd.
  In this case, define
  \[
    \Phi
    := \Big(
         (A_1,b_1),
         (A_2,b_2),
         (A_3,b_3),
         (A,0),
         \underbrace{
           (B,0),(A,0),\dots,(B,0),(A,0)
         }_{(L-5)/2 \text{ copies of ``} (B,0),(A,0) \text{''}} ,
         (E, 0)
       \Big) .
  \]
  A variant of the arguments in Case~2 shows that
  \(
    R_\varrho \Phi
    = C \cdot (C^2 \, n)^{(L-5)/2} \cdot \frac{C^4 n^2}{4 M} \vartheta_{M,y}
    = \frac{C^L \cdot n^{\lfloor L/2 \rfloor}}{4 M} \vartheta_{M,y}
  \)
  and $W(\Phi) \leq 15 d n + 2 n + \frac{L-5}{2} \cdot 2 n + 1 \leq 15 (d + L) n$, and hence
  $\frac{C^L \cdot n^{\lfloor L/2 \rfloor}}{4 M} \vartheta_{M,y} \in \NNSigma[15 (d + L) n]$
  also in this last case.
\end{proof}

\begin{lemma}\label{lem:MultidimensionalHatInSpace}
  Let $\LayerFunc,\CoeffFunc : \N \to \N \cup \{ \infty \}$ be non-decreasing
  with $\LayerFunc^\ast \geq 3$.
  Let $d \in \N$, $\alpha \in (0,\infty)$, and $0 < \gamma < \gamma^{\flat}(\LayerFunc,\CoeffFunc)$.
  Then there exists a constant $\kappa = \kappa(\gamma,\alpha,d,\LayerFunc,\CoeffFunc) > 0$
  such that for any $M \in [1,\infty)$, we have
  \[
    g_{M,y}
    := \kappa \cdot M^{-\alpha / (\alpha + \gamma)} \, \vartheta_{M,y}
    \in \ApproxSpace[\infty]
    \qquad \text{with} \qquad
    \big\| g_{M,y} \big\|_{\ApproxSpace[\infty]} \leq 1.
  \]
\end{lemma}

\begin{proof}
  Since $\gamma < \gamma^{\flat}(\LayerFunc,\CoeffFunc)$,
  there exist $L = L(\gamma,\LayerFunc,\CoeffFunc) \in \N_{\geq \ell^\ast}$
  and $C_1 = C_1 (\gamma,\LayerFunc,\CoeffFunc) > 0$ satisfying
  $n^\gamma \leq C_1 \cdot (\CoeffFunc (n))^L \cdot n^{\lfloor L/2 \rfloor}$ for all $n \in \N$.
  Since $\LayerFunc^\ast \geq 3$, we can assume without loss of generality that $L \geq 3$.
  Furthermore, since $L \leq \ell^\ast$, there exists
  $n_0 = n_0(\gamma,\LayerFunc,\CoeffFunc) \in \N$ satisfying $L \leq \LayerFunc(n_0)$.

  Given $M \in [1,\infty)$, set $n := n_0 \cdot \big\lceil M^{1/(\alpha+\gamma)} \big\rceil$,
  noting that $n \geq n_0$.
  Since $n^\gamma \leq C_1 \cdot (\CoeffFunc(n))^L \cdot n^{\lfloor L/2 \rfloor}$,
  there exists $0 < C \leq \CoeffFunc(n)$ satisfying
  $n^\gamma \leq C_1 \cdot C^L n^{\lfloor L/2 \rfloor}$.

  Set $\kappa := \min \{ (15 (d+L))^{-\alpha} (2 n_0)^{-\alpha}, \, (4 \, C_1)^{-1} \} > 0$
  and note $\kappa = \kappa(d,\alpha,\gamma,\LayerFunc,\CoeffFunc)$.
  Furthermore, note that $n \geq M^{1/(\alpha + \gamma)}$ and hence
  \(
    \kappa \, M^{-\frac{\alpha}{\alpha+\gamma}}
    = \frac{\kappa}{M} \, M^{\frac{\gamma}{\alpha+\gamma}}
    \leq \kappa \, \frac{n^\gamma}{M}
    \leq 4 C_1 \, \kappa \, \frac{C^L \, n^{\lfloor L/2 \rfloor}}{4 M}
    \leq \frac{C^L \, n^{\lfloor L/2 \rfloor}}{4 M} .
  \)
  Combining this with the inclusion $c \NNSigma[t] \subset \NNSigma[t]$ for $c \in [-1,1]$,
  we see from \Cref{lem:MultidimensionalHatImplementation}
  and because of $3 \leq L \leq \LayerFunc(n_0) \leq \LayerFunc(n)$ that
  $g_{M,y} = \kappa \, M^{-\alpha/(\alpha+\gamma)} \, \vartheta_{M,y} \in \NNSigma[15 (d+L) n]$

  We claim that $\PreNorm[\infty](g_{M,y}) \leq 1$.
  To see this, first note $\| g_{M,y} \|_{L^\infty} \leq \| \vartheta_{M,y} \|_{L^\infty} \leq 1$.
  Furthermore, for $t \in \N$, there are two cases:
  For $t \geq 15 (d+L) n$, we have $g_{M,y} \in \NNSigma[t]$, and hence
  $t^\alpha \, d_{\infty} (g_{M,y}, \NNSigma[t]) = 0 \leq 1$.
  On the other hand, if $t \leq 15 (d+L) n$, then we see because of
  $n \leq 1 + n_0 \, M^{1/(\alpha+\gamma)} \leq 2 n_0 \, M^{1/(\alpha+\gamma)}$ that
  \begin{align*}
    t^\alpha \, d_\infty (g_{M,y}, \NNSigma[t])
    & \leq \bigl(15 (d+L)\bigr)^\alpha \, n^\alpha \, \| g_{M,y} \|_{L^\infty}
      \leq \bigl(15 (d+L)\bigr)^\alpha \, \kappa \, n^\alpha \, M^{-\alpha / (\alpha+\gamma)} \\
    & \leq \bigl(15 (d+L)\bigr)^\alpha \,
           (2 n_0)^\alpha \,
           \kappa \,
           M^{\alpha/(\alpha+\gamma)} \,
           M^{-\alpha / (\alpha+\gamma)}
      \leq 1
      .
  \end{align*}
  Overall, this shows $\PreNorm(g_{M,y}) \leq 1$, so that \Cref{lem:ApproximationSpaceProperties}
  shows as claimed that $\| g_{M,y} \|_{\ApproxSpace[\infty]} \leq 1$.
\end{proof}


\section{Error bounds for uniform approximation}
\label{sec:UniformApproximationErrorBounds}


In this section, we derive an upper bound on how many point samples of a function
$f \in \ApproxSpace[\infty]$ are needed in order to uniformly approximate $f$
up to error $\eps \in (0,1)$.
The crucial ingredient will be the following estimate of the Lipschitz constant
of functions $F \in \NNSigma$.
The bound in the lemma is one of the reasons for our choice of the quantities
$\gamma^{\flat}$ and $\gamma^{\sharp}$ introduced in \Cref{eq:GammaDefinition}.

\begin{lemma}\label{lem:NetworkLipschitzEstimate}
  Let $\LayerFunc : \N \to \N \cup \{ \infty \}$ and $\CoeffFunc : \N \to [1,\infty]$.
  Let $n \in \N$ and assume that $L := \LayerFunc(n)$ and $C := \CoeffFunc(n)$ are finite.
  Then each $F \in \NNSigma$ satisfies
  \[
    \Lip_{(\R^d, \| \cdot \|_{\ell^1}) \to \R}(F) \leq C^L \cdot n^{\lfloor L/2 \rfloor}
    \quad \text{ and } \quad
    \Lip_{(\R^d, \| \cdot \|_{\ell^\infty}) \to \R}(F) \leq d \cdot C^L \cdot n^{\lfloor L/2 \rfloor} .
  \]
\end{lemma}

\begin{proof}
  \textbf{Step~1:} For any matrix $A \in \R^{k \times m}$,
  define $\| A \|_{\infty} := \max_{i,j} |A_{i,j}|$ and denote by $\| A \|_{\ell^0}$
  the number of non-zero entries of $A$.
  In this step, we show that
  \begin{equation}
    \| A \|_{\ell^1 \to \ell^\infty} \leq \| A \|_{\infty}
    \quad \text{ and } \quad
    \| A \|_{\ell^\infty \to \ell^1} \leq \| A \|_{\infty} \, \| A \|_{\ell^0} .
    \label{eq:MatrixOperatorNormEstimates}
  \end{equation}
  To prove the first part, note for arbitrary $x \in \R^m$ and any $i \in \FirstN{k}$ that
  \[
    \bigl|(A x)_i\bigr|
    \leq \sum_{j=1}^m
           |A_{i,j}| \, |x_j|
    \leq \| A \|_{\infty} \, \sum_{j=1}^m |x_j|
    =    \| A \|_{\infty} \, \| x \|_{\ell^1} ,
  \]
  showing that $\| A x \|_{\ell^\infty} \leq \| A \|_{\infty} \, \| x \|_{\ell^1}$.
  To prove the second part, note for arbitrary $x \in \R^m$ that
  \[
    \| A x \|_{\ell^1}
    = \sum_{i=1}^k
        \bigl|(A x)_i\bigr|
    \leq \sum_{i,j}
           |A_{i,j}| \, |x_j|
    \leq \| x \|_{\ell^\infty} \,
         \| A \|_{\infty} \,
         \sum_{i,j}
           \indicator_{A_{i,j} \neq 0}
    =    \| A \|_{\infty} \, \| A \|_{\ell^0} \, \| x \|_{\ell^\infty} .
  \]

  \medskip{}

  \noindent
  \textbf{Step~2 (Completing the proof):}
  Let $F \in \NNSigma$ be arbitrary, so that $F = R_\varrho \Phi$ for a network
  $\Phi = \big( (A_1,b_1),\dots,(A_{\widetilde{L}},b_{\widetilde{L}}) \big)$
  satisfying $\widetilde{L} \leq \LayerFunc(n) = L$ and
  $\| A_j \|_{\infty} \leq \WeightSize{\Phi} \leq \CoeffFunc(n) = C$,
  as well as $\| A_j \|_{\ell^0} \leq W(\Phi) \leq n$ for all $j \in \FirstN{\widetilde{L}}$.

  Set $p_j := 1$ if $j$ is even and $p_j := \infty$ otherwise.
  Choose $N_j$ such that $A_j \in \R^{N_j \times N_{j-1}}$, and define $T_j \, x := A_j \, x + b_j$.
  By Step~1, we then see that
  \(
    T_j :
    \bigl(
      \R^{N_{j-1}}, \| \cdot \|_{\ell^{p_j - 1}}
    \bigr) \to
    \bigl(
      \R^{N_j}, \| \cdot \|_{\ell^{p_j}}
    \bigr)
  \)
  is Lipschitz with
  \[
    \Lip(T_j)
    = \| A_j \|_{\ell^{p_{j-1}} \to \ell^{p_j}}
    \leq \begin{cases}
           \| A_j \|_{\infty} \, \| A_j \|_{\ell^0} \leq C n, & \text{if } j \text{ is even}, \\
           \| A_j \|_{\infty} \leq C ,                        & \text{if } j \text{ is odd}.
         \end{cases}
  \]
  Next, a straightforward computation shows that the ``vector-valued ReLU'' is $1$-Lipschitz
  as a map $\varrho : (\R^k, \| \cdot \|_{\ell^p}) \to (\R^k, \| \cdot \|_{\ell^p})$,
  for arbitrary $p \in [1,\infty]$ and any $k \in \N$.
  As a consequence, we see that
  \[
    F
    = R_\varrho \Phi
    = T_{\widetilde{L}}
      \circ (\varrho \circ T_{\widetilde{L} - 1})
      \circ \cdots
      \circ (\varrho \circ T_1)
    : \quad (\R^d, \| \cdot \|_{\ell^1}) \to (\R, \| \cdot \|_{\ell^{p_{\widetilde{L}}}} ) = (\R, |\cdot|)
  \]
  is Lipschitz continuous as a composition of Lipschitz maps, with overall Lipschitz constant
  \[
    \Lip(R_\varrho \Phi)
    \leq \prod_{j=1}^{\widetilde{L}}
           \bigl(C \cdot n_j\bigr)
    =    C^{\widetilde{L}} \cdot n^{\lfloor \widetilde{L} / 2 \rfloor}
    \leq C^L \cdot n^{\lfloor L / 2 \rfloor} .
  \]
  where we used the notation $n_j := n$ if $j$ is even and $n_j := 1$ otherwise.
  Furthermore, we used in the last step that $C \geq 1$.
  The final claim of the lemma follows from the elementary estimate
  $\| x \|_{\ell^1} \leq d \cdot \| x \|_{\ell^\infty}$ for $x \in \R^d$.
\end{proof}

Based on the preceding lemma, we can now prove an error bound
for the computational problem of uniform approximation on the neural network
approximation space $\ApproxSpace[\infty]([0,1]^d)$.

\begin{theorem}\label{thm:ErrorBoundUniformApproximation}
  Let $\LayerFunc,\CoeffFunc : \N \to \N \cup \{ \infty \}$ be non-decreasing,
  and suppose that $\gamma^{\sharp}(\LayerFunc,\CoeffFunc) < \infty$.
  Let $d \in \N$ and $\alpha \in (0,\infty)$ be arbitrary,
  and let $\UnitBall[\infty]([0,1]^d)$ as in \Cref{eq:UnitBallDefinition}.
  Furthermore, let ${\iota_\infty : \ApproxSpace[\infty]([0,1]^d) \to C([0,1]^d), f \mapsto f}$.
  Then, we have
  \[
    \beta_\ast^{\deterministic} \bigl(\UnitBall[\infty]([0,1]^d), \iota_\infty\bigr)
    \geq \frac{1}{d} \cdot \frac{\alpha}{\gamma^{\sharp}(\LayerFunc,\CoeffFunc) + \alpha} .
  \]
\end{theorem}

\begin{rem*}
  \emph{a)} The proof shows that choosing the uniform grid
  $\{ 0, \frac{1}{N}, \dots, \frac{N-1}{N} \}^d$ as the set of sampling points
  (with $N \sim m^{1/d}$) yields an essentially optimal sampling scheme.

  \medskip{}

  \emph{b)} It is well-known (see \cite[Proposition~3.3]{HeinrichRandomApproximation})
  that the error of an optimal Monte Carlo algorithm is at most two times the error
  of an optimal deterministic algorithm.
  Therefore, the theorem also implies that
  \[
    \beta_\ast^{\MonteCarlo} \bigl(\UnitBall[\infty]([0,1]^d), \iota_\infty\bigr)
    \geq \frac{1}{d} \cdot \frac{\alpha}{\gamma^{\sharp}(\LayerFunc,\CoeffFunc) + \alpha} .
  \]
\end{rem*}

\begin{proof}
  Since $\gamma^{\sharp}(\LayerFunc,\CoeffFunc) < \infty$,
  \Cref{rem:GammaRemark} shows that $L := \LayerFunc^\ast < \infty$.
  Let $\gamma > \gamma^{\sharp}(\LayerFunc,\CoeffFunc) \geq 1$ be arbitrary.
  By definition of $\gamma^{\sharp}(\LayerFunc,\CoeffFunc)$, it follows that there exists some
  $\gamma' \in \bigl(\gamma^{\sharp}(\LayerFunc,\CoeffFunc), \gamma\bigr)$
  and a constant $C_0 = C_0(\gamma', \LayerFunc,\CoeffFunc) = C_0(\gamma,\LayerFunc,\CoeffFunc) > 0$
  satisfying
  \({
    (\CoeffFunc(n))^L \cdot n^{\lfloor L/2 \rfloor}
    \leq C_0 \cdot n^{\gamma'}
    \leq C_0 \cdot n^{\gamma}
  }\)
  for all $n \in \N$.
  Let $m \in \N$ be arbitrary and choose
  \[
    N := \big\lfloor m^{1/d} \big\rfloor \geq 1
    \qquad \text{ and } \qquad
    n := \big\lceil m^{1/(d \cdot (\gamma + \alpha))} \big\rceil \in \N.
  \]
  Furthermore, let $I := \bigl\{ 0, \frac{1}{N}, \dots, \frac{N-1}{N} \bigr\}^d \subset [0,1]^d$
  and set $C := \CoeffFunc(n)$ and $\mu := d \cdot C^L \cdot n^{\lfloor L/2 \rfloor}$,
  noting that $\mu \leq d \, C_0 \, n^{\gamma} =: C_1 \, n^{\gamma}$ and $|I| = N^d \leq m$.

  Next, set
  \(
    B
    := U
    := \UnitBall[\infty] ([0,1]^d)
    = \bigl\{
        f \in \ApproxSpace[\infty]([0,1]^d)
        \colon
        \| f \|_{\ApproxSpace[\infty]}
        \leq 1
      \bigr\}
  \)
  and define $S := \Omega(B)$ for
  \[
    \Omega : \quad
    C([0,1]^d) \to \R^I, \quad
    f \mapsto \big( f(i) \big)_{i \in I} .
  \]
  For each $y = (y_i)_{i \in I} \in S$, choose some $f_y \in B$ satisfying $y = \Omega(f_y)$.
  Note by \Cref{lem:ApproximationSpaceProperties} that $\PreNorm[\infty](f_y) \leq 1$;
  by definition of $\PreNorm[\infty]$, we can thus choose $F_y \in \NNSigma$ satisfying
  $\| f_y - F_y \|_{L^\infty} \leq 2 \cdot n^{-\alpha}$.
  Given this choice, define
  \[
    Q : \quad
    \R^I \to C([0,1]^d), \quad
    y \mapsto \begin{cases}
                F_y, & \text{if } y \in S, \\
                0  , & \text{otherwise} .
              \end{cases}
  \]
  We claim that
  $\| f - Q (\Omega(f)) \|_{L^\infty} \leq C_2 \cdot m^{-\alpha / (d \cdot (\gamma + \alpha))}$
  for all $f \in B$, for a suitable constant $C_2 = C_2(d,\gamma,\LayerFunc,\CoeffFunc)$.
  Once this is shown, it follows that
  \(
    \beta_\ast^{\deterministic} (U, \iota_\infty)
    \geq \frac{1}{d} \frac{\alpha}{\gamma + \alpha} ,
  \)
  which then implies the claim of the theorem,
  since $\gamma > \gamma^{\sharp}(\LayerFunc,\CoeffFunc)$ was arbitrary.

  Thus, let $f \in B$ be arbitrary and set $y := \Omega(f) \in S$.
  By the same arguments as above, there exists $F \in \NNSigma$ satisfying
  $\| f - F \|_{L^\infty} \leq2 \cdot n^{-\alpha}$.
  Now, we see for each $i \in I$ because of $f(i) = (\Omega(f))_i = y_i = (\Omega(f_y))_i = f_y(i)$
  that
  \[
    | F(i) - F_y (i) |
    \leq |F(i) - f(i)| + |f_y (i) - F_y (i)|
    \leq \| F - f \|_{L^\infty} + \| f_y - F_y \|_{L^\infty}
    \leq 4 \cdot n^{-\alpha} .
  \]
  Furthermore, \Cref{lem:NetworkLipschitzEstimate} shows that
  $F - F_y : (\R^d, \| \cdot \|_{\ell^\infty}) \to (\R, |\cdot|)$ is Lipschitz continuous
  with Lipschitz constant at most $2 \mu$.
  Now, given any $x \in [0,1]^d$, we can choose $i = i(x) \in I$ satisfying
  $\| x - i \|_{\ell^\infty} \leq N^{-1}$.
  Therefore,
  \(
    |(F - F_y)(x)|
    \leq \frac{2 \mu}{N} + |(F - F_y)(i)|
    \leq \frac{2 \mu}{N} + 4 \, n^{-\alpha} .
  \)
  Overall, we have thus shown $\| F - F_y \|_{L^\infty} \leq \frac{2 \mu}{N} + 4 \, n^{-\alpha}$,
  which finally implies because of $Q(\Omega(f)) = Q(y) = F_y$ that
  \[
    \big\| f - Q(\Omega(f)) \big\|_{L^\infty}
    \leq \| f - F \|_{L^\infty} + \| F - F_y \|_{L^\infty}
    \leq 6 \, n^{-\alpha} + \frac{2 \mu}{N} .
  \]
  It remains to note that our choice of $N$ and $n$ implies $m^{1/d} \leq 1 + N \leq 2 N$
  and hence $\frac{1}{N} \leq 2 m^{-1/d}$ and furthermore
  $n \leq 1 + m^{1/(d \cdot (\gamma + \alpha))} \leq 2 \, m^{1/(d \cdot (\gamma + \alpha))}$.
  Hence, recalling that $\mu \leq C_1 \, n^{\gamma}$, we see
  \[
    \frac{\mu}{N}
    \leq 2 C_1 m^{-1/d} n^{\gamma}
    \leq 2^{1+\gamma} C_1 m^{\frac{1}{d} (\frac{\gamma}{\gamma+\alpha} - 1)}
    =    2^{1+\gamma} C_1 m^{-\frac{\alpha}{d \cdot (\gamma + \alpha)}}
  \]
  Furthermore, since $n \geq m^{1/(d \cdot (\gamma+\alpha))}$, we also have
  $n^{-\alpha} \leq m^{-\frac{\alpha}{d \cdot (\gamma + \alpha)}}$.
  Combining all these observations, it is easy to see that
  ${\| f - Q(\Omega(f)) \|_{L^\infty} \leq C_2 \cdot m^{-\frac{\alpha}{d \cdot (\gamma+\alpha)}}}$,
  for a suitable constant $C_2 = C_2(d,\gamma,\LayerFunc,\CoeffFunc) > 0$.
  Since $f \in B$ was arbitrary, this completes the proof.
\end{proof}


\section{Hardness of uniform approximation}
\label{sec:UniformApproximationHardness}


In this section, we show that the error bound for uniform approximation
provided by \Cref{thm:ErrorBoundUniformApproximation} is optimal,
at least in the common case where
$\gamma^{\flat}(\LayerFunc,\CoeffFunc) = \gamma^\sharp (\LayerFunc,\CoeffFunc)$
and $\LayerFunc^\ast \geq 3$.
This latter condition means that the approximation for defining the approximation
space $\ApproxSpace[\infty]$ is performed using networks with at least \emph{two hidden layers}.
We leave it as an interesting question for future work
whether a similar result even holds for approximation spaces associated
to shallow networks.

\begin{theorem}\label{thm:UniformApproximationHardness}
  Let $\LayerFunc: \N \to \N_{\geq 2} \cup \{ \infty \}$
  and $\CoeffFunc : \N \to \N \cup \{ \infty \}$ be non-decreasing with $\LayerFunc^\ast \geq 3$.
  Given $d \in \N$ and $\alpha \in (0,\infty)$,
  let $\UnitBall[\infty] = \UnitBall[\infty]([0,1]^d)$ as in \Cref{eq:UnitBallDefinition}
  and consider the embedding
  $\iota_\infty : \ApproxSpace[\infty]([0,1]^d) \hookrightarrow C([0,1]^d)$.
  Then
  \[
    \beta_\ast^{\deterministic} (\UnitBall[\infty], \iota_\infty),
    \beta_\ast^{\MonteCarlo} (\UnitBall[\infty], \iota_\infty)
    \leq \frac{1}{d} \frac{\alpha}{\alpha + \gamma^{\flat}(\LayerFunc,\CoeffFunc)} .
  \]
\end{theorem}

\begin{proof}
  Set $K := [0,1]^d$ and $U := \UnitBall[\infty]$.

  \textbf{Step~1:}
  Let $0 < \gamma < \gamma^{\flat}(\LayerFunc,\CoeffFunc)$.
  Let $m \in \N$ be arbitrary and $\Gamma_m := \FirstN{2k}^d \times \{ \pm 1 \}$,
  where $k := \big\lceil m^{1/d} \big\rceil$.
  In this step, we show that there is a constant
  $\kappa = \kappa(d,\alpha,\gamma,\LayerFunc,\CoeffFunc) > 0$ (independent of $m$)
  and a family of functions $(f_{\ell,\nu})_{(\ell,\nu) \in \Gamma_m} \subset U$
  which satisfies
  \begin{equation}
    \avsum_{(\ell,\nu) \in \Gamma_m}
      \big\| f_{\ell,\nu} - A(f_{\ell,\nu}) \big\|_{L^\infty}
    \geq \kappa \cdot m^{-\frac{1}{d} \frac{\alpha}{\alpha+\gamma}}
    \qquad \forall \, A \in \Alg_m\bigl(U,C([0,1]^d)\bigr) .
    \label{eq:AverageCaseHardnessUniformApproximation}
  \end{equation}

  To see this, set $M := 4 k$, and for $\ell \in \FirstN{2k}^d$ define
  $y^{(\ell)} := \frac{(1,\dots,1)}{4 k} + \frac{\ell - (1,\dots,1)}{2 k} \in \R^d$.
  Then, we have
  \begin{align*}
    y^{(\ell)} + (-M^{-1}, M^{-1})^d
    & = \frac{2}{M} \bigl(\ell - (1,\dots,1)\bigr) + \frac{(1,\dots,1)}{M} + (-M^{-1}, M^{-1})^d \\
    & = \frac{2}{M} \Big( \ell - (1,\dots,1) + (0,1)^d \Big)
      \subset (0,1)^d ,
  \end{align*}
  which shows that the functions $\vartheta_{M,y^{(\ell)}}$, $\ell \in \FirstN{2k}^d$,
  (with $\vartheta_{M,y}$ as defined in \Cref{lem:MultidimensionalHatProperties}),
  have disjoint supports contained in $[0,1]^d$.
  Furthermore, \Cref{lem:MultidimensionalHatInSpace} yields a constant
  $\kappa_1 = \kappa_1(\gamma,\alpha,d,\LayerFunc,\CoeffFunc) > 0$ such that
  \(
    f_{\ell,\nu}
    := \kappa_1
       \cdot M^{-\alpha/(\alpha+\gamma)}
       \cdot \nu
       \cdot \vartheta_{M,y^{(\ell)}} \in U
  \)
  for arbitrary $(\ell,\nu) \in \Gamma_m$.

  To prove \Cref{eq:AverageCaseHardnessUniformApproximation}, let $A \in \Alg_m (U, C([0,1]^d))$
  be arbitrary.
  By definition, there exist $\x = (x_1,\dots,x_m) \in K^m$ and a function $Q : \R^m \to \R$
  satisfying $A(f) = Q(f(x_1),\dots,f(x_m))$ for all $f \in U$.
  Choose
  \(
    I
    := I_{\x}
    := \big\{
         \ell \in \FirstN{2k}^d
         \colon
         \forall \, n \in \FirstN{m} : \vartheta_{M,y^{(\ell)}} (x_n) = 0
       \big\} .
  \)
  Then for each $\ell \in I^c = \FirstN{2k}^d \setminus I$, there exists $n_\ell \in \FirstN{m}$
  such that $\vartheta_{M,y^{(\ell)}}(x_{n_\ell}) \neq 0$.
  Then the map $I^c \to \FirstN{m}, \ell \mapsto n_\ell$ is injective,
  since $\vartheta_{M,y^{(\ell)}} \, \vartheta_{M,y^{(t)}} = 0$ for $t,\ell \in \FirstN{2k}^d$
  with $t \neq \ell$.
  Therefore, $|I^c| \leq m$ and hence $|I| \geq (2k)^d - m \geq m$, because of $k \geq m^{1/d}$.

  Define $h := Q(0,\dots,0)$.
  Then for each $\ell \in I_{\x}$ and $\nu \in \{ \pm 1 \}$, we have $f_{\ell,\nu}(x_n) = 0$
  for all $n \in \FirstN{m}$ and hence $A(f_{\ell,\nu}) = Q(0,\dots,0) = h$.
  Therefore,
  \begin{equation}
    \begin{split}
      & \| f_{\ell,1} - A(f_{\ell,1}) \|_{L^\infty}
        + \| f_{\ell,-1} - A(f_{\ell,-1}) \|_{L^\infty} \\
      & = \| f_{\ell,1} - h \|_{L^\infty}
          + \| - f_{\ell,1} - h \|_{L^\infty}
        = \| f_{\ell,1} - h \|_{L^\infty}
          + \| h + f_{\ell,1} \|_{L^\infty} \\
      & \geq \| f_{\ell,1} - h + h + f_{\ell,1} \|_{L^\infty}
        =    2 \, \| f_{\ell,1} \|_{L^\infty}
        =    2\kappa_1 \cdot M^{-\alpha/(\alpha+\gamma)}
        \qquad \forall \, \ell \in I_{\x} .
    \end{split}
    \label{eq:UniformApproximationHardnessElementaryEstimate}
  \end{equation}
  Furthermore, since $k \leq 1 + m^{1/d} \leq 2 m^{1/d}$, we see $k^d \leq 2^d m$
  and $M = 4 k \leq 8 \, m^{1/d}$ and hence
  \(
    M^{\frac{\alpha}{\alpha+\gamma}}
    \leq 8^{\frac{\alpha}{\alpha+\gamma}} m^{\frac{1}{d} \frac{\alpha}{\alpha+\gamma}}
    .
  \)
  Combining these estimates with \Cref{eq:UniformApproximationHardnessElementaryEstimate}
  and recalling that $|I| \geq m$, we finally see
  \begin{align*}
    \avsum_{(\ell,\nu) \in \Gamma_m}
      \| f_{\ell,\nu} - A(f_{\ell,\nu}) \|_{L^\infty}
    & \geq (2k)^{-d} \,
           \sum_{\ell \in I_{\x}} \,\,
             \avsum_{\nu \in \{ \pm 1 \}}
               \| f_{\ell,\nu} - A(f_{\ell,\nu}) \|_{L^\infty} \\
    & \geq (2k)^{-d}
           \cdot |I|
           \cdot \kappa_1
           \cdot M^{-\frac{\alpha}{\alpha+\gamma}}
      \geq \frac{\kappa_1}{4^d}
           \cdot m^{-1} \, |I| 
           \cdot M^{-\frac{\alpha}{\alpha+\gamma}} \\
    & \geq \frac{\kappa_1 / 8}{4^d}
           \cdot m^{-\frac{1}{d} \frac{\alpha}{\alpha + \gamma}}
      ,
  \end{align*}
  which establishes \Cref{eq:AverageCaseHardnessUniformApproximation}
  for $\kappa := \frac{\kappa_1/8}{4^d}$.

  \medskip{}

  \textbf{Step~2 (Completing the proof):}
  Given \Cref{eq:AverageCaseHardnessUniformApproximation}, a direct application of
  \Cref{lem:MonteCarloHardnessThroughAverageCase} shows that
  \(
    \beta_\ast^{\deterministic}(U,\iota_\infty), \beta_\ast^{\MonteCarlo}(U,\iota_\infty)
    \leq \frac{1}{d} \frac{\alpha}{\alpha + \gamma} .
  \)
  Since this holds for arbitrary $0 < \gamma < \gamma^{\flat}(\LayerFunc,\CoeffFunc)$,
  we easily obtain the claim of the theorem.
\end{proof}

%
%


\section{Error bounds for approximation in \texorpdfstring{$L^2$}{L²}}%
\label{sec:L2ApproximationErrorBounds}


This section provides error bounds for the approximation of functions in
$\ApproxSpace[\infty]([0,1]^d)$ based on point samples, with error measured in $L^2$.
In a nutshell, the argument is based on combining bounds from statistical learning theory
(specifically from \cite{CuckerSmaleMathematicalFoundationsOfLearning})
with bounds for the \emph{covering numbers} of the neural network sets $\NNSigma$.

For completeness, we mention that the \emph{$\eps$-covering number} $\Covering (\Sigma, \eps)$
(with $\eps > 0$) of a (non-empty) subset $\Sigma$ of a metric space $(X,d)$
is the minimal number $N \in \N$ for which there exist $f_1,\dots,f_N \in \Sigma$ satisfying
$\Sigma \subset \bigcup_{j=1}^N \ClosedBall_\eps (f_j)$.
Here, $\ClosedBall_\eps (f) := \{ g \in X \colon d(f,g) \leq \eps \}$.
If no such $N \in \N$ exists, then $\Covering(\Sigma,\eps) = \infty$.
If we want to emphasize the metric space $X$, we also write $\Covering_X (\Sigma,\eps)$.

For the case where one considers networks \emph{of a given architecture}, bounds
for the covering numbers of network sets have been obtained for instance
in \cite[Proposition~2.8]{BlackScholesGeneralizationError}.
Here, however, we are interested in \emph{sparsely connected networks with unspecified architecture}.
For this case, the following lemma provides covering bounds.

\begin{lemma}\label{lem:NetworkSetCoveringBound}
  Let $\LayerFunc : \N \to \N_{\geq 2}$ and $\CoeffFunc : \N \to \N$ be non-decreasing.
  The covering numbers of the neural network set $\NNSigma$
  (considered as a subset of the metric space $C([0,1]^d)$)
  can be estimated by
  \[
    \Covering_{C([0,1]^d)} (\NNSigma, \eps)
    \leq \Bigl(
           \tfrac{44}{\eps}
           \cdot \bigl(\LayerFunc(n)\bigr)^4
           \cdot \bigl(\CoeffFunc(n) \max \{ d,n \}\bigr)^{1 + \LayerFunc(n)}
         \Bigr)^n
  \]
  for arbitrary $\eps \in (0,1]$ and $n \in \N$.
\end{lemma}


\begin{proof}
  Define $L := \LayerFunc(n)$ and $R := \CoeffFunc(n)$
  We will use some results and notation from \cite{BlackScholesGeneralizationError}.
  Precisely, given a network architecture $\Architecture = (a_0,\dots,a_K) \in \N^{K+1}$,
  we denote by
  \[
    \CalNN (\Architecture)
    := \prod_{j=1}^K
         \big(
           [-R,R]^{a_j \times a_{j-1}} \times [-R,R]^{a_j}
         \big)
  \]
  the set of all network weights with architecture $\Architecture$ and all weights bounded
  (in magnitude) by $R$.
  Let us also define the index set
  \(
    I(\Architecture)
    := \biguplus_{j=1}^{K}
       \big(
         \{ j \}
         \times \{ 1,\CompressedDots,a_j \}
         \times \{ 1,\CompressedDots,1+a_{j-1} \}
       \big) ,
  \)
  noting that $\CalNN(\Architecture) \cong [-R,R]^{I(\Architecture)}$.
  In the following, we will equip $\CalNN(\Architecture)$ with the $\ell^\infty$-norm.
  Then, \mbox{\cite[Theorem~2.6]{BlackScholesGeneralizationError}} shows that the realization map
  $R_\varrho : \CalNN(\Architecture) \to C([0,1]^d), \Phi \mapsto R_\varrho \Phi$
  is Lipschitz continuous on $\CalNN(\Architecture)$, with Lipschitz constant bounded by
  $2 K^2 \, R^{K-1} \, \| \Architecture \|_\infty^{K}$,
  a fact that we will use below.

  For $\ell \in \{ 1,\dots,L \}$, define $\Architecture^{(\ell)} := (d,n,\dots,n,1) \in \N^{\ell+1}$
  and $I_\ell := I(\Architecture^{(\ell)})$, as well as
  \[
    \Sigma_\ell
    := \Big\{
         R_\varrho \Phi
         \,\, \colon
         \begin{array}{l}
           \Phi \text{ NN with }
           \din(\Phi) = d,
           \dout(\Phi) = 1,
           \\
           W(\Phi) \leq n,
           L(\Phi) = \ell,
           \WeightSize{\Phi} \leq R
         \end{array}
       \Big\} .
  \]
  By dropping ``dead neurons,'' it is easy to see that each $f \in \Sigma_\ell$
  is of the form ${f = R_\varrho \Phi}$ for some ${\Phi \in \CalNN(\Architecture^{(\ell)})}$
  satisfying $W(\Phi) \leq n$.
  Thus, keeping the identification ${\CalNN(\Architecture) \cong [-R,R]^{I(\Architecture)}}$,
  given a subset $S \subset I_\ell$, let us write
  $\CalNN_{S,\ell} := \big\{ \Phi \in \CalNN(\Architecture^{(\ell)}) \colon \supp \Phi \subset S \big\}$;
  then we have ${\Sigma_\ell = \bigcup_{S \subset I_\ell, |S| = n} R_\varrho (\CalNN_{S,\ell})}$.
  Moreover, it is easy to see that $|I_\ell| = 2d$ if $\ell = 1$
  while if $\ell \geq 2$ then $|I_\ell| = 1 + n (d+2) + (\ell-2) (n^2 + n)$.
  This implies in all cases that $|I_\ell| \leq 2 n (L n + d)$.

  Now we collect several observations which in combination will imply the claimed bound.
  First, directly from the definition of covering numbers, we see that
  if $\Theta$ is Lipschitz continuous, then
  $\Covering (\Theta(\Omega), \eps) \leq \Covering (\Omega, \frac{\eps}{\mathrm{Lip}(\Theta)})$,
  and furthermore
  ${\Covering (\bigcup_{j=1}^K \Omega_j, \eps) \leq \sum_{j=1}^K \Covering(\Omega_j, \eps)}$.
  Moreover, since $\CalNN_{S,\ell} \cong [-R,R]^{|S|}$, we see by
  \cite[Lemma~2.7]{BlackScholesGeneralizationError} that
  $\Covering_{\ell^\infty}(\CalNN_{S,\ell}, \eps) \leq \lceil R/\eps \rceil^n \leq (2R/\eps)^n$.
  Finally, \cite[Exercise~0.0.5]{VershyninHighDimensionalProbability} provides the bound
  $\binom{N}{n} \leq (e N / n)^n$ for $n \leq N$.

  Recall that the realization map $R_\varrho : \CalNN(\Architecture^{(\ell)}) \to C([0,1]^d)$
  is Lipschitz continuous with ${\mathrm{Lip}(R_{\varrho}) \leq C := 2 L^2 R^{L-1} \max \{ d,n \}^L}$.
  Combining this with the observations from the preceding paragraph and recalling that
  ${|I_\ell| \leq 2 n (L n + d)}$, we see
  \[
    \begin{split}
      \Covering_{C([0,1]^d)}(\Sigma_\ell, \eps)
      & \leq \sum_{S \subset I_\ell, |S| = n}
               \Covering_{C([0,1]^d)} \bigl(R_\varrho (\CalNN_{S,\ell}), \eps\bigr) \\
      & \leq \sum_{S \subset I_\ell, |S| = n}
               \Covering_{\ell^\infty} (\CalNN_{S,\ell}, \tfrac{\eps}{C}) \\
      & \leq \sum_{S \subset I_\ell, |S| = n}
               \Bigl(\frac{2 C R}{\eps}\Bigr)^{|S|}
        =    \binom{|I_\ell|}{n}
             \cdot \Bigl(\frac{2 C R}{\eps}\Bigr)^{n} \\
      & \leq \Bigl( \frac{e |I_\ell|}{n} \Bigr)^n
             \cdot \Bigl(\frac{2 C R}{\eps}\Bigr)^n
      \leq \bigl(2 e (L n + d)\bigr)^n
           \cdot \Bigl(\frac{2 C R}{\eps}\Bigr)^n
      .
    \end{split}
  \]
  Finally, noting that $\NNSigma = \bigcup_{\ell=1}^L \Sigma_\ell$
  and setting ${\eta := \max \{ d,n \}}$, we see via elementary estimates that
  \[
    \Covering_{C([0,1]^d)}(\NNSigma, \eps)
    \leq L
         \cdot \big(
                 4 e (L n + d) R C / \eps
               \big)^n
    \leq L \cdot \bigl(16 e \, L^3 \eta^{L+1} R^L / \eps\bigr)^n
    \leq \bigl(44 \, L^4 \, \eta^{L+1} R^L / \eps\bigr)^n ,
  \]
  which implies the claim of the lemma.
\end{proof}

Using the preceding bounds for the covering numbers of the network sets $\NNSigma$,
we now derive covering number bounds for the (closure of the) unit ball
$\UnitBall[\infty]$ of the approximation space $\ApproxSpace[\infty]$.

\begin{lemma}\label{lem:ApproxSpaceCoveringBounds}
  Let $d \in \N$, $C_1,C_2,\alpha \in (0,\infty)$, and $\theta,\nu \in [0,\infty)$.
  Assume that $\CoeffFunc(n) \leq C_1 \cdot n^\theta$ and $\LayerFunc(n) \leq C_2 \cdot \ln^\nu(2 n)$
  for all $n \in \N$.

  Then there exists $C = C(d,\alpha,\theta,\nu,C_1,C_2) > 0$ such that
  for any $\eps \in (0,1]$, the unit ball
  \[
    \UnitBall[\infty]
    := \bigl\{ f \in \ApproxSpace[\infty]([0,1]^d) \colon \| f \|_{\ApproxSpace[\infty]} \leq 1 \bigr\}
  \]
  satisfies
  \[
    \Covering_{C([0,1]^d)} \big(\, \ClosedUnitBall[\infty], \eps \,\big)
    \leq \exp\bigl(C \cdot \eps^{-1/\alpha} \cdot \ln^{\nu+1}(2/\eps)\bigr) .
  \]
  Here, we denote by $\ClosedUnitBall[\infty]$ the closure of $\UnitBall[\infty]$
  in $C([0,1]^d)$.
\end{lemma}

\begin{proof}
  Let $n := \big\lceil (8/\eps)^{1/\alpha} \big\rceil \in \N_{\geq 2}$,
  noting $n^{-\alpha} \leq \eps / 8$.
  Set $C := \CoeffFunc(n)$ and $L := \LayerFunc(n)$.
  \Cref{lem:NetworkSetCoveringBound} provides an absolute constant $C_3 > 0$ and $N \in \N$
  such that $N \leq \bigl(\frac{C_3}{\eps} \, L^4 \cdot (C \, \max\{ d,n \})^{1+L} \bigr)^n$
  and functions $h_1,\dots,h_N \in \NNSigma$ satisfying
  $\NNSigma \subset \bigcup_{j=1}^N \ClosedBall_{\eps/4} (h_j)$;
  here, $\ClosedBall_\eps (h)$ is the closed ball in $C([0,1]^d)$ of radius $\eps$
  around $h$.
  For each $j \in \FirstN{N}$ choose $g_j \in \UnitBall[\infty] \cap \ClosedBall_{\eps/2} (h_j)$,
  provided that the intersection is non-empty; otherwise choose $g_j := 0 \in \UnitBall[\infty]$.

  We claim that $\UnitBall[\infty] \subset \bigcup_{j=1}^N \ClosedBall_{\eps}(g_j)$.
  To see this, let $f \in \UnitBall[\infty]$ be arbitrary;
  then \Cref{lem:ApproximationSpaceProperties} shows that $\PreNorm[\infty](f) \leq 1$.
  Directly from the definition of $\PreNorm[\infty]$
  we see that we can choose $h \in \NNSigma$ satisfying $n^\alpha \, \| f - h \|_{L^\infty} \leq 2$
  and hence $\| f - h \|_{L^\infty} \leq \frac{\eps}{4}$.
  By choice of $h_1,\dots,h_N$, there exists $j \in \FirstN{N}$
  satisfying $\| h - h_j \|_{L^\infty} \leq \frac{\eps}{4}$.
  This implies $\| f - h_j \|_{L^\infty} \leq \frac{\eps}{2}$ and therefore
  $f \in \ClosedBall_{\eps/2}(h_j) \cap \UnitBall[\infty] \neq \emptyset$.
  By our choice of $g_j$, we thus have $g_j \in \UnitBall[\infty] \cap \ClosedBall_{\eps/2}(h_j)$
  and hence $\| f - g_j \|_{L^\infty} \leq \eps$.
  All in all, we have thus shown $\UnitBall[\infty] \subset \bigcup_{j=1}^N \ClosedBall_{\eps/2}(g_j)$
  and hence also $\ClosedUnitBall[\infty] \subset \bigcup_{j=1}^N \ClosedBall_{\eps/2}(g_j)$.
  This implies $\Covering_{C([0,1]^d)} (\ClosedUnitBall[\infty], \eps) \leq N$,
  so that it remains to estimate $N$ sufficiently well.

  To estimate $N$, first note that
  \begin{equation}
    n
    \leq 1 + (\tfrac{8}{\eps})^{1/\alpha}
    \leq 2 \cdot 8^{1/\alpha} \, \eps^{-1/\alpha}
    \quad \text{ and } \quad
    \ln(n)
    \leq \ln(2 n)
    \leq \ln(4 \cdot 8^{1/\alpha}) + \tfrac{1}{\alpha} \ln(\tfrac{1}{\eps})
    \leq C_4 \cdot \ln(\tfrac{2}{\eps})
    \label{eq:ApproxSpaceCoveringNumerProofNEstimate}
  \end{equation}
  for a suitable constant $C_4 = C_4(\alpha) > 0$.
  This implies
  \[
    L
    \leq 1 + L
    \leq 2 L
    \leq 2 C_2 \, \ln^{\nu} (2 n)
    \leq 2 C_2 C_4^\nu \cdot \ln^{\nu}(\tfrac{2}{\eps})
    \leq C_5 \cdot \ln^\nu(\tfrac{2}{\eps})
  \]
  with a constant $C_5 = C_5(C_2,\nu,\alpha) \geq 1$.

  Now, using \Cref{eq:ApproxSpaceCoveringNumerProofNEstimate} and noting
  $\max \{ d,n \} \leq d \, n$, we obtain ${C_6 = C_6 (d,\alpha,C_1) > 0}$
  and $C_7 = C_7 (d,\alpha,\theta,\nu,C_1,C_2) > 0$ satisfying
  \begin{equation}
    \begin{split}
      \big(
        C \, \max \{ d, n \}
      \big)^{1 + L}
      & \leq \big( C_1 \, d \cdot n^{\theta + 1} \big)^{1 + L}
        \leq \bigl(C_6 \cdot n^{(1+\theta) / \alpha}\bigr)^{1 + L}
        \leq \bigl(C_6 \cdot n^{(1+\theta) / \alpha}\bigr)^{C_5 \, \ln^{\nu}(2/\eps)} \\
      & =    \exp
             \Big(
               \big(
                 \ln(C_6) + \tfrac{1+\theta}{\alpha} \, \ln(n)
               \big)
               \cdot C_5 \, \ln^{\nu}(2/\eps)
             \Big) \\
      & =    \exp
             \Big(
               \big(
                 \ln(C_6) + \tfrac{1+\theta}{\alpha} \, C_4 \, \ln(2/\eps)
               \big)
               \cdot C_5 \, \ln^{\nu}(2/\eps)
             \Big) \\
      & \leq \exp \Big( C_7 \cdot \ln^{\nu + 1}(2 / \eps) \Big)
    \end{split}
    \label{eq:ApproxSpaceCoveringNumerProof1stIngredient}
  \end{equation}
  Furthermore, using the elementary estimate $\ln x \leq x$ for $x > 0$, we see
  \begin{equation}
    \begin{split}
      \frac{C_3}{\eps} \, L^4
      & \leq C_3 C_5^4 \cdot \ln^{4 \nu}(2/\eps) \cdot \eps^{-1}
        \leq 2^{4\nu} C_3 C_5^4 \cdot \eps^{-(1 + 4 \nu)} \\
      & =    \exp \big( C_8 + (1 + 4 \nu) \cdot \ln(1/\eps) \big)
        \leq \exp \big( C_9 \, \ln(2/\eps) \big)
        \leq \exp \bigl(C_{10} \, \ln^{\nu+1}(2/\eps)\bigr)
    \end{split}
    \label{eq:ApproxSpaceCoveringNumerProof2ndIngredient}
  \end{equation}
  for suitable constants $C_8,C_9,C_10$ all only depending on $\nu,\alpha,C_2$.

  Overall, recalling the estimate for $N$ from the beginning of the proof and using
  \Cref{eq:ApproxSpaceCoveringNumerProofNEstimate,%
  eq:ApproxSpaceCoveringNumerProof1stIngredient,eq:ApproxSpaceCoveringNumerProof2ndIngredient},
  we finally see
  \begin{align*}
    N
    & \leq \Big(
             \frac{C_3}{\eps} \, L^4
             \cdot \big(
                     C \, \max \{ d,n \}
                   \big)^{1+L}
           \Big)^n
      \leq \exp
           \Big(
             (C_{10} + C_7) \cdot n \cdot \ln_{\nu+1} (2/\eps)
           \Big) \\
    & \leq \exp \Big(
                  2 \cdot 8^{1/\alpha}
                  \cdot (C_{10} + C_7)
                  \cdot \eps^{-1/\alpha}
                  \cdot \ln^{\nu+1}(2/\eps)
                \Big)
      ,
  \end{align*}
  which easily implies the claim of the lemma.
\end{proof}

Combining the preceding covering number bounds with bounds from statistical learning theory,
we now prove the following error bound for approximating functions $f \in \ApproxSpace[\infty]$
from point samples, with error measured in $L^2$.

\begin{theorem}\label{thm:L2ErrorBound}
  Let $d \in \N$, $C_1,C_2,\alpha \in (0,\infty)$, and $\theta,\nu \in [0,\infty)$.
  Let $\LayerFunc,\CoeffFunc : \N \to \N$ be non-decreasing and such that
  $\CoeffFunc(n) \leq C_1 \cdot n^\theta$ and $\LayerFunc(n) \leq C_2 \cdot \ln^\nu(2 n)$
  for all $n \in \N$.
  Let $\UnitBall[\infty]$ as in \Cref{eq:UnitBallDefinition}, and denote by
  $\ClosedUnitBall[\infty]$ the closure of $\UnitBall[\infty]$ in $C([0,1]^d)$.

  Then there exists a constant $C = C(\alpha,\theta,\nu,d,C_1,C_2) > 0$ such that for each $m \in \N$,
  there are points $x_1,\dots,x_m \in [0,1]^d$ with the following property:
  \begin{equation}
    \forall \, f,g \in \ClosedUnitBall[\infty]
            \text{ with } f(x_i) = g(x_i) \text{ for all } i \in \FirstN{m}: \quad
    \| f - g \|_{L^2([0,1]^d)}
    \leq C \cdot
    \big(
      \ln^{1+\nu}(2 m) \big/ m
    \big)^{\frac{\alpha/2}{1+\alpha}} .
    \label{eq:L2LearningBoundExplicit}
  \end{equation}
  In particular, this implies for the embedding
  $\iota_2 : \ApproxSpace[\infty]([0,1]^d) \hookrightarrow L^2([0,1]^d)$ that
  \[
    \beta_{\ast}^{\deterministic} \big(\, \ClosedUnitBall[\infty], \iota_2 \,\big)
    \geq \frac{\alpha/2}{1 + \alpha} .
  \]
\end{theorem}

\begin{rem*}
  The proof shows that the points $x_1,\dots,x_m$ can be obtained with positive probability
  by uniformly and independently sampling $x_1,\dots,x_m$ from $[0,1]^d$.
  In fact, an inspection of the proof shows for each $m \in \N$ that this sampling procedure
  will result in ``good'' points with probability at least
  \[
    1 - \exp
        \Big(
          - \big[
              m \cdot \ln^{\alpha \cdot (1+\nu)}(2m)
            \big]^{1/(1+\alpha)}
        \Big) .
  \]
\end{rem*}

\begin{proof}
  \textbf{Step~1:}
  An essential ingredient for our proof is
  \cite[Proposition~7]{CuckerSmaleMathematicalFoundationsOfLearning}.
  In this step, we briefly recall the general setup from
  \cite{CuckerSmaleMathematicalFoundationsOfLearning} and describe how it applies to our setting.

  Let us fix a function $f_0 \in \ClosedUnitBall[\infty]$ for the moment.
  In \cite{CuckerSmaleMathematicalFoundationsOfLearning}, one starts with a probability measure
  $\rho$ on $Z = X \times Y$, where $X$ is a compact domain and $Y = \R$.
  In our case we take $X = [0,1]^d$ and we define
  $\rho(M) := \rho_{f_0}(M) := \LebesgueMeasure(\{ x \in [0,1]^d \colon (x,f_0(x)) \in M \})$
  for any Borel set $M \subset X \times Y$.
  In other words, $\rho$ is the distribution of the random variable $\xi = (\eta, f_0(\eta))$,
  where $\eta$ is uniformly distributed in $X = [0,1]^d$.
  Then, in the notation of \cite{CuckerSmaleMathematicalFoundationsOfLearning}, the measure
  $\rho_X$ on $X$ is simply the Lebesgue measure on $[0,1]^d$ and the conditional probability measure
  $\rho(\bullet \mid x)$ on $Y$ is $\rho(\bullet \mid x) = \delta_{f_0(x)}$.
  Furthermore, the \emph{regression function} $f_\rho$
  considered in \cite{CuckerSmaleMathematicalFoundationsOfLearning} is simply $f_\rho = f_0$,
  and the \emph{(least squares) error} $\CalE(f)$ of $f : X \to Y$ is
  $\CalE(f) = \int_{[0,1]^d} |f(x) - f_0(x)|^2 \, d\LebesgueMeasure(x) = \| f - f_0 \|_{L^2}^2$;
  to emphasize the role of $f_0$, we shall write $\CalE(f; f_0) = \| f - f_0 \|_{L^2}^2$ instead.
  The \emph{empirical error} of $f : X \to Y$ with respect to a sample $\z \in Z^m$ is
  \[
    \CalE_{\z} (f) := \frac{1}{m} \sum_{i=1}^m \big( f(x_i) - y_i \big)^2
    \quad \text{where} \quad
    \z = \bigl( (x_1,y_1), \dots, (x_m,y_m)\bigr) .
  \]
  We shall also use the notation
  \[
    \CalE_{\x} (f; f_0)
    := \CalE_{\z} (f)
    = \frac{1}{m}
      \sum_{i=1}^m
        \big(
          f(x_i) - f_0(x_i)
        \big)^2
    \quad \text{where} \quad y_i = f_0(x_i) \text{ for } i \in \FirstN{m}.
  \]

  Furthermore, as the \emph{hypothesis space} $\CalH$ we choose $\CalH := \ClosedUnitBall[\infty]$.
  As required in \cite{CuckerSmaleMathematicalFoundationsOfLearning}, this is a compact subset
  of $C(X)$; indeed $\ClosedUnitBall[\infty] \subset C([0,1]^d)$ is closed
  and has finite covering numbers $\Covering (\ClosedUnitBall[\infty],\eps)$
  for arbitrarily small $\eps > 0$ (see \Cref{lem:ApproxSpaceCoveringBounds}).
  Thus, $\ClosedUnitBall[\infty] \subset C([0,1]^d)$ is compact;
  see for instance \cite[Theorem~3.28]{AliprantisBorderHitchhiker}.

  Moreover, since every $(x,y) \in Z$ satisfies $y = f_0(x)$ almost surely
  (with respect to $\rho = \rho_{f_0}$), and since all $f \in \CalH = \ClosedUnitBall[\infty]$
  satisfy $\| f \|_{C([0,1]^d)} \leq 1$, we see that $\rho_{f_0}$-almost surely, the estimate
  $|f(x) - y| = |f(x) - f_0(x)| \leq 2 =: M$ holds for all $f \in \CalH$.
  Furthermore, in \cite{CuckerSmaleMathematicalFoundationsOfLearning},
  the function $f_{\CalH} \in \CalH$ is a minimizer of $\CalE$ over $\CalH$;
  in our case, since $f_0 \in \CalH$, we easily see that $f_{\CalH} = f_0$ and $\CalE(f_{\CalH}) = 0$.
  Therefore, the \emph{error in $\CalH$} of $f \in \CalH$
  as considered in \cite{CuckerSmaleMathematicalFoundationsOfLearning} is simply
  $\CalE_{\CalH}(f) = \CalE(f) - \CalE(f_{\CalH}) = \CalE(f)$.
  Finally, the \emph{empirical error in $\CalH$} of $f \in \CalH$ is given by
  $\CalE_{\CalH,\z}(f) = \CalE_{\z}(f) - \CalE_{\z}(f_{\CalH})$.
  Hence, if $\z = \bigl( (x_1,y_1),\dots,(x_m,y_m)\bigr)$ satisfies
  $y_i = f_0(x_i)$ for all $i \in \FirstN{m}$,
  then $\CalE_{\CalH,\z}(f) = \CalE_{\z}(f) = \CalE_{\x}(f; f_0)$, because of $f_{\CalH} = f_0$.

  Now, let $\x = (x_1,\dots,x_m)$ be i.i.d.~uniformly distributed in $[0,1]^d$
  and set $y_i = f_0(x_i)$ for $i \in \FirstN{m}$ and
  $\z = (z_1,\dots,z_m) = \bigl( (x_1,y_1),\dots,(x_m,y_m)\bigr)$.
  Then $z_1,\dots,z_m \overset{iid}{\sim} \rho_{f_0}$.
  Therefore, \mbox{\cite[Proposition~7]{CuckerSmaleMathematicalFoundationsOfLearning}}
  (applied with $\alpha = \frac{1}{6}$) shows for arbitrary $\eps > 0$ and $m \in \N$
  that there is a measurable set
  \begin{equation}
    \begin{split}
      & E = E(m,\eps,f_0) \subset ([0,1]^d)^m \cong [0,1]^{d m} \\
        \quad \text{with} \quad
      & \LebesgueMeasure (E)
        \leq \Covering\bigl(\ClosedUnitBall[\infty], \tfrac{\eps}{48}\bigr)
             \cdot e^{-m \eps / 288} \\[0.1cm]
      \text{satisfying} \quad
      & \sup_{f \in \CalH}
          \frac{\CalE(f;f_0) - \CalE_{\x} (f;f_0)}{\CalE(f;f_0) + \eps}
        =    \sup_{f \in \CalH}
               \frac{\CalE_{\CalH}(f) - \CalE_{\CalH,\z} (f)}{\CalE_{\CalH}(f) + \eps}
        \leq \frac{1}{2}
        \qquad \forall \, \x \in \bigl([0,1]^d\bigr)^m \setminus E .
    \end{split}
    \label{eq:CuckerSmaleApplicatiom}
  \end{equation}
  Here, we remark that \cite[Proposition~7]{CuckerSmaleMathematicalFoundationsOfLearning}
  requires the hypothesis space $\CalH$ to be convex, which is not in general satisfied in our case.
  However, as shown in \cite[Remark~13]{CuckerSmaleMathematicalFoundationsOfLearning},
  the assumption of convexity can be dropped provided that $f_\rho \in \CalH$,
  which is satisfied in our case.

  \medskip{}

  \noindent
  \textbf{Step~2:} In this step, we prove the first claim of the theorem.
  To this end, we first apply \Cref{lem:ApproxSpaceCoveringBounds} to obtain a constant
  $C_3 = C_3(\alpha,\nu,\theta,d,C_1,C_2) > 0$ satisfying
  \begin{equation}
    \Covering\bigl(\, \ClosedUnitBall[\infty], \eps \bigr)
    \leq N_\eps
    :=   \Covering\bigl(\, \ClosedUnitBall[\infty], \tfrac{\eps}{48} \bigr)
    \leq \exp \big( C_3 \cdot \eps^{-1/\alpha} \cdot \ln^{1+\nu} (2/\eps) \big)
    \qquad \forall \, \eps \in (0,1] .
    \label{eq:L2ErrorBoundCoveringNumberApplication}
  \end{equation}
  Next, define $C_4 := 1 + \frac{\alpha}{1 + \alpha}$ and $C_5 := C_4^{1+\nu}$,
  and choose $C_6 = C_6(\alpha,\nu,\theta,d,C_1,C_2) \geq 1$ such that
  $2 C_3 C_5 - \frac{C_6}{288} \leq -1 < 0$.

  Let $m \in \N$ be arbitrary with $m \geq m_0 = m_0(\alpha,\nu,\theta,d,C_1,C_2) \geq 2$,
  where $m_0$ is chosen such that
  $\eps := C_6 \cdot \big( \ln^{1+\nu}(2 m) \big/ m \big)^{\alpha / (1+\alpha)}$
  satisfies $\eps \in (0,1]$; the case $m \leq m_0$ will be considered below.
  Let $N := N_\eps$ as in \Cref{eq:L2ErrorBoundCoveringNumberApplication}.
  Since $\Covering(\ClosedUnitBall[\infty], \eps) \leq N$, we can choose
  $f_1,\dots,f_N \in \ClosedUnitBall[\infty]$ such that
  $\ClosedUnitBall[\infty] \subset \bigcup_{j=1}^N \overline{B}_\eps (f_j)$, where
  $\overline{B}_\eps (f) := \bigl\{ g \in C([0,1]^d) \colon \| f - g \|_{L^\infty} \leq \eps \bigr\}$.
  Now, for each $j \in \FirstN{N}$, choose $E_j := E(m,\eps,f_j) \subset ([0,1]^d)^m$
  as in \Cref{eq:CuckerSmaleApplicatiom}, and define $E^\ast := \bigcup_{j=1}^N E_j$.

  Note because of $C_6 \geq 1$ and $\ln(2m) \geq \ln(4) \geq 1$ that
  $\eps \geq \big( \ln^{1+\nu}(2 m) \big/ m \big)^{\alpha/(1+\alpha)} \geq m^{-\alpha/(1+\alpha)}$
  and hence
  \[
    \ln(2/\eps)
    \leq \ln(2) + \tfrac{\alpha}{1+\alpha} \ln(m)
    \leq C_4 \, \ln(2m)
    \quad \text{and thus} \quad
    \ln^{1+\nu}(2/\eps) \leq C_5 \, \ln^{1+\nu}(2 m) .
  \]
  Using the estimate for $N = N_\eps$ from \Cref{eq:L2ErrorBoundCoveringNumberApplication}
  and the bound for the measure of $E_j$ from \Cref{eq:CuckerSmaleApplicatiom}, we thus see
  \begin{align*}
    \LebesgueMeasure(E^\ast)
    & \leq N
           \cdot \Covering\bigl(\, \ClosedUnitBall[\infty], \tfrac{\eps}{48} \bigr)
           \cdot e^{- m \eps / 288}
      \leq \exp
           \big(
             2 C_3 \cdot \eps^{-1/\alpha} \cdot \ln^{1+\nu}(2/\eps)
             - m \eps / 288
           \big) \\
    & \leq \exp
           \Big(
             2 C_3 C_5
             \cdot \big( m \big/ \ln^{1+\nu}(2m) \big)^{1/(1+\alpha)}
             \cdot \ln^{1+\nu}(2 m)
             - \tfrac{C_6}{288}
               \cdot m^{1 - \frac{\alpha}{1+\alpha}}
               \cdot \bigl(\ln(2m)\bigr)^{(1+\nu) \frac{\alpha}{1+\alpha}}
           \Big) \\
    & \leq \exp
           \Big(
             m^{\frac{1}{1+\alpha}}
             \cdot \big( \ln(2m) \big)^{(1+\nu)\frac{\alpha}{1+\alpha}}
             \cdot \big( 2 C_3 C_5 - \tfrac{C_6}{288} \big)
           \Big) \\
    & \leq \exp
           \Big(
             - m^{\frac{1}{1+\alpha}}
               \cdot \big( \ln(2m) \big)^{(1+\nu)\frac{\alpha}{1+\alpha}}
           \Big)
      <    1 .
  \end{align*}
  Thus, we can choose $\x = (x_1,\dots,x_m) \in ([0,1]^d)^m \setminus E^\ast$.
  We claim that every such choice satisfies the property stated in the first part of the theorem.

  \smallskip{}

  To see this, let $f,g \in \ClosedUnitBall[\infty]$ be arbitrary with $f(x_i) = g(x_i)$
  for all $i \in \FirstN{m}$.
  By choice of $f_1,\dots,f_N$, there exists some $j \in \FirstN{N}$ satisfying
  $\| f - f_j \|_{L^\infty} \leq \eps$.
  Since $\x \notin E^\ast$, we have $\x \notin E_j = E(m,\eps,f_j)$.
  In view of \Cref{eq:CuckerSmaleApplicatiom},
  this implies $\CalE(g;f_j) - \CalE_{\x}(g;f_j) \leq \frac{1}{2} (\CalE(g;f_j) + \eps)$,
  and after rearranging, this yields $\CalE(g;f_j) \leq 2 \, \CalE_{\x}(g;f_j) + \eps$.
  Because of $\| g - f_j \|_{L^2} \leq \| g \|_{L^\infty} + \| f_j \|_{L^\infty} \leq 2$
  and thanks to the elementary estimate $(a + \eps)^2 = a^2 + 2 a \eps + \eps^2 \leq a^2 + 5 \eps$
  for $0 \leq a \leq 2$, we thus see
  \[
    \| g - f \|_{L^2}^2
    \leq \big( \| g - f_j \|_{L^2} + \| f_j - f \|_{L^2} \big)^2
    \leq \| g - f_j \|_{L^2}^2 + 5 \eps
    =    \CalE(g; f_j) + 5 \eps
    \leq 2 \, \CalE_{\x}(g;f_j) + 6 \eps .
  \]
  But directly from the definition and because of $g(x_i) = f(x_i)$
  and $\| f - f_j \|_{L^\infty} \leq \eps$, we see
  \({
    \CalE_{\x}(g;f_j)
    = \frac{1}{m}
      \sum_{i=1}^m
        \bigl( g(x_i) - f_j(x_i) \bigr)^2
    \leq \eps^2
    \leq \eps
    .
  }\)
  Overall, we thus see that
  \[
    \| g - f \|_{L^2}^2
    \leq 8 \eps
    =    8 C_6 \,
         \big( \ln^{1+\nu}(2m) \big/ m \big)^{\frac{\alpha}{1+\alpha}}
    \quad \forall \, f,g \in \ClosedUnitBall[\infty]
                     \text{ satisfying } f(x_i) = g(x_i)
                     \text{ for all } i \in \FirstN{m}.
  \]
  We have thus proved the claim for $m \geq m_0$.
  Since $\| g - f \|_{L^2} \leq \| f \|_{L^\infty} + \| g \|_{L^\infty} \leq 2$
  for \emph{arbitrary} $f,g \in \ClosedUnitBall[\infty]$,
  it is easy to see that this proves the claim for all $m \in \N$,
  possibly after enlarging $C$.

  \medskip{}

  \noindent
  \textbf{Step~3:}
  To complete the proof of the theorem, for each $\y = (y_1,\dots,y_m) \in \R^m$,
  choose a fixed $f_{\y} \in \ClosedUnitBall[\infty]$ satisfying
  \[
    f_{\y}
    \in \argmin_{f \in \ClosedUnitBall[\infty]}
          \sum_{i=1}^m \bigl(f(x_i) - y_i\bigr)^2 ;
  \]
  existence of $f_{\y}$ is an easy consequence
  of the compactness of $\ClosedUnitBall[\infty] \subset C([0,1]^d)$.
  Define
  \[
    \Phi : \quad
    \R^m \to \ClosedUnitBall[\infty], \quad
    \y \mapsto f_{\y}
    \qquad \text{and} \qquad
    A : \quad
    \ClosedUnitBall[\infty] \to \ClosedUnitBall[\infty], \quad
    f \mapsto \Phi\bigl( (f(x_1),\dots,f(x_m)) \bigr) .
  \]
  Then given any $f \in \ClosedUnitBall[\infty]$, the function $g := A f \in \ClosedUnitBall[\infty]$
  satisfies $f(x_i) = g(x_i)$ for all $i \in \FirstN{m}$, and hence
  $\| f - A f \|_{L^2} \leq C \cdot \bigl(\ln^{1+\nu}(2 m) \big/ m\bigr)^{\frac{\alpha/2}{1 + \alpha}}$,
  as shown in the previous step.
  By definition of $\beta_\ast^{\deterministic}(U,\iota_2)$, this easily entails
  $\beta_\ast^{\deterministic}(U,\iota_2) \geq \frac{\alpha/2}{1 + \alpha}$.
\end{proof}


\section{Hardness of approximation in \texorpdfstring{$L^2$}{L²}}%
\label{sec:L2ApproximationHardness}


This section presents hardness results for approximating the embedding
$\ApproxSpace[\infty]([0,1]^d) \hookrightarrow L^2([0,1]^d)$ using point samples.

\begin{theorem}\label{thm:L2HardnessResult}
  Let $\LayerFunc,\CoeffFunc : \N \to \N \cup \{ \infty \}$ be non-decreasing with
  $\LayerFunc^\ast \geq 2$.
  Let $d \in \N$ and $\alpha \in (0,\infty)$
  Set $\gamma^\flat := \gamma^{\flat}(\LayerFunc,\CoeffFunc)$ as in \Cref{eq:GammaDefinition}
  and let $\UnitBall[\infty]$ as in \Cref{eq:UnitBallDefinition}.
  For the embedding $\iota_2 : \UnitBall[\infty] \to L^2([0,1]^d), f \mapsto f$,
  we then have
  \begin{equation}
    \begin{split}
      \beta_\ast^{\deterministic} \bigl(\UnitBall[\infty], \iota_2\bigr) , \,\,
      \beta_\ast^{\MonteCarlo} \bigl(\UnitBall[\infty], \iota_2\bigr)
      & \leq \begin{cases}
               \min
               \big\{
                 \frac{1}{2} + \frac{\alpha}{\alpha + \gamma^{\flat}}, \,\,
                 \frac{2 \alpha}{\alpha + \gamma^{\flat}}
               \big\},
               & \text{if } \alpha + \gamma^{\flat} < 2, \\
               \min
               \big\{
                 \frac{1}{2} + \frac{\alpha}{\alpha + \gamma^{\flat}}, \,\,
                 \alpha, \,\,
                 \frac{1}{2} + \frac{\alpha - \frac{1}{2}}{\alpha + \gamma^{\flat} - 1}
               \big\},
               & \text{if } \alpha + \gamma^{\flat} \geq 2
             \end{cases} \\
      & =    \begin{cases}
               \frac{2 \alpha}{\alpha + \gamma^{\flat}} ,
               & \text{if } \alpha + \gamma^{\flat} < 2, \\
               \alpha ,
               & \text{if } \alpha + \gamma^{\flat} \geq 2
                 \text{ and } \alpha \leq \frac{1}{2}, \\
               \frac{1}{2} + \frac{\alpha - \frac{1}{2}}{\alpha + \gamma^{\flat} - 1},
               & \text{if } \alpha + \gamma^{\flat} \geq 2
                 \text{ and } \frac{1}{2} \leq \alpha \leq \gamma^{\flat}, \\
               \frac{1}{2} + \frac{\alpha}{\alpha + \gamma^{\flat}},
               & \text{if } \alpha + \gamma^{\flat} \geq 2
                 \text{ and } \alpha \geq \gamma^{\flat}.
             \end{cases}
    \end{split}
    \label{eq:L2Hardness}
  \end{equation}
\end{theorem}

\begin{rem*}
  The bound from above might seem intimidating at first sight,
  so we point out two important consequences:
  First, we always have
  \(
    \beta_\ast^{\deterministic} \bigl(\UnitBall[\infty], \iota_2\bigr) , \,\,
    \beta_\ast^{\MonteCarlo} \bigl(\UnitBall[\infty], \iota_2\bigr)
    \leq \frac{1}{2} + \frac{\alpha}{\alpha + \gamma^{\flat}}
    \leq \frac{3}{2} ,
  \)
  which shows that \emph{no matter how large the approximation rate $\alpha$ is},
  one can never get a better convergence rate than $m^{-3/2}$.
  Furthermore, in the important case where $\gamma^{\flat} = \infty$
  (for instance if the depth-growth function $\LayerFunc$ is unbounded), then
  \(
    \beta_\ast^{\deterministic} \bigl(\UnitBall[\infty], \iota_2\bigr) , \,\,
    \beta_\ast^{\MonteCarlo} \bigl(\UnitBall[\infty], \iota_2\bigr)
    \leq \frac{1}{2} + \frac{\alpha}{\alpha + \gamma^{\flat}}
    =    \frac{1}{2} .
  \)
  These two bounds are the interesting bounds for the regime of large $\alpha$.

  For small $\alpha > 0$, the theorem shows
  \[
    \beta_\ast^{\deterministic} \bigl(\UnitBall[\infty], \iota_2\bigr) , \,\,
    \beta_\ast^{\MonteCarlo} \bigl(\UnitBall[\infty], \iota_2\bigr)
    \leq \max \big\{ \tfrac{2\alpha}{\alpha + \gamma^{\flat}}, \alpha \big\}
    \leq \max \bigl\{ \tfrac{2}{\gamma^{\flat}}, 1 \bigr\} \cdot \alpha
    \leq 2 \alpha ,
  \]
  since $\gamma^{\flat} \geq 1$.
  This shows that one can not get a good rate of approximation
  for small exponents $\alpha > 0$.
\end{rem*}

\begin{proof}
  \textbf{Step~1 (preparation):}
  Let $0 < \gamma < \gamma^{\flat}$ be arbitrary and let
  $\theta \in (0,\infty)$ and $\lambda \in [0,1]$ with $\theta \lambda \leq 1$ and set
  $\omega := \min \{ -\theta \alpha , \,\, \theta \cdot (\gamma - \lambda) - 1 \} \in (-\infty,0)$.

  Let $m \in \N$ be arbitrary and set $M := 4 m$ and $z_j := \frac{1}{4 m} + \frac{j-1}{2 m}$
  for $j \in \FirstN{2 m}$.
  Then, \Cref{lem:HatSumsInUnitBall} yields a constant
  $\kappa = \kappa(\gamma,\alpha,\lambda,\theta,\LayerFunc,\CoeffFunc) > 0$ (independent of $m$)
  such that
  \[
    f_{\boldnu,J}
    := \kappa \cdot m^\omega \cdot \sum_{j \in J} \nu_j \, \Lambda_{M,z_j}^\ast
    \in \UnitBall[\infty]
    \quad \forall \, J \subset \FirstN{2m} \text{ with } |J| \leq 2 \cdot m^{\theta \lambda}
                  \text{ and } {\boldnu = (\nu_j)_{j \in \FirstN{2m}} \in [-1,1]^{2 m}}
    .
  \]
  Furthermore, \Cref{lem:HatSumsInUnitBall} shows that the functions
  $(\Lambda_{M,z_i}^\ast)_{i \in \FirstN{2 m}}$ have disjoint supports contained in $[0,1]^d$
  which are pairwise disjoint (up to null-sets).
  By continuity, this implies $\Lambda_{M,z_i}^\ast \Lambda_{M,z_\ell}^\ast \equiv 0$
  for $i \neq \ell$.

  Let $k := \lceil m^{\theta \lambda} \rceil$, noting because of $\theta \lambda \leq 1$
  that $k \leq \lceil m \rceil = m$ and $k \leq 1 + m^{\theta \lambda} \leq 2 \cdot m^{\theta \lambda}$.
  Set $\SpecialSet := \bigl\{ J \subset \FirstN{2m} \colon |J| = k \bigr\}$
  and $\Gamma_m := \{ \pm 1 \}^{2 m} \times \SpecialSet$.
  The idea of the proof is to show that \Cref{lem:MonteCarloHardnessThroughAverageCase}
  is applicable to the family $(f_{\boldnu,J})_{(\boldnu,J) \in \Gamma_m}$.

  \medskip{}

  \textbf{Step~2:}
  In this step, we prove
  \begin{equation}
    \avsum_{(\boldnu,J) \in \Gamma_m}
      \big\| f_{\boldnu,J} - A(f_{\boldnu,J}) \big\|_{L^2([0,1]^d)}
    \geq \frac{\kappa}{32} \cdot m^{\omega + \frac{1}{2}(\theta \lambda - 1)}
    \qquad \forall \, A \in \Alg_m \bigl(U, L^2([0,1]^d)\bigr) .
    \label{eq:L2HardnessMainEstimate}
  \end{equation}
  To see this, let $\x = (x_1,\dots,x_m) \in ([0,1]^d)^{m}$
  and $Q : \R^m \to L^2([0,1]^d)$ be arbitrary.
  Define
  \({
    I
    := I_{\x}
    := \big\{
         i \in \FirstN{2 m}
         \,\,\colon\,\,
         \forall \, n \in \FirstN{m} : \Lambda_{M,z_i}^\ast (x_n) = 0
       \big\}
  }\)
  as in \Cref{lem:AvoidingSamplingPoints} and recall the estimate $|I| \geq m$ from that lemma.

  Now, given $\boldnu^{(1)} \in \{ \pm 1 \}^{I}$ and $\boldnu^{(2)} \in \{ \pm 1 \}^{I^c}$
  as well as $J \in \SpecialSet$, define
  \[
    F_{\boldnu^{(1)}, J}
    := \kappa
       \cdot m^\omega
       \cdot \sum_{j \in I \cap J}
               \nu_j^{(1)} \Lambda_{M,z_j}^\ast
    \qquad \text{and} \qquad
    g_{\boldnu^{(2)}, J}
    := \kappa
       \cdot m^\omega
       \cdot \sum_{j \in I^c \cap J}
               \nu_j^{(2)} \Lambda_{M,z_j}^\ast
  \]
  and finally
  \(
    h_{\boldnu^{(2)}, J}
    := g_{\boldnu^{(2)}, J}
       - Q
         \bigl(
           g_{\boldnu^{(2)}, J} (x_1),
           \dots,
           g_{\boldnu^{(2)}, J} (x_m)
         \bigr)
    .
  \)
  Note by choice of $I = I_{\x}$ that $f_{\boldnu,J} (x_n) = g_{\boldnu^{(2)},J}(x_n)$
  for all $n \in \FirstN{m}$, if we identify $\boldnu$ with $(\boldnu^{(1)}, \boldnu^{(2)})$,
  as we will continue to do for the remainder of the proof.
  Thus, we see for fixed but arbitrary $\boldnu^{(2)} \in \{ \pm 1 \}^{I^c}$
  and $J \in \SpecialSet$ that
  \begin{equation}
    \begin{split}
      & \sum_{\boldnu^{(1)} \in \{ \pm 1 \}^I}
          \big\|
            f_{\boldnu,J} - Q \big( f_{\boldnu,J}(x_1), \dots, f_{\boldnu,J}(x_m) \big)
          \big\|_{L^2([0,1]^d)} \\
      & = \sum_{\boldnu^{(1)} \in \{ \pm 1 \}^I}
            \big\|
              F_{\boldnu^{(1)},J} + h_{\boldnu^{(2)}, J}
            \big\|_{L^2([0,1]^d)} \\
      & = \frac{1}{2}
          \sum_{\boldnu^{(1)} \in \{ \pm 1 \}^I}
            \Big(
              \big\|
                F_{\boldnu^{(1)},J} + h_{\boldnu^{(2)}, J}
              \big\|_{L^2([0,1]^d)}
              + \big\|
                  F_{-\boldnu^{(1)},J} + h_{\boldnu^{(2)}, J}
                \big\|_{L^2([0,1]^d)}
            \Big) \\
      & \overset{(\ast)}{\geq}
          \sum_{\boldnu^{(1)} \in \{ \pm 1 \}^I}
            \| F_{\boldnu^{(1)},J} \|_{L^2([0,1]^d)} \\
      & \overset{(\blacklozenge)}{\geq}
          2^{|I|}
          \cdot \frac{\kappa}{8}
          \cdot m^\omega
          \cdot \bigg( \frac{|I \cap J|}{m} \bigg)^{1/2} .
    \end{split}
    \label{eq:L2HardnessStep1}
  \end{equation}
  Here, the step marked with $(\ast)$ used the identity
  $F_{-\boldnu^{(1)}, J} = - F_{\boldnu^{(1)}, J}$ and the elementary estimate
  \(
    \| f + g \|_{L^2} + \| - f + g \|_{L^2}
    = \| f + g \|_{L^2} + \| f - g \|_{L^2}
    \geq \| f + g + f - g \|_{L^2}
    =  2 \, \| f \|_{L^2} .
  \)
  Finally, the step marked with $(\ast)$ used that the functions
  $\bigl(\Lambda_{M,z_i}^\ast\bigr)_{i \in \FirstN{2m}}$ have disjoint supports
  (up to null-sets) contained in $[0,1]^d$ and that $\Lambda_{M,z_j}^\ast (x) \geq \frac{1}{2}$
  for all $x \in [0,1]^d$ satisfying $|x_1 - z_j| \leq \frac{1}{2 M}$;
  since $M = 4 m$, this easily implies
  \(
    \| \Lambda_{M,z_i}^{\ast} \|_{L^2([0,1]^d)}
    \geq \frac{1}{2}
         \big(
           \frac{1}{2 M}
         \big)^{1/2}
    \geq \frac{m^{-1/2}}{8}
  \)
  and hence
  \begin{align*}
    \| F_{\boldnu^{(1)}, J} \|_{L^2([0,1]^d)}
    & = \kappa \cdot m^\omega
        \cdot \Big\|
                \sum_{j \in I \cap J}
                  \nu_j^{(1)} \, \Lambda_{M,z_j}^\ast
              \Big\|^{L^2([0,1]^d)} \\
    & = \kappa \cdot m^\omega
        \cdot \Big(
                |\nu_j^{(1)}|^2 \, \| \Lambda_{M,z_j}^\ast \|_{L^2([0,1]^d)}^2
              \Big)^{1/2}
      \geq \frac{\kappa}{8} \cdot m^\omega \cdot \Big( |I \cap J| \,\Big/\, m \Big)^{1/2} .
  \end{align*}

  Combining \Cref{eq:L2HardnessStep1} with \Cref{lem:RandomSubsetExpectationBound}
  and recalling that $k \geq m^{\theta \lambda}$, we finally see
  \[
    \begin{split}
      & \avsum_{(\boldnu,J) \in \Gamma_m}
          \big\|
            f_{\boldnu,J} - Q \big( f_{\boldnu,J}(x_1), \dots, f_{\boldnu,J}(x_m) \big)
          \big\|_{L^2([0,1]^d)} \\
      & \geq \avsum_{J \in \SpecialSet} \,\,\,
               \avsum_{\boldnu^{(2)} \in \{ \pm 1 \}^{I^c}} \,\,
                 \avsum_{\boldnu^{(1)} \in \{ \pm 1 \}^{I}}
                   \big\|
                     f_{\boldnu,J} - Q \big( f_{\boldnu,J}(x_1), \dots, f_{\boldnu,J}(x_m) \big)
                   \big\|_{L^2([0,1]^d)} \\
      & \geq \frac{\kappa}{8}
             \cdot m^\omega
             \avsum_{J \in \SpecialSet}
               \bigg( \frac{|I_{\x} \cap J|}{m} \bigg)^{1/2}
        \geq \frac{\kappa}{32} \cdot m^{\omega + \frac{1}{2}(\theta \lambda - 1)} .
    \end{split}
  \]
  Recall that this holds for any $m \in \N$, arbitrary $\x = (x_1,\dots,x_m) \in ([0,1]^d)^m$
  and any map $Q : \R^m \to L^2([0,1]^d)$.
  Thus, we have established \Cref{eq:L2HardnessMainEstimate}.


  \medskip{}

  \textbf{Step~3:}
  In view of \Cref{eq:L2HardnessMainEstimate},
  an application of \Cref{lem:MonteCarloHardnessThroughAverageCase} shows that
  \begin{equation}
    \beta_\ast^{\deterministic}(U, \iota_2), \beta_\ast^{\MonteCarlo}(U, \iota_2)
    \leq \tfrac{1}{2} - \omega - \tfrac{\theta \lambda}{2}
    =    \tfrac{1}{2}
         + \max
           \big\{
             \theta \cdot (\alpha - \tfrac{\lambda}{2}), \,\,
             1 + \theta \cdot (\tfrac{\lambda}{2} - \gamma)
           \big\}
    \label{eq:L2HardnessAlmostDone}
  \end{equation}
  for arbitrary $0 < \gamma < \gamma^\flat$, $\theta \in (0,\infty)$
  and $\lambda \in [0,1]$ with $\theta \lambda \leq 1$;
  here, we note that $\frac{1}{2} - \frac{\theta \lambda}{2} \geq 0$ and $-\omega \geq 0$.

  From \Cref{eq:L2HardnessAlmostDone}, it is easy (but slightly tedious) to deduce the first line of
  \Cref{eq:L2Hardness}; the details are given in \Cref{lem:OptimizationLemma1}.
  Finally, the second line of \Cref{eq:L2Hardness} follows by a straightforward case distinction.
\end{proof}


\section{Error bounds for numerical integration}%
\label{sec:IntegrationErrorBounds}


In this section, we derive error bounds for the numerical integration
of functions ${f \in \ApproxSpace[\infty]([0,1]^d)}$ based on point samples.
We first consider (in \Cref{thm:DeterministicIntegrationErrorBound}) deterministic algorithms,
which surprisingly provide a strictly positive rate of convergence,
even for neural network approximation spaces
\emph{without restrictions on the size of the network weights}.
Then, in \Cref{thm:MonteCarloIntegrationErrorBound}, we consider the case of Monte Carlo algorithms.
As usual for such algorithms, they improve on the deterministic rate of convergence
(essentially) by a factor of $m^{-1/2}$, at the cost of having a non-deterministic
algorithm and (in our case) of requiring a non-trivial (albeit mild)
condition on the growth function $\CoeffFunc$ used to define the space $\ApproxSpace[\infty]$.

\begin{theorem}\label{thm:DeterministicIntegrationErrorBound}
  Let $d \in \N$ and $C,\sigma, \alpha \in (0,\infty)$.
  Let $\LayerFunc,\CoeffFunc : \N \to \N \cup \{ \infty \}$ be non-decreasing
  and assume that $\LayerFunc(n) \leq C \cdot (\ln(e n))^{\sigma}$ for all $n \in \N$.
  Then, with $\UnitBall[\infty]$ as in \Cref{eq:UnitBallDefinition}
  and with $T_{\int} : \ApproxSpace[\infty] \to \R, f \mapsto \int_{[0,1]^d} f(x) \, d x$,
  we have
  \[
    \beta_{\ast}^{\deterministic} \bigl(\UnitBall[\infty], T_{\int}\bigr)
    \geq \frac{\alpha}{1 + 2 \alpha}
    \in  \Bigl(0, \frac{1}{2}\Bigr) .
  \]
\end{theorem}

The proof relies on VC-dimension based bounds for empirical processes.
For the convenience of the reader, we briefly review the notion of VC dimension.
Let $\Omega \neq \emptyset$ be a set,
and let ${\emptyset \neq \CalH \subset \{ 0,1 \}^{\Omega}}$ be arbitrary.
In the terminology of machine learning, $\CalH$ is called a \emph{hypothesis class}.
The \emph{growth function} of $\CalH$ is defined as
\[
  \tau_{\CalH} : \quad
  \N \to \N, \quad
  m \mapsto \sup_{x_1,\dots,x_m \in \Omega}
              \big|
                \big\{
                  \big( f(x_1),\CompressedDots,f(x_m) \big)
                  \colon
                  f \in \CalH
                \big\}
              \big| ,
\]
see \cite[Definition~3.6]{MohriFoundationsOfML}.
That is, $\tau_{\CalH}(m)$ describes the maximal number of different ways
in which the hypothesis class $\CalH$ can partition points $x_1,\dots,x_m \in \Omega$.
Clearly, $\tau_{\CalH}(m) \leq 2^m$ for each $m \in \N$.
This motivates the definition of the \emph{VC-dimension} $\VC(\CalH) \in \N_0 \cup \{ \infty \}$
of $\CalH$ as
\[
  \VC(\CalH)
  := \begin{cases}
       0,
       & \text{if } \tau_{\CalH}(1) < 2^1, \\
       \sup \bigl\{ m \in \N \colon \tau_{\CalH}(m) = 2^m \bigr\} \in \N \cup \{ \infty \},
       & \text{otherwise}.
     \end{cases}
\]
For applying existing learning bounds based on the VC dimension
in our setting, the following lemma will be essential.

\begin{lemma}\label{lem:VCUnionBound}
  Let $C_1, C_2, \sigma_1, \sigma_2 > 0$.
  Then there exist constants ${n_0 = n_0(C_1,C_2,\sigma_1,\sigma_2) \in \N}$ and ${C = C(C_1) > 0}$
  such that for every $n \in \N_{\geq n_0}$
  and every $L \in \N$ with $L \leq C_2 \cdot (\ln(e n))^{\sigma_2}$,
  the following holds:

  For any set $\Omega \neq \emptyset$ and any hypothesis classes
  ${\emptyset \neq \CalH_1,\dots,\CalH_N \subset \{ 0, 1 \}^{\Omega}}$ satisfying
  \[
    N \leq L \cdot \tbinom{L n^2}{n}
    \quad \text{ and } \quad
    \VC(\CalH_j) \leq C_1 \cdot n \cdot (\ln(e n))^{\sigma_1}
    \text{ for all } j \in \FirstN{N},
  \]
  we have
  \[
    \VC (\CalH_1 \cup \cdots \cup \CalH_N)
    \leq C \cdot n \cdot (\ln(e n))^{1 + \sigma_1} .
  \]
\end{lemma}

\begin{proof}
  Choose $C_0 = 10 \, C_1$ so that $\ln 2 - \frac{C_1}{C_0} \geq \frac{1}{2}$;
  here we used that $\ln 2 \approx 0.693 \geq \frac{6}{10}$.
  Set $C_3 := 1 + \ln (C_2) + \sigma_2$ and choose $n_0 = n_0(C_1,C_2,\sigma_1,\sigma_2) \in \N$
  so large that for every $n \geq n_0$, we have $C_3 \cdot (\ln (e n))^{-\sigma_1} \leq \frac{1}{6}$
  and $C_1 \, \ln(20 e) \cdot (\ln(e n))^{-1} \leq \frac{1}{6}$.

  For any subset $\emptyset \neq \CalH \subset \{ 0, 1 \}^{\Omega}$,
  Sauer's lemma shows that if $d_\CalH := \VC (\CalH) \in \N$,
  then $\tau_{\CalH}(m) \leq (e m / d_{\CalH})^{d_{\CalH}}$ for all $m \geq d_{\CalH}$;
  see \cite[Corollary~3.18]{MohriFoundationsOfML}.
  An elementary calculation shows that the function $(0,\infty) \to \R, x \mapsto (e m / x)^x$
  is non-decreasing on $(0,e^2 m] \supset (0,m]$; thus, we see
  \begin{equation}
    \tau_{\CalH}(m)
    \leq (e m / d)^d
    \qquad \forall \, m \in \N \text{ and } d \in [d_\CalH, m] \cap [1,\infty) ;
    \label{eq:SauerLemmaApplication}
  \end{equation}
  this trivially remains true if $d_{\CalH} = 0$.

  Let $n \in \N_{\geq n_0}$, $L$, and $\CalH_1,\dots,\CalH_N$ as in the statement of the lemma.
  Set ${\CalH := \CalH_1 \cup \cdots \cup \CalH_N}$
  and $m := \big\lceil C_0 \cdot n \cdot (\ln(e n))^{\sigma_1 + 1}  \big\rceil$;
  we want to show that $\VC(\CalH) \leq m$.
  By definition of the VC dimension, it is sufficient to show that $\tau_\CalH (m) < 2^m$.
  To this end, first note by a standard estimate for binomial coefficients
  (see \mbox{\cite[Exercise~0.0.5]{VershyninHighDimensionalProbability}}) that
  \[
    N
    \leq L \cdot \binom{L n^2}{n}
    \leq L \cdot \bigl(e L n^2 \big/ n\bigr)^n
    \leq (e L^2 n)^n
    =    \exp \bigl(n \cdot \ln(e L^2 n)\bigr)
    \leq \exp \bigl(C_3 n \ln(e n)\bigr) ,
  \]
  thanks to the elementary estimate $\ln x \leq x$,
  since $\ln(e n) \geq 1$ and $L \leq C_2 \cdot (\ln(e n))^{\sigma_2}$,
  and by our choice of $C_3$ at the beginning of the proof.

  Next, recall that $C_0 = 10 \, C_1$ and note
  ${d_{\CalH_j} \leq d := C_1 \cdot n \cdot (\ln(e n))^{\sigma_1} \in [1,m]}$,
  so that \Cref{eq:SauerLemmaApplication} shows because of
  ${m \leq 2 C_0 \cdot n \cdot \ln(e n)^{\sigma_1 + 1}}$ that
  \[
    \tau_{\CalH_j} (m)
    \leq \Big(
           \frac{e m}{C_1 \cdot n \cdot (\ln(e n))^{\sigma_1}}
         \Big)^{C_1 \, n \, (\ln(en))^{\sigma_1}}
    \leq \bigl(20 e \ln(e n)\bigr)^{C_1 \, n \, (\ln(en))^{\sigma_1}} .
  \]
  Combining all these observations and using the subadditivity property
  $\tau_{\CalH_1 \cup \CalH_2} \leq \tau_{\CalH_1} + \tau_{\CalH_2}$
  and the bounds ${m \geq C_0 \, n \, (\ln(e n))^{\sigma_1 + 1}}$
  and $\ln (2) - \frac{C_1}{C_0} \geq \frac{1}{2}$ as well as $C_0 \geq 1$,
  we see with ${\theta := C_0 \, n \, (\ln(e n))^{\sigma_1 + 1}}$ that
  \[
    \begin{split}
      \frac{\tau_{\CalH}(m)}{2^m}
      & \leq \frac{N}{2^m} \cdot
             \bigl(20 e \ln(e n)\bigr)^{C_1 \, n \, (\ln(en))^{\sigma_1}} \\
      & \leq \exp \!
             \big(
               C_3 n \, \ln(e n)
               + C_1 n \, (\ln(e n))^{\sigma_1} \ln(20 e \ln(e n))
               - m \ln(2)
             \big) \\
      & \leq \exp \!
             \Big(
               \!\! - \theta
               \! \cdot \!
               \Bigl[
                 \ln(2)
                 - \frac{C_1}{C_0}
                 - \frac{C_1 \ln(20 e)}{\ln(e n)} 
                 - \frac{C_3}{(\ln(en))^{\sigma_1}} 
               \Bigr]
             \Big) \\
      & \leq \exp \!
             \Big(
               - \theta \cdot \Bigl[\frac{1}{2} - \frac{1}{6} - \frac{1}{6}\Bigr]
             \Big)
        =    \exp \bigl(- \theta \big/ 6\bigr)
        <    1 ,
    \end{split}
  \]
  since $n \geq n_0$ and thanks to our choice of $n_0$ from the beginning of the proof.

  Overall, we have thus shown $\tau_{\CalH}(m) < 2^m$ and hence
  \(
    \VC(\CalH)
    \leq m
    \leq 2 C_0 \cdot n \cdot \, (\ln(en))^{\sigma_1 + 1} ,
  \)
  which completes the proof, for $C := 2 C_0 = 20 \, C_1$.
\end{proof}

As a consequence, we get the following VC-dimension bounds for the network classes
$\NNSigma[n][\infty]$.

\begin{lemma}\label{lem:NNVCDimensionBound}
  Let $d \in \N$ and $\LayerFunc : \N \to \N$
  such that $\LayerFunc(n) \leq C \cdot (\ln(e n))^{\sigma}$ for all $n \in \N$
  and certain $C,\sigma > 0$.
  Then there exist ${n_0 = n_0(C,\sigma,d) \in \N}$ and $C' = C'(C) > 0$ such that for all
  $\lambda \in \R$ and $n \geq n_0$, we have
  \[
    \VC \bigl(\bigl\{ \indicator_{g > \lambda} \colon g \in \NNSigma[n][\infty] \bigr\}\bigr)
    \leq C' \cdot n \cdot (\ln(e n))^{\sigma + 2} .
  \]
\end{lemma}

\begin{proof}
  Given a network architecture $\Architecture = (a_0,\dots,a_K) \in \N^{K+1}$,
  we denote the set of all networks with architecture $\Architecture$ by
  \[
    \CalNN (\Architecture)
    := \prod_{j=1}^K
         \big(
           \R^{a_j \times a_{j-1}} \times \R^{a_j}
         \big) ,
  \]
  and by
  \(
    I(\Architecture)
    := \biguplus_{j=1}^{K}
       \big(
         \{ j \}
         \times \{ 1,\CompressedDots,a_j \}
         \times \{ 1,\CompressedDots,1+a_{j-1} \}
       \big)
  \)
  the corresponding index set, so that $\CalNN(\Architecture) \cong \R^{I(\Architecture)}$.

  Define $L := \LayerFunc(n)$.
  For $\ell \in \{ 1,\dots,L \}$, define $I_\ell := I(\Architecture^{(\ell)})$ and
  $\Architecture^{(\ell)} := (d,n,\dots,n,1) \in \N^{\ell+1}$, as well as
  \[
    \Sigma_\ell
    := \Big\{
         R_\varrho \Phi
         \,\, \colon
         \begin{array}{l}
           \Phi \text{ NN with }
           \din(\Phi) = d,
           \dout(\Phi) = 1,
           \\
           W(\Phi) \leq n,
           L(\Phi) = \ell,
         \end{array}
       \Big\} .
  \]
  By dropping ``dead neurons,'' it is easy to see that each $f \in \Sigma_\ell$
  is of the form ${f = R_\varrho \Phi}$ for some ${\Phi \in \CalNN(\Architecture^{(\ell)})}$
  satisfying $W(\Phi) \leq n$.
  In other words, keeping the identification ${\CalNN(\Architecture) \cong \R^{I(\Architecture)}}$,
  given a subset $S \subset I_\ell$, let us write
  \[
    \CalNN_{S,\ell}
    := \big\{
         R_\varrho \Phi \in \CalNN(\Architecture^{(\ell)})
         \,\,\colon\,\,
         \supp \Phi \subset S
       \big\} ;
  \]
  then ${\Sigma_\ell = \bigcup_{S \subset I_\ell, |S| = n} \CalNN_{S,\ell}}$.
  Moreover, $|I_\ell| = 2d$ if $\ell = 1$
  while $|I_\ell| = 1 + n (d+2) + (\ell-2) (n^2 + n)$ for $\ell \geq 2$,
  and this implies in all cases that $|I_\ell| \leq 2 n (L n + d) \leq L' \cdot n^2$
  for $L' := 4 d \, L$.

  Overall, given a class $\CalF \subset \{ f : \R^d \to \R \}$ and $\lambda \in \R$,
  let us write $\CalF(\lambda) := \{ \indicator_{f > \lambda} \colon f \in \CalF \}$.
  Then the considerations from the preceding paragraph show that
  \begin{equation}
    \NNSigma[n][\infty] (\lambda)
    \subset \bigcup_{\ell=1}^L \,\,
              \bigcup_{S \subset I_\ell, |S| = n} \,\,
                \CalNN_{S,\ell} (\lambda) .
    \label{eq:NetworkUnion}
  \end{equation}
  Now, the set $\CalNN_{S,\ell}$ can be seen as all functions obtained by a fixed ReLU network
  (architecture) with $n$ nonzero weights and $\ell$ layers, in which the weights are allowed to vary.
  Therefore, \mbox{\cite[Equation~(2)]{BartlettNealyTightVCBounds}} shows for a suitable absolute constant
  ${C^{(0)} > 0}$ that
  \[
    \VC(\CalNN_{S,\ell}(\lambda))
    \leq C^{(0)} \cdot n \ell \ln(e n)
    \leq C^{(0)} C \cdot n \cdot (\ln(e n))^{\sigma + 1}
    .
  \]
  Finally, noting that the number of sets over which the union is taken in \Cref{eq:NetworkUnion}
  is bounded by
  \(
    \sum_{\ell=1}^L \binom{|I_\ell|}{n}
    \leq \sum_{\ell=1}^L \binom{L' \, n^2}{n}
    \leq L \cdot \binom{L' \, n^2}{n}
    \leq L' \cdot \binom{L' \, n^2}{n} ,
  \)
  we can apply \Cref{lem:VCUnionBound}
  (with $\sigma_1 = \sigma + 1$, $\sigma_2 = \sigma$, $C_1 = C^{(0)} C$, and $C_2 = 4 d C$)
  to obtain $n_0 = n_0(d,C,\sigma) \in \N$ and $C' = C'(C) > 0$ satisfying
  \(
    \VC (\NNSigma[n][\infty] (\lambda))
    \leq C' \cdot n \cdot (\ln(e n))^{\sigma + 2}
  \)
  for all $n \geq n_0$.
\end{proof}

\begin{proof}[Proof of \Cref{thm:DeterministicIntegrationErrorBound}]
  Define $\theta := \frac{1}{1 + 2 \alpha}$ and ${\gamma := - \frac{\sigma + 2}{1 + 2 \alpha}}$.
  Let $m \geq m_0$ with $m_0$ chosen such that
  ${n := \lfloor m^\theta \cdot (\ln(e m))^\gamma \rfloor}$
  satisfies $n \geq n_0$ for $n_0 = n_0(\sigma,C) \geq 1$ provided by \Cref{lem:NNVCDimensionBound}.
  Let $\CalG := \{ g \in \NNSigma[n][\infty] \colon \| g \|_{L^\infty} \leq 3 \}$
  and note that \Cref{lem:NNVCDimensionBound} shows for every $\lambda \in \R$ that
  ${\VC(\{ \indicator_{g > \lambda} \colon g \in \CalG \}) \leq C' \cdot n \cdot (\ln(en))^{\sigma + 2}}$
  for a suitable constant $C' = C'(C) > 0$.
  Therefore, \mbox{\cite[Proposition~A.1]{CarageaBarronBoundary}} yields a universal constant
  $\kappa > 0$ such that if $X_1,\dots,X_m \overset{\mathrm{iid}}{\sim} U([0,1]^d)$, then
  \[
    \EE
    \bigg[
      \sup_{g \in \CalG}
      \bigg|
        \int_{[0,1]^d}
          g(x)
        d x
        - \frac{1}{m} \sum_{j=1}^m g(X_j)
      \bigg|
    \bigg]
    \leq 6\kappa \sqrt{\frac{C' \, n \, (\ln(en))^{\sigma + 2}}{m}}
    .
  \]
  In particular, there exists ${\x = (X_1,\dots,X_m) \in ([0,1]^d)^m}$ such that
  \[
    \bigg|
      \int_{[0,1]^d}
        g(x)
      d x
      - \frac{1}{m} \sum_{j=1}^m g(X_j)
    \bigg|
    \leq 6\kappa \sqrt{\frac{C' \, n \, (\ln(en))^{\sigma + 2}}{m}}
    =:   \eps_1
    \qquad \forall \, g \in \CalG .
  \]

  Next, note because of $\gamma < 0$ that $n \leq m^\theta \, (\ln (e m))^{\gamma} \leq m^\theta$
  and hence $\ln(e n) \lesssim \ln(m)$.
  Therefore,
  \[
    \eps_1
    \lesssim \sqrt{\frac{n \cdot (\ln(e n))^{\sigma+2}}{m}}
    \lesssim m^{\frac{\theta-1}{2}} \cdot (\ln(e m))^{\frac{\sigma+2+\gamma}{2}}
    =        m^{-\frac{\alpha}{1 + 2\alpha}} \cdot (\ln(em))^{-\alpha \gamma}
    =:       \eps_2 ,
  \]
  where the implied constant only depends on $\alpha$.
  Similarly, we have $n^{-\alpha} \lesssim m^{-\alpha \theta} (\ln(em))^{-\alpha \gamma} = \eps_2$,
  because of $m^\theta \cdot (\ln(em))^{\gamma} \leq n+1 \leq 2 n$.

  Finally, set $Q : \R^m \to \R, (y_1,\dots,y_m) \mapsto \frac{1}{m} \sum_{j=1}^m y_j$ and
  let $f \in \ApproxSpace[\infty]<\infty>$ with $\| f \|_{\ApproxSpace[\infty]<\infty>} \leq 1$
  be arbitrary.
  By \Cref{lem:ApproximationSpaceProperties}, we have $\PreNorm[\infty](f) \leq 1$,
  which implies that $\| f \|_{L^\infty} \leq 1$, and furthermore that there is some
  $g \in \NNSigma[n][\infty]$ satisfying $\| f - g \|_{L^\infty} \leq 2 n^{-\alpha} \leq 2$,
  which in particular implies that $g \in \CalG$.
  Therefore,
  \[
    \begin{split}
      & \Big| \int_{[0,1]^d} f(x) d x - Q\bigl(f(X_1),\dots,f(X_m)\bigr) \Big| \\
      & \leq \Big|
               \int_{[0,1]^d} \!\!\!\!
                 f(x) - g(x)
               \, d x
             \Big|
             + \Big|
                 \int_{[0,1]^d} \!\!\!\!
                   g(x)
                 \, d x
                 - \frac{1}{m} \sum_{j=1}^m g(X_i)
               \Big|
             + \Big|
                 \frac{1}{m}
                 \sum_{j=1}^m
                   (g - f) (X_i)
               \Big| \\
      & \leq 2 \| f - g \|_{L^\infty} + \eps_1
        \lesssim \eps_2 .
    \end{split}
  \]
  Since this holds for all $f \in \UnitBall[\infty]$, with an implied constant
  independent of $f$ and $m$, and since
  $\eps_2 = m^{-\frac{\alpha}{1 + 2 \alpha}} \cdot (\ln(em))^{-\alpha \gamma}$,
  this easily implies
  $\beta_\ast^{\deterministic}(\UnitBall[\infty], T_{\int}) \geq \frac{\alpha}{1 + 2 \alpha}$.
\end{proof}

Our next result shows that Monte Carlo algorithms can improve the rate of convergence
of the deterministic algorithm from \Cref{thm:DeterministicIntegrationErrorBound} by (essentially)
a factor $m^{-1/2}$.
The proof is based on our error bounds for $L^2$ approximation from \Cref{thm:L2ErrorBound}.

\begin{theorem}\label{thm:MonteCarloIntegrationErrorBound}
  Let $d \in \N$, $C_1,C_2,\alpha \in (0,\infty)$, and $\theta,\nu \in [0,\infty)$.
  Let $\LayerFunc : \N \to \N_{\geq 2}$ and $\CoeffFunc : \N \to \N$ be non-decreasing
  and such that $\CoeffFunc(n) \leq C_1 \cdot n^\theta$
  and $\LayerFunc(n) \leq C_2 \cdot \ln^\nu(2 n)$ for all $n \in \N$.
  Let
  \(
    U := \UnitBall[\infty]
       = \big\{
           f \in \ApproxSpace[\infty]
           \colon
           \| f \|_{\ApproxSpace[\infty]}
           \leq 1
         \big\}
  \).

  There exists $C = C(\alpha,\theta,\nu,d,C_1,C_2) > 0$ such that for every $m \in \N$,
  there exists a strongly measurable Monte Carlo algorithm $(\A,\m)$ with $\m \equiv m$
  and $\A = (A_\omega)_{\omega \in \Omega}$ that satisfies
  \begin{equation}
    \bigg(
      \EE \, \Big| A_\omega (f) - \int_{[0,1]^d} f(t) \, d t \Big|
    \bigg)^2
    \leq \EE \bigg[ \Big| A_\omega (f) - \int_{[0,1]^d} f(t) \, d t \Big|^2 \bigg]
    \leq C \cdot \frac{1}{m} \cdot \big( \ln^{1+\nu}(2 m) \big/ m \big)^{\frac{\alpha}{1 + \alpha}}
    \label{eq:MonteCarloIntegrationExplicitBound}
  \end{equation}
  for all $f \in U$.
  In particular, this implies
  \begin{equation}
    \beta_\ast^{\MonteCarlo} \bigl(\UnitBall[\infty], T_{\int}\bigr)
    \geq \frac{1}{2} + \frac{\alpha/2}{1 + \alpha} .
    \label{eq:MonteCarloIntegrationExponentLowerBound}
  \end{equation}
\end{theorem}

\begin{proof}
  Set $Q := [0,1]^d$.
  Let $m \in \N_{\geq 2}$ and $m' := \lfloor \frac{m}{2} \rfloor \in \N$
  and note that $\frac{m}{2} \leq m' + 1 \leq 2 m'$ and hence $\frac{m}{4} \leq m' \leq \frac{m}{2}$.
  Let $C = C(\alpha,\theta,\nu,d,C_1,C_2) > 0$ and $\x = (x_1,\dots,x_{m'}) \in Q^{m'} \strut$
  as provided by \Cref{thm:L2ErrorBound} (applied with $m'$ instead of $m$).
  Note that ${\CalH := \ClosedUnitBall[\infty] \subset C(Q)}$ is closed
  and nonempty, with finite covering numbers $\Covering_{C(Q)}(\CalH,\eps)$,
  for arbitrary $\eps > 0$; see \Cref{lem:ApproxSpaceCoveringBounds}.
  Hence, $\CalH \subset C(Q)$ is compact,
  see for instance \mbox{\cite[Theorem~3.28]{AliprantisBorderHitchhiker}}.
  Let us equip $\CalH$ with the Borel $\sigma$-algebra induced by $C(Q)$.
  Then, it is easy to see from \Cref{lem:MeasurableERM} that the map
  \({
    M :
    \CalH \to \R^{m'},
    f \mapsto \bigl(f(x_1),\dots,f(x_{m'})\bigr)
  }\)
  is measurable and that there is a measurable map $B : \R^{m'} \to \CalH$
  satisfying $B(\y) \in \argmin_{g \in \CalH} \sum_{i=1}^{m'} \bigl(g(x_i) - y_i\bigr)^2$
  for all $\y \in \R^{m'}$.

  \smallskip{}

  Given $f \in \CalH$, note that $g := B(M(f)) \in \CalH$ satisfies $g(x_i) = f(x_i)$
  for all $i \in \FirstN{m'}$, so that \Cref{thm:L2ErrorBound} shows
  \begin{equation}
    \big\| f - B(M(f)) \big\|_{L^2}
    \leq C \cdot \big( \ln^{1+\nu}(2 m') \big/ m' \big)^{\frac{\alpha/2}{1 + \alpha}}
    \leq C' \cdot \big( \ln^{1+\nu}(2 m) \big/ m \big)^{\frac{\alpha/2}{1 + \alpha}} ,
    \label{eq:MonteCarloIntegrationIngredient1}
  \end{equation}
  for a suitable constant $C' = C'(\alpha,\theta,\nu,d,C_1,C_2) > 0$.

  \smallskip{}

  Now, consider the probability space $\Omega = Q^{m'} \cong [0,1]^{m' d}$,
  equipped with the Lebesgue measure $\LebesgueMeasure$.
  For $\z \in \Omega$, write $\Omega \ni \z = (z_1,\dots,z_{m'})$ and define
  \[
    \Psi : \quad
    \Omega \times C(Q) \to \R, \quad
    (\z, g) \mapsto \frac{1}{m'} \sum_{j=1}^{m'} g(z_j) .
  \]
  It is easy to see that $\Psi$ is continuous and hence measurable;
  see \Cref{eq:PointEvaluationJointlyContinuous} for more details.

  Note that for $\z = (z_1,\dots,z_{m'}) \in \Omega$,
  the random vectors $z_1,\dots,z_{m'} \in Q$ are stochastically independent.
  Furthermore, for arbitrary $g \in C(\Omega)$, we have
  $\EE_{\z} [g(z_j)] = \int_{[0,1]^d} g(t) \, dt = T_{\int}(g)$.
  Using the additivity of the variance for independent random variables, this entails
  \begin{equation}
    \begin{split}
      \EE_{\z} \Big[ \big( \Psi(\z,g) - T_{\int}(g) \big)^2 \Big]
      & = \mathrm{Var} \bigl(A(\z,g)\bigr)
        = \big( 1 \big/ m' \big)^{2}
          \sum_{j=1}^{m'} \mathrm{Var} \bigl(g(z_j)\bigr) \\
      & \leq \big( 1 \big/ m' \big)^{2}
             \sum_{j=1}^{m'}
               \int_{[0,1]^d}
                 |g(x)|^2
               \, d x
        =    \frac{\| g \|_{L^2}^2}{m'} .
    \end{split}
    \label{eq:HonestToGodMonteCarloIntegration}
  \end{equation}

  Finally, for each $\z \in \Omega$ define
  \[
    A_{\z} : \quad
    \CalH \to \R, \quad
    f \mapsto \Psi\bigl(\z, f - B(M(f))\bigr) + T_{\int}\bigl(B(M(f))\bigr)
  \]
  Since the map $T_{\int} : C([0,1]^d) \to \R$ is continuous and hence measurable,
  it is easy to verify that $\Omega \times \UnitBall[\infty] \ni (\z,f) \mapsto A_{\z}(f)$
  is measurable.
  Furthermore, explicitly writing out the definition of $A_{\z}$ shows that
  \[
    A_{\z} (f)
    = \frac{1}{m'}
      \sum_{j=1}^{m'}
        f(z_j)
      - \frac{1}{m'}
        \sum_{j=1}^{m'}
          B\bigl(f(x_1),\dots,f(x_{m'})\bigr) (z_j)
      + T_{\int} \bigl(B (f(x_1),\dots,f(x_{m'}))\bigr)
  \]
  only depends on $m' + m' \leq m$ point samples of $f$.
  Thus, if we set $\m \equiv m$, then $(\A,\m)$ is a strongly measurable
  Monte Carlo algorithm $(\A,\m) \in \Alg_m^{\MonteCarlo}(\UnitBall[\infty], \R)$.

  To complete the proof, note that a combination of
  \Cref{eq:MonteCarloIntegrationIngredient1,eq:HonestToGodMonteCarloIntegration} shows
  \begin{align*}
    \EE_{\z}
    \Big[
      \big( A_{\z}(f) - T_{\int}(f) \big)^2
    \Big]
    & = \EE_{\z}
        \Big[
          \big( \Psi(\z, f - B(M(f))) - T_{\int}(f - B(M(f))) \big)^2
        \Big] \\
    & \leq \frac{1}{m'} \big\| f - B(M(f)) \big\|_{L^2}^2
      \leq 4 \, (C')^2 \cdot m^{-1} \cdot \big( \ln^{1+\nu}(2m) \big/ m \big)^{\frac{\alpha}{1+\alpha}} .
  \end{align*}
  for all $f \in U$.
  Combined with Jensen's inequality, this proves \Cref{eq:MonteCarloIntegrationExplicitBound}
  for the case $m \in \N_{\geq 2}$.
  The case $m = 1$ can be handled by taking $A_{\omega} \equiv 0$ and possibly enlarging
  the constant $C$ in \Cref{eq:MonteCarloIntegrationExplicitBound}.
  Directly from the definition of $\beta_{\ast}^{\MonteCarlo} (\UnitBall[\infty], T_{\int})$,
  we see that \Cref{eq:MonteCarloIntegrationExplicitBound} implies
  \Cref{eq:MonteCarloIntegrationExponentLowerBound}.
\end{proof}


\section{Hardness of numerical integration}%
\label{sec:IntegrationHardness}


Our goal in this section is to prove upper bounds for the optimal order
$\beta_\ast (\UnitBall[\infty], T_{\int})$ of quadrature on the neural network
approximation spaces, both for deterministic and randomized algorithms.
Our bounds for the deterministic setting in particular show that
\emph{regardless of the ``approximation exponent'' $\strut \alpha$},
the quadrature error given $m$ point samples
can never decay faster than $\CalO\bigl(m^{- \min \{2, 2 \alpha\}}\bigr)$.
In fact, if the depth growth function $\LayerFunc$ is unbounded,
or if the weight growth function $\strut \CoeffFunc$ grows sufficiently fast
(so that ${\gamma^{\flat}(\LayerFunc,\CoeffFunc) = \infty}$),
then no better rate than $\CalO\bigl(m^{- \min\{1,\alpha\}}\bigr)$ is possible.

For the case of of Monte Carlo algorithms, the bound that we derive shows that
the expected quadrature error given at most $m$ point samples (in expectation)
can never decay faster than $\CalO \big( m^{- \min \{ 2, \frac{1}{2} + 2 \alpha \}} \big)$.
In fact, if $\gamma^\flat = \infty$ then the error can not decay faster than
$\CalO \big( m^{- \min \{ 1, \frac{1}{2} + \alpha \}} \big)$.


Our precise bound for the deterministic setting reads as follows:

\begin{theorem}\label{thm:QuadratureDeterministicHardness}
  Let $\LayerFunc : \N \to \N_{\geq 2} \cup \{ \infty \}$ and $\CoeffFunc : \N \to \N \cup \{ \infty \}$
  be non-decreasing, and let $d \in \N$ and $\alpha > 0$.
  Let $\gamma^\flat := \gamma^\flat (\LayerFunc,\CoeffFunc)$ as in \Cref{eq:GammaDefinition}
  and $\UnitBall[\infty] ([0,1]^d)$ as in \Cref{eq:UnitBallDefinition}.
  For the operator
  $T_{\int} : \UnitBall[\infty] ([0,1]^d) \to \R, f \mapsto \int_{[0,1]^d} f(x) \, d x$,
  we then have
  \begin{align}
    \beta_\ast^{\deterministic}(\UnitBall[\infty], T_{\int})
    & \leq \begin{cases}
             \frac{2 \alpha}{\alpha + \gamma^\flat} ,
             & \text{if } \alpha + \gamma^\flat < 2, \\
             \min \big\{ \alpha, 1 + \frac{\alpha - 1}{\alpha + \gamma^\flat - 1} \big\}
             & \text{if } \alpha + \gamma^\flat \geq 2
           \end{cases}
           \label{eq:DeterministicIntegrationHardnessBound1}
           \\
    & =  \begin{cases}
           \frac{2 \alpha}{\alpha + \gamma^{\flat}},
           & \text{if } \alpha + \gamma^{\flat} \leq 2 \\
           \alpha ,
           & \text{if } \alpha + \gamma^{\flat} > 2
             \text{ and } \alpha \leq 1 , \\
           1 + \frac{\alpha - 1}{\alpha + \gamma^\flat - 1}
           & \text{if } \alpha + \gamma^{\flat} \geq 2
             \text{ and } \alpha >    1 .
         \end{cases}
         \label{eq:DeterministicIntegrationHardnessBound2}
  \end{align}
\end{theorem}

\begin{rem*}
  Since the bound above might seem intimidating at first sight,
  we discuss a few specific consequences.
  First, the theorem implies
  \({
    \beta_{\ast}^{\deterministic} (\UnitBall[\infty], T_{\int})
    \leq \max
         \big\{
           \alpha, \frac{2 \alpha}{\alpha + \gamma^{\flat}}
         \big\}
    \leq \max \{ 1, \frac{2}{\gamma^\flat} \} \alpha
    \leq 2 \alpha
  }\)
  and hence $\beta_{\ast}^{\deterministic} (\UnitBall[\infty], T_{\int}) \to 0$
  as $\alpha \downarrow 0$.
  Furthermore, the theorem shows that
  ${\beta_{\ast}^{\deterministic} (\UnitBall[\infty], T_{\int}) \leq 2}$,
  and if $\gamma^\flat = \infty$, then in fact
  $\beta_{\ast}^{\deterministic} (\UnitBall[\infty], T_{\int}) \leq \min \{ \alpha, 1 \}$.
\end{rem*}

\begin{proof}
  For brevity, set $U := \UnitBall[\infty]$.

  \textbf{Step~1:} Let $0 < \gamma < \gamma^\flat$, $\theta \in (0,\infty)$,
  and $\lambda \in [0,1]$ with $\theta \lambda \leq 1$ be arbitrary and define
  ${\omega := \min \{ -\theta \alpha, \theta \cdot (\gamma - \lambda) - 1 \}}$.
  In this step, we show that
  \begin{equation}
    e(A,U,T_{\int})
    \geq \kappa_2 \cdot m^{-(1 - \omega - \theta \lambda)}
    \qquad \forall \, m \in \N \text{ and } A \in \Alg_m(U,\R),
    \label{eq:DeterministicIntegrationHardnessProofStep1}
  \end{equation}
  for a suitable constant
  ${\kappa_2 = \kappa_2 (\alpha,\gamma,\theta,\lambda,\LayerFunc,\CoeffFunc) > 0}$.

  To see this, let $m \in \N$ and $A \in \Alg_m(U,\R)$ be arbitrary.
  By definition, this means that there exist $Q : \R^m \to \R$
  and $\x = (x_1,\dots,x_m) \in ([0,1]^d)^m$ satisfying
  $A(f) = Q\bigl(f(x_1),\dots,f(x_m)\bigr)$ for all $f \in U$.
  Set $M := 4 m$ and let $z_j := \frac{1}{4m} + \frac{j-1}{2m}$ for $j \in \FirstN{2m}$
  as in \Cref{lem:HatSumsInUnitBall}.
  Furthermore, choose
  \(
    I
    := I_{\x}
    := \big\{
         i \in \FirstN{2 m}
         \,\,\, \colon \,\,\,
         \forall \, n \in \FirstN{m}:
           \Lambda_{M,z_i}^\ast (x_n) = 0
       \big\}
  \)
  and recall from \Cref{lem:AvoidingSamplingPoints} that $|I| \geq m$.
  Define $k := \lceil m^{\theta \lambda} \rceil$ and note
  $k \leq 1 + m^{\theta \lambda} \leq 2 \, m^{\theta \lambda}$.
  Since $\theta \lambda \leq 1$, we also have $k \leq \lceil m \rceil = m \leq |I|$.
  Hence, there is a subset $J \subset I$ satisfying $|J| = k$.

  Now, an application of \Cref{lem:HatSumsInUnitBall} yields a constant
  $\kappa_1 = \kappa_1(\alpha,\gamma,\theta,\lambda,\LayerFunc,\CoeffFunc) > 0$
  (independent of $m$ and $A$) such that
  $f := \kappa_1 \, m^\omega \, \sum_{j \in J} \Lambda_{M,z_j}^\ast$ satisfies $\pm f \in U$.
  Since $J \subset I$, we see by definition of $I = I_{\x}$ that $f(x_n) = 0$
  for all $n \in \FirstN{2m}$ and hence $A(\pm f) = Q(0,\dots,0) =: \mu$.
  Using the elementary estimate
  \(
    \max \{ |x-\mu|, |-x-\mu| \}
    \geq \frac{1}{2} \big( |x-\mu| + |x+\mu| \big)
    \geq \frac{1}{2} |x-\mu+x+\mu|
    =    |x| ,
  \)
  we thus see
  \begin{align*}
    e(A,U,T_{\int})
    & \geq \max
           \Big\{
             \bigl|T_{\int}(f) - Q\bigl(f(x_1),\dots,f(x_m)\bigr)\bigr|, \quad
             \bigl|T_{\int}(-f) - Q\bigl(-f(x_1),\dots,-f(x_m)\bigr)\bigr|
           \Big\} \\
    & \geq \max
           \Big\{
             \bigl| T_{\int}(f) - \mu \bigr|, \quad
             \bigl| - T_{\int}(f) - \mu \bigr|
           \Big\} \\
    & \geq |T_{\int}(f)|
      =    \kappa_1 \cdot m^\omega \cdot \frac{|J|}{M}
      \overset{(\ast)}{\geq} \frac{\kappa_1}{4} \cdot m^{\omega - 1 + \theta \lambda}
      =:   \kappa_2 \cdot m^{-(1 - \omega - \theta \lambda)} ,
  \end{align*}
  as claimed in \Cref{eq:DeterministicIntegrationHardnessProofStep1}.
  Here, the step marked with $(\ast)$ used that $|J| = k \geq m^{\theta \lambda}$
  and that $M = 4 m$.

  \medskip{}

  \textbf{Step~2 (Completing the proof):}
  \Cref{eq:DeterministicIntegrationHardnessProofStep1} shows that
  $e_m^{\deterministic}(U,T_{\int}) \geq \kappa_2 \cdot m^{-(1-\omega-\theta\lambda)}$
  for all $m \in \N$, with $\kappa_2 > 0$ independent of $m$.
  Directly from the definition of $\beta_\ast^{\deterministic}(U,T_{\int})$ and $\omega$,
  this shows
  \[
    \beta_\ast^{\deterministic} (U, T_{\int})
    \leq 1 - \omega - \theta \lambda
    =    1
         + \max
           \big\{
             \theta \cdot (\alpha - \lambda), \quad
             1 + \theta \cdot (\lambda - \gamma) - \theta \lambda
           \big\}
    =    1
         + \max
           \big\{
             \theta \cdot (\alpha - \lambda), \quad
             1 - \theta \gamma
           \big\} ,
  \]
  and this holds for arbitrary $0 < \gamma < \gamma^\flat$, $\theta \in (0,\infty)$,
  and $\lambda \in [0,1]$ satisfying $\theta \lambda \leq 1$.
  It is easy (but somewhat tedious) to shows that this implies
  \Cref{eq:DeterministicIntegrationHardnessBound1}; see \Cref{lem:OptimizationLemma2}
  for the details.
  Finally, \Cref{eq:DeterministicIntegrationHardnessBound2} follows from
  \Cref{eq:DeterministicIntegrationHardnessBound1} via an easy case distinction.
\end{proof}

As our next result, we derive a hardness results for Monte Carlo algorithms
for integration on the neural network approximation space $\ApproxSpace[\infty]$.
The proof hinges on \emph{Khintchine's inequality}, which states the following:

\begin{proposition}\label{prop:KhintchineInequality}%
  (\cite[Theorem~1 in Section~10.3]{ChowTeicherProbabilityTheory})
  Let $n \in \N$ and let $(X_i)_{i=1,\dots,n}$ be independent random variables
  (one some probability space $(\Omega,\CalF,\PP)$) that are Rademacher distributed
  (i.e., $\PP(X_i = 1) = \frac{1}{2} = \PP(X_i = -1)$ for each $i \in \FirstN{n}$).
  Then for each $p \in (0,\infty)$ there exist constants $A_p,B_p \in (0,\infty)$
  (only depending on $p$) such that for arbitrary $c = (c_i)_{i=1,\dots,n} \subset \R$,
  the following holds:
  \[
    A_p \cdot \bigg( \sum_{i=1}^n c_i^2 \bigg)^{1/2}
    \leq \bigg\| \sum_{i=1}^n c_i \, X_i \bigg\|_{L^p(\PP)}
    = \bigg(\, \EE \bigg| \sum_{i=1}^n c_i \, X_i \bigg|^p \,\bigg)^{1/p}
    \leq B_p \cdot \bigg( \sum_{i=1}^n c_i^2 \bigg)^{1/2}
  \]
\end{proposition}

\begin{remark}\label{rem:KhintchineRemark}
  Applying Khintchine's inequality for $p = 1$ and $c_i = 1$, we see
  \begin{equation}
    \avsum_{\nu \in \{ \pm 1 \}^n} \,\,
    \bigg|\,
      \sum_{i=1}^n \nu_i
    \,\bigg|
    \geq A_1 \cdot n^{1/2} ,
    \label{eq:SpecialKhintchine}
  \end{equation}
  which is what we will actually use below.
\end{remark}

Our precise hardness result for integration using Monte Carlo algorithms reads as follows.

\begin{theorem}\label{thm:QuadratureMonteCarloHardness}
  Let $\LayerFunc: \N \to \N_{\geq 2} \cup \{ \infty \}$
  and $\CoeffFunc : \N \to \N \cup \{ \infty \}$ be non-decreasing.
  Let $d \in \N$ and $\alpha \in (0,\infty)$.
  Let $\gamma^\flat := \gamma^\flat (\LayerFunc,\CoeffFunc)$ as in \Cref{eq:GammaDefinition}
  and $\UnitBall[\infty] ([0,1]^d)$ as in \Cref{eq:UnitBallDefinition}.
  For the operator
  $T_{\int} : \UnitBall[\infty] ([0,1]^d) \to \R, f \mapsto \int_{[0,1]^d} f(x) \, d x$,
  we then have
  \begin{equation}
    \begin{split}
      \beta_\ast^{\MonteCarlo}\bigl(\UnitBall[\infty], T_{\int}\bigr)
      & \leq \begin{cases}
               \min
               \big\{
                 1 + \frac{\alpha}{\alpha + \gamma^\flat}, \,\,
                 \frac{1}{2} + \frac{2 \alpha}{\alpha + \gamma^\flat}
               \big\} ,
               & \text{if } \alpha + \gamma^\flat <    2, \\
               \min
               \big\{
                 1 + \frac{\alpha}{\alpha + \gamma^\flat}, \,\,
                 \frac{1}{2} + \alpha, \,\,
                 1 + \frac{\alpha - \frac{1}{2}}{\alpha + \gamma^\flat - 1}
               \big\} ,
               & \text{if } \alpha + \gamma^\flat \geq 2.
             \end{cases} \\
      & =    \begin{cases}
               \frac{1}{2} + \frac{2 \alpha}{\alpha + \gamma^{\flat}} ,
               & \text{if } \alpha + \gamma^{\flat} < 2, \\
               \frac{1}{2} + \alpha ,
               & \text{if } \alpha + \gamma^{\flat} \geq 2
                 \text{ and } \alpha \leq \frac{1}{2}, \\
               1 + \frac{\alpha - \frac{1}{2}}{\alpha + \gamma^{\flat} - 1},
               & \text{if } \alpha + \gamma^{\flat} \geq 2
                 \text{ and } \frac{1}{2} \leq \alpha \leq \gamma^{\flat}, \\
               1 + \frac{\alpha}{\alpha + \gamma^{\flat}},
               & \text{if } \alpha + \gamma^{\flat} \geq 2
                 \text{ and } \alpha \geq \gamma^{\flat}.
             \end{cases}
    \end{split}
    \label{eq:QuadratureMonteCarloHardness}
  \end{equation}
\end{theorem}

\begin{rem*}
  We discuss a few special cases.
  First, we always have
  \(
    \beta_\ast^{\MonteCarlo}(\UnitBall[\infty], T_{\int})
    \leq 1 + \frac{\alpha}{\alpha + \gamma^\flat}
    \leq 2 ,
  \)
  which shows that \emph{no matter how large the approximation rate $\alpha$ is},
  one can never get an (asymptotically) better error bound than $m^{-2}$.
  Furthermore, if $\gamma^\flat = \infty$ (for instance if $\LayerFunc$ is unbounded), then
  \(
    \beta_\ast^{\MonteCarlo}(\UnitBall[\infty], T_{\int})
    \leq 1 + \frac{\alpha}{\alpha + \gamma^\flat}
    = 1 .
  \)

  The previous bounds are informative for (somewhat) large $\alpha$.
  For small $\alpha > 0$, the theorem shows
  \(
    \beta_\ast^{\MonteCarlo}(\UnitBall[\infty])
    \leq \frac{1}{2}
         + \max
           \big\{
             \frac{2 \alpha}{\alpha + \gamma^\flat}, \,\,
             \alpha
           \big\}
    \leq \frac{1}{2}
         + \max \big\{ \frac{2}{\gamma^\flat}, \,\, 1 \big\} \alpha
    \leq \frac{1}{2} + 2 \alpha .
  \)
\end{rem*}

\begin{proof}
  For brevity, set $U := \UnitBall[\infty]$ and $\gamma^\flat := \gamma^\flat (\LayerFunc,\CoeffFunc)$.
  The main idea of the proof is to apply \Cref{lem:MonteCarloHardnessThroughAverageCase}
  for a suitable choice of the family of functions
  $(f_{\boldnu,J})_{(\boldnu,J) \in \Gamma_m} \subset U$.

  \smallskip{}

  \textbf{Step~1 (Preparation):}
  Let $0 < \gamma < \gamma^\flat$, $\theta \in (0,\infty)$, and $\lambda \in [0,1]$
  with $\theta \lambda \leq 1$ be arbitrary and define
  $\omega := \min \{ -\theta \alpha, \theta \cdot (\gamma - \lambda) - 1 \}$.
  Given a fixed but arbitrary $m \in \N$,
  set $M := 4 m$ and $z_j := \frac{1}{4 m} + \frac{j - 1}{2 m}$ as in \Cref{lem:HatSumsInUnitBall}.
  Furthermore, let $k := \big\lceil m^{\theta \lambda} \big\rceil$ and note
  because of $\theta \lambda \leq 1$ that $k \leq \lceil m \rceil = m$
  and $k \leq 1 + m^{\theta \lambda} \leq 2 \, m^{\theta \lambda}$.

  Define $\CalP_k (\FirstN{2 m}) := \{ J \subset \FirstN{2m} \colon |J| = k \}$
  and $\Gamma_m := \{ \pm 1 \}^{2 m} \times \CalP_k (\FirstN{2 m})$.
  Then, \Cref{lem:HatSumsInUnitBall} yields a constant
  $\kappa_1 = \kappa_1(\gamma,\theta,\lambda,\alpha,\LayerFunc,\CoeffFunc) > 0$
  such that for any $(\boldnu,J) \in \Gamma_m$, the function
  \[
    f_{\boldnu,J}
    := \kappa_1 \, m^\omega \, \sum_{j \in J} \nu_j \, \Lambda_{M,z_j}^\ast
    \quad \text{satisfies} \quad
    f_{\boldnu,J} \in U .
  \]

  \smallskip{}

  \textbf{Step~2:} We show for $\gamma,\theta,\lambda,\omega$ as in Step~2 that there exists
  ${\kappa_3 = \kappa_3(\gamma,\theta,\lambda,\alpha,\LayerFunc,\CoeffFunc) \!>\! 0}$
  (independent of $m \in \N$) such that
  \begin{equation}
    \avsum_{(\boldnu,J) \in \Gamma_m}
      \bigl|T_{\int}(f_{\boldnu, J}) - A(f_{\boldnu,J})\bigr|
    \geq \kappa_3 \cdot m^{-(1 - \frac{\theta \lambda}{2} - \omega)}
    \qquad \forall \, m \in \N \text{ and } A \in \Alg_m (U, \R) .
    \label{eq:IntegrationAverageCaseHardness}
  \end{equation}
  To see this, let $A \in \Alg_m (U, \R)$ be arbitrary.
  By definition, we have $A(f) = Q\bigl(f(x_1),\dots,f(x_m)\bigr)$ for all $f \in U$,
  for suitable $\x = (x_1,\dots,x_m) \in ([0,1]^d)^m$ and $Q : \R^m \to \R$.
  Now, define
  \({
    I
    := I_{\x}
    := \{
         j \in \FirstN{2 m}
         \,\,\colon\,\,
         \forall \, n \in \FirstN{m} :
           \Lambda_{M,z_j}^\ast (x_n) = 0
       \}
  }\)
  and recall from \Cref{lem:AvoidingSamplingPoints} that $|I| \geq m$.

  Set $I^c := \FirstN{2m} \setminus I$.
  For $\boldnu^{(1)} = (\nu_j)_{j \in I} \in \{ \pm 1 \}^I$ and
  $\boldnu^{(2)} := (\nu_j)_{j \in I^c} \in \{ \pm 1 \}^{I^c}$ and $J \in \CalP_k(\FirstN{2m})$,
  define
  \[
    g_{\boldnu^{(1)},J}
    := \kappa_1 \, m^\omega \,
       \sum_{j \in J \cap I}
         \nu_j^{(1)} \, \Lambda_{M,z_j}^\ast
    \qquad \text{and} \qquad
    h_{\boldnu^{(2)}, J}
    := \kappa_1 \, m^\omega \,
       \sum_{j \in J \cap I^c}
         \nu_j^{(2)} \, \Lambda_{M,z_j}^\ast
    .
  \]
  Furthermore, define
  \(
    \mu_{\boldnu^{(2)}, J}
    := T_{\int}(h_{\boldnu^{(2)},J})
       - Q\big( h_{\boldnu^{(2)},J}(x_1), \dots, h_{\boldnu^{(2)},J}(x_m) \big) .
  \)
  By choice of $I$, we have $g_{\boldnu^{(1)},J}(x_n) = 0$ for all $n \in \FirstN{m}$,
  and hence $f_{\boldnu,J}(x_n) = h_{\boldnu^{(2)},J}(x_n)$, if we identify $\boldnu$
  with $(\boldnu^{(1)},\boldnu^{(2)})$, as we will do for the remainder of this step.

  Finally, recall from \Cref{lem:HatSumsInUnitBall} that $\supp \Lambda_{M,z_j}^\ast \subset [0,1]^d$
  and hence $T_{\int}(\Lambda_{M,z_j}^\ast) = M^{-1} = \frac{1}{4 m}$.
  Overall, we thus see for arbitrary $J \in \CalP_k(\FirstN{2m})$
  and $\boldnu^{(2)} \in \{ \pm 1 \}^{I^c}$ that
  \begin{equation}
    \begin{split}
      & \avsum_{\boldnu^{(1)} \in \{ \pm 1 \}^I}
          \Big|
            T_{\int}(f_{\boldnu,J})
            - Q \big( f_{\boldnu,J}(x_1), \dots, f_{\boldnu,J}(x_m) \big)
          \Big| \\
      & = \avsum_{\boldnu^{(1)} \in \{ \pm 1 \}^I}
            \Big|
              T_{\int}(g_{\boldnu^{(1)},J})
              + \mu_{\boldnu^{(2)},J}
            \Big|
        = \avsum_{\boldnu^{(1)} \in \{ \pm 1 \}^I}
            \bigg|
              \frac{\kappa_1 \, m^\omega}{M}
              \sum_{j \in J \cap I}
                \nu_j^{(1)}
              + \mu_{\boldnu^{(2)},J}
            \bigg| \\
      & \overset{(\ast)}{=}
        \frac{1}{2}
        \avsum_{\boldnu^{(3)} \in \{ \pm 1 \}^{I \cap J}}
        \bigg(
          \Big|
            \frac{\kappa_1 \, m^\omega}{M}
            \sum_{j \in J \cap I}
              \nu^{(3)}_j
            + \mu_{\boldnu^{(2)},J}
          \Big|
          + \Big|
              \frac{\kappa_1 \, m^\omega}{M}
              \sum_{j \in J \cap I}
                (-\nu^{(3)}_j)
              + \mu_{\boldnu^{(2)},J}
            \Big|
        \bigg)
        \\
      & \overset{(\blacklozenge)}{\geq}
        \avsum_{\boldnu^{(3)} \in \{ \pm 1 \}^{I \cap J}}
          \Big|
            \frac{\kappa_1 \, m^\omega}{M}
            \sum_{j \in J \cap I}
              \nu^{(3)}_j
          \Big|
        \geq \kappa_2 \, m^{\omega - 1} \cdot |J \cap I|^{1/2}
    \end{split}
    \label{eq:IntegrationMonteCarloHardnessMainStep}
  \end{equation}
  for a suitable constant $\kappa_2 = \kappa_2(\gamma,\theta,\lambda,\alpha,\LayerFunc,\CoeffFunc) > 0$.
  Here, the very last step used \Cref{eq:SpecialKhintchine} and the identity $M = 4m$.
  Furthermore, the step marked with $(\ast)$ used that
  \[
    \avsum_{\boldsymbol{\sigma} \in \{ \pm 1 \}^K}
      a_{\boldsymbol{\sigma}}
    = \frac{1}{2}
      \bigg(
        \avsum_{\boldsymbol{\sigma} \in \{ \pm 1 \}^K} a_{\boldsymbol{\sigma}}
        + \avsum_{\boldsymbol{\sigma} \in \{ \pm 1 \}^K} a_{\boldsymbol{\sigma}}
      \bigg)
    = \frac{1}{2}
      \bigg(
        \avsum_{\boldsymbol{\sigma} \in \{ \pm 1 \}^K} a_{\boldsymbol{\sigma}}
        + \avsum_{\boldsymbol{\sigma} \in \{ \pm 1 \}^K} a_{-\boldsymbol{\sigma}}
      \bigg) ,
  \]
  while the elementary estimate $|x + y| + |-x + y| = |x+y| + |x-y| \geq |x+y+x-y| = 2 |x|$
  was used at the step marked with $(\blacklozenge)$.

  Combining \Cref{eq:IntegrationMonteCarloHardnessMainStep}
  and \Cref{lem:RandomSubsetExpectationBound}, we finally obtain
  $\kappa_3 = \kappa_3(\gamma,\theta,\lambda,\alpha,\LayerFunc,\CoeffFunc) > 0$ satisfying
  \begin{align*}
    \avsum_{(\boldnu,J) \in \Gamma_m}
      \big|
        T_{\int}(f_{\boldnu,J}) - A(f_{\boldnu,J})
      \big|
    & = \avsum_{J \in \CalP_k(m)}
          \avsum_{\boldnu^{(2)} \in \{ \pm 1 \}^{I^c}}
            \avsum_{\boldnu^{(1)} \in \{ \pm 1 \}^{I}}
              \big|
                T_{\int}(f_{\boldnu,J}) - A(f_{\boldnu,J})
              \big| \\
    & \geq \kappa_2 \, m^{\omega - 1}
           \avsum_{J \in \CalP_k(m)}
             |J \cap I|^{1/2}
      \geq \kappa_3 \, m^{\omega - 1} \cdot k^{1/2}
      \geq \kappa_3 \, m^{\omega - 1 + \frac{\theta\lambda}{2}} ,
  \end{align*}
  as claimed in \Cref{eq:IntegrationAverageCaseHardness}.
  Since $m \in \N$ and $A \in \Alg_m(U;\R)$ were arbitrary and $\kappa_3$
  is independent of $A$ and $m$, Step~2 is complete.

  \medskip{}

  \textbf{Step~3:}
  In view of \Cref{eq:IntegrationAverageCaseHardness}, a direct application of
  \Cref{lem:MonteCarloHardnessThroughAverageCase} shows that
  \[
    \beta_\ast^{\MonteCarlo} (U, T_{\int})
    \leq 1 - \omega - \tfrac{\theta \lambda}{2}
    =    1
         + \max
           \big\{
             \theta \cdot (\alpha - \tfrac{\lambda}{2}), \,\,
             1 + \theta \cdot (\tfrac{\lambda}{2} - \gamma)
           \big\}
  \]
  for arbitrary $0 < \gamma < \gamma^\flat$, $\theta \in (0,\infty)$, and $\lambda \in [0,1]$
  with $\theta \lambda \leq 1$.
  From this, the first part of \Cref{eq:QuadratureMonteCarloHardness} follows by a straightforward
  but technical computation; see \Cref{lem:OptimizationLemma1} for the details.
  The second part of \Cref{eq:QuadratureMonteCarloHardness} follows from the first one
  by a straightforward case distinction.
\end{proof}

\appendix

\section{Postponed technical results and proofs}%
\label{sec:TechnicalResults}


\subsection{Proof of Lemma~\ref{lem:ApproximationSpaceProperties}}%
\label{sub:NNApproximationSpacesProofs}

This section provides the proof of \Cref{lem:ApproximationSpaceProperties},
which is based on the following lemma concerning closure properties
of the sets $\NNSigma$.


\begin{lemma}\label{lem:NetworkSetsClosedUnderSummation}
  With $\widetilde{\LayerFunc}(n) := \min \{ \LayerFunc(n), n \}$, we have
  $\NNSigma = \NNSigma<\tilde{\LayerFunc}>$.
  Furthermore, for every $n \in \N$, we have
  $\NNSigma + \NNSigma \subset \NNSigma[9n]$.
\end{lemma}

\begin{proof}
  We first prove $\NNSigma = \NNSigma<\tilde{\LayerFunc}>$.
  To this end, we prove for fixed $n \in \N$ by induction on $\ell \in \N_{\geq n}$ that
  $\NNSigma<\ell> \subset \NNSigma<n>$.
  For $\ell = n$, this is trivial.
  Thus, suppose that $\NNSigma<\ell> \subset \NNSigma<n>$ for some $\ell \in \N_{\geq n}$
  and let $f \in \NNSigma<\ell+1>$, say $f = R_\varrho \Phi$ with $\WeightSize{\Phi} \leq \CoeffFunc(n)$
  and $W(\Phi) \leq n$, as well as $L(\Phi) \leq \ell + 1$.
  If $L(\Phi) \leq \ell$, then $f \in \NNSigma<\ell> \subset \NNSigma<n>$ by induction.
  Hence, we can assume that $L(\Phi) = \ell+1$.

  Writing $\Phi = \big( (A_1,b_1), \CompressedDots, (A_{\ell+1}, b_{\ell+1}) \big)$ with
  $b_m \in \R^{N_m}$ and $A_m \in \R^{N_m \times N_{m-1}}$,
  we have ${A_j = b_j = 0}$ for some $j \in \FirstN{\ell+1}$, since otherwise
  \({
    n\!+\!1
    \leq \ell\!+\!1
    \leq \sum_{j=1}^{\ell+1} \bigl(\| A_j \|_{\ell^0} \!+\! \| b_j \|_{\ell^0}\bigr)
    \!=\! W(\Phi)
    \leq n
    .
  }\)
  If $j = \ell+1$, we trivially have $f \equiv 0 \in \NNSigma<n>$;
  thus, let us assume $j \leq \ell$ and define
  \[
    \widetilde{\Phi}
    := \big(
         (0_{N_{j+1} \times d}, b_{j+1}),
         (A_{j+2}, b_{j+2}),
         \dots,
         (A_{\ell+1}, b_{\ell+1})
       \big)
    .
  \]
  Since $A_j = b_j = 0$ and $\varrho(0) = 0$, it is straightforward to verify
  $R_\varrho \widetilde{\Phi} = R_\varrho \Phi = f$.
  Since furthermore $\WeightSize{\widetilde{\Phi}} \leq \WeightSize{\Phi} \leq \CoeffFunc(n)$
  and $W(\widetilde{\Phi}) \leq W(\Phi) \leq n$, as well as
  $L(\widetilde{\Phi}) \leq \ell - j + 1 \leq \ell$,
  this implies $f \in \NNSigma<\ell> \subset \NNSigma<n>$,
  where the last inclusion holds by induction.
  This completes the induction.

  \medskip{}

  To prove $\NNSigma + \NNSigma \subset \NNSigma[5n]$, let $f,g \in \NNSigma$,
  so that $f = R_\varrho \Phi$ and $g = R_\varrho \Psi$ for networks $\Phi,\Psi$ satisfying
  ${W(\Phi), W(\Psi) \leq n}$ and $\WeightSize{\Phi}, \WeightSize{\Psi} \leq \CoeffFunc(n)$,
  as well as $L(\Phi), L(\Psi) \leq \min \{ n, \LayerFunc(n) \}$;
  here we used the first part of the lemma.
  By possibly swapping $\Phi,\Psi$ and $\vphantom{\sum_i} f,g$,
  we can assume that $k := L(\Phi) \leq L(\Psi) =: \ell$.
  If $k = \ell$, define $\widetilde{\Phi} := \Phi$.
  If otherwise $k < \ell$, write $\Phi = \big( (A_1,b_1),\dots,(A_k,b_k) \big)$ where
  $A_k \in \R^{1 \times N_{k-1}}$ and $b_k \in \R^1$, and define
  \(
    \Gamma
    := \big(
         \left(
           \begin{smallmatrix}
             1 \\ 1
           \end{smallmatrix}
         \right),
         \left(
           \begin{smallmatrix}
             0 \\ 0
           \end{smallmatrix}
         \right)
       \big)
  \)
  and
  \(
    \Lambda
    := \big(
         (1, -1),
         0
       \big)
  \)
  and finally
  \[
    \widetilde{\Phi}
    := \!
    \bigg(\!
      (A_1,b_1),
      \dots,
      (A_{k-1}, b_{k-1}),
      \Big(\!\!
        \left(
        \begin{smallmatrix}
          A_k \\ - A_k
        \end{smallmatrix}
        \right) \!,
        \left(
        \begin{smallmatrix}
          \smash{b_k} \vphantom{A_k} \\ - \smash{b_k} \vphantom{A_k}
        \end{smallmatrix}
        \right)
      \!\!\Big),
      \Gamma,
      \dots,
      \Gamma,
      \Lambda
    \!\bigg)
    ,
  \]
  where $\Gamma$ appears $\ell - k - 1$ times, so that $L(\widetilde{\Phi}) = \ell$.
  Using the identities $x = \varrho(x) - \varrho(-x)$ and $\varrho(\varrho(x)) = \varrho(x)$,
  it is easy to see $R_\varrho \widetilde{\Phi} = R_\varrho \Phi = f$.
  Moreover, $\WeightSize{\widetilde{\Phi}} \leq \max \{ 1, \CoeffFunc(n) \} = \CoeffFunc(n)$
  and $W(\widetilde{\Phi}) \leq 2 W(\Phi) + 2 (\ell - k) \leq 4 n$.

  Finally, explicitly writing $\widetilde{\Phi} = \big( (B_1, c_1), \dots, (B_\ell, c_\ell) \big)$
  and $\Psi = \big( (C_1, e_1), \dots, (C_\ell, e_\ell) \big)$ with $c_\ell, e_\ell \in \R^{1}$
  and $B_\ell,C_\ell \in \R^{1 \times N_{\ell-1}}$, define
  \[
    \Theta_1
    := \left(
         \left(
           \begin{smallmatrix}
             B_1 \\
             C_1 \\
             0_{4 \times 1}
           \end{smallmatrix}
         \right) ,
         \left(
           \begin{smallmatrix}
             c_1 \\
             e_1 \\
             c_\ell \\
             -c_\ell \\
             e_\ell \\
             -e_\ell
           \end{smallmatrix}
         \right)
       \right)
    \quad \text{and} \quad
    \Theta_m
    := \left(
         \left(
           \begin{smallmatrix}
             B_m & 0   & 0 \\
             0   & C_m & 0 \\
             0   & 0   & I_{4 \times 4}
           \end{smallmatrix}
         \right) ,
         \left(
           \begin{smallmatrix}
             c_m \vphantom{B_m} \\
             e_m \vphantom{C_m} \\
             0_{4 \times 1}
           \end{smallmatrix}
         \right)
       \right)
    \quad \text{for } m \in \{ 2,\dots,\ell-1 \} ,
  \]
  and set
  \[
    \Xi
    := \Big( 
         \Theta_1,
         \dots,
         \Theta_{\ell-1},
         \big(
           (B_\ell \mid C_\ell \mid 1 \mid -1 \mid 1 \mid -1),
           0
         \big)
       \Big) .
  \]
  Using the identities $\varrho(\varrho(x)) = \varrho(x)$ and $x = \varrho(x) - \varrho(-x)$,
  it is then straightforward to verify
  $R_\varrho \Xi = R_\varrho \widetilde{\Phi} + R_\varrho \Psi = f + g$.
  Moreover, $\WeightSize{\Xi} \leq \CoeffFunc(n) \leq \CoeffFunc(9n)$,
  $L(\Xi) = \ell \leq \LayerFunc(n) \leq \LayerFunc(9n)$,
  and $W(\Xi) \leq W(\widetilde{\Phi}) + W(\Psi) + 4 \, \ell \leq 9 n$.
  Here, we used that $\LayerFunc$ and $\CoeffFunc$ are non-decreasing and that $\ell \leq n$.
  Overall, we have shown $f + g \in \NNSigma[9n]$, as claimed.
\end{proof}

With \Cref{lem:NetworkSetsClosedUnderSummation} at our disposal,
we can now prove \Cref{lem:ApproximationSpaceProperties}.

\begin{proof}[Proof of \Cref{lem:ApproximationSpaceProperties}]
  \textbf{Step~1} \emph{(Showing $\PreNorm(f+g) \leq C \cdot (\PreNorm (f) + \PreNorm(g))$):}
  To see this, let $n \in \N_{\geq 9}$ and write $n = 9 m + k$ with $m \in \N$
  and $k \in \{ 0, \dots, 8 \}$, noting that $n \leq 17 m$.
  By \Cref{lem:NetworkSetsClosedUnderSummation},
  we have $\NNSigma[n] \supset \NNSigma[9m] \supset \NNSigma[m] + \NNSigma[m]$ and hence
  \begin{align*}
    n^\alpha \, d_p (f+g, \NNSigma[n])
    & \leq n^\alpha \, d_{p} (f + g, \NNSigma[m] + \NNSigma[m]) \\
    & \leq 17^\alpha m^\alpha \cdot \bigl(d_p (f, \NNSigma[m]) + d_p(g, \NNSigma[m])\bigr) \\
    & \leq 17^\alpha \cdot \big( \PreNorm(f) + \PreNorm(g) \big) .
  \end{align*}
  Moreover, if $n \leq 8$, then we see because of $0 \in \NNSigma[n]$ that
  \[
    n^\alpha \, d_p(f+g, \NNSigma[n])
    \leq 8^\alpha \| f+g \|_{L^p}
    \leq 8^\alpha \cdot (\| f \|_{L^p} + \| g \|_{L^p})
    \leq 8^\alpha \cdot (\PreNorm(f) + \PreNorm(g))
    .
  \]
  Overall, we thus see for every $n \in \N$ that
  $n^\alpha \, d_p (f+g, \NNSigma) \leq C \cdot (\PreNorm(f) + \PreNorm(g))$.
  Since also
  \(
    \| f+g \|_{L^p}
    \leq \| f \|_{L^p} + \| g \|_{L^p}
    \leq \PreNorm(f) + \PreNorm(g)
    \leq C \cdot (\PreNorm(f) + \PreNorm(g)) ,
  \)
  we see that $\PreNorm(f+g) \leq C \cdot (\PreNorm(f) + \PreNorm(g))$,
  as claimed in this step.

  \medskip{}

  \noindent
  \textbf{Step~2} \emph{(Showing $\PreNorm(c f) \leq |c| \, \PreNorm(f)$ for $|c| \leq 1$):}
  Since $|c| \leq 1$, it is straightforward to see $c \NNSigma \subset \NNSigma$ and hence
  \(
    n^\alpha \, d_p(c f, \NNSigma)
    \leq n^\alpha \, d_p (c f, c \NNSigma)
    = |c| n^\alpha \, d_p(f, \NNSigma)
    .
  \)
  This implies
  \(
    \PreNorm(c f)
    \leq \max
         \big\{
           \| c f \|_{L^p} ,
           |c| \sup_{n \in \N} \big[ n^\alpha \, d_p(f, \NNSigma) \big]
         \big\}
    = |c| \, \PreNorm(f) .
  \)

  \medskip{}

  \noindent
  \textbf{Step~3}
  \emph{(Showing $\PreNorm(f) < \infty \Longleftrightarrow \| f \|_{\ApproxSpace} < \infty$):}

  ``$\Rightarrow$:'' For $\theta := 1 + \PreNorm(f) \in [1,\infty)$, Step~2 shows
  $\PreNorm(f/\theta) \leq \frac{1}{\theta} \PreNorm(f) \leq 1$,
  and hence $\| f \|_{\ApproxSpace} < \infty$.

  ``$\Leftarrow$:'' Let $\| f \|_{\ApproxSpace} < \infty$.
  Hence, there exists $\theta > 0$ satisfying $\PreNorm(f/\theta) \leq 1 < \infty$.
  Step~1 shows $\PreNorm(2 g) = \PreNorm(g+g) \leq 2 C \PreNorm(g)$.
  Inductively, this implies $\PreNorm(2^m g) \leq (2 C)^m \PreNorm(g)$ for every $m \in \N$.
  Now, choosing $m \in \N$ such that $\theta \leq 2^m$, Step~2 shows
  \[
    \PreNorm(f)
    =    \PreNorm(\tfrac{\theta}{2^m} 2^m \tfrac{f}{\theta})
    \leq \PreNorm(2^m \tfrac{f}{\theta})
    \leq (2 C)^m \PreNorm(\tfrac{f}{\theta})
    < \infty .
  \]

  \medskip{}

  \noindent
  \textbf{Step~4} \emph{(Homogeneity of $\| \cdot \|_{\ApproxSpace}$):}
  It is easy to see $\| 0 \|_{\ApproxSpace} = 0$.
  Moreover, given $c \in \R \setminus \{ 0 \}$, Step~2 shows that $\PreNorm(\pm f) = \PreNorm(f)$.
  Therefore,
  \[
    \| c f \|_{\ApproxSpace}
    = \inf \{ \theta > 0 \colon \PreNorm( c f / \theta) \leq 1 \} \\
    = |c| \cdot
      \inf
      \big\{
        \tfrac{\theta}{|c|}
        \colon
        \theta > 0
        \text{ and }
        \PreNorm\bigl(\tfrac{f}{\theta / |c|}\bigr) \leq 1
      \big\} \\
    = |c| \, \| f \|_{\ApproxSpace} .  
  \]

  \medskip{}

  \noindent
  \textbf{Step~5} \emph{(Definiteness of $\| \cdot \|_{\ApproxSpace}$):}
  If $\| f \|_{\ApproxSpace} = 0$, then for each $n \in \N$ there exists
  $\theta_n \in (0, \frac{1}{n})$ satisfying $\PreNorm(f / \theta_n) \leq 1$.
  By Step~2, this implies
  \[
    \| f \|_{L^p}
    \leq \PreNorm(f)
    = \PreNorm\bigl(\theta_n \tfrac{f}{\theta_n}\bigr)
    \leq \theta_n \PreNorm\bigl(\tfrac{f}{\theta_n}\bigr)
    \leq \theta_n
    \xrightarrow[n\to\infty]{} 0 ,
  \]
  and hence $f = 0$.

  \medskip{}

  \noindent
  \textbf{Step~6}
  \emph{(If $\| f \|_{\ApproxSpace} \in (0,\infty)$, then $\PreNorm(f / \| f \|_{\ApproxSpace}) \leq 1$):}
  By definition of $\| f \|_{\ApproxSpace}$, there exists a sequence
  $(\theta_n)_{n \in \N} \subset (0,\infty)$ satisfying $\theta_n \to \theta := \| f \|_{\ApproxSpace}$
  and $\PreNorm(f / \theta_n) \leq 1$ for all $n \in \N$.
  Since $\vphantom{\sum_j} \frac{f}{\theta_n} \to \frac{f}{\theta}$ and since $d_p (\cdot, \NNSigma[m])$
  is continuous with respect to $\| \cdot \|_{L^p}$, this implies for each $m \in \N$ that
  \[
    \max
    \big\{
      \big\| \tfrac{f}{\theta} \big\|_{L^p} , \,\,
      m^\alpha \, d_p \! \bigl(\tfrac{f}{\theta}, \NNSigma[m]\bigr)
    \big\}
    = \lim_{n \to \infty}
      \max
      \big\{
        \big\| \tfrac{f}{\theta_n} \big\|_{L^p} , \,\,
        m^\alpha \, d_p \! \bigl(\tfrac{f}{\theta_n}, \NNSigma[m]\bigr)
      \big\}
    \leq 1 ,
  \]
  and hence $\PreNorm(f / \theta) \leq 1$.

  \medskip{}

  \noindent
  \textbf{Step~7}
  \emph{(Showing $\| f+g \|_{\ApproxSpace} \leq C \cdot (\| f \|_{\ApproxSpace} + \| g \|_{\ApproxSpace})$):}
  The claim is trivial if $\| f \|_{\ApproxSpace} \in \{ 0,\infty \}$
  or $\| g \|_{\ApproxSpace} \in \{ 0,\infty \}$.
  Hence, we can assume that $A := \| f \|_{\ApproxSpace} \in (0,\infty)$
  and $B := \| g \|_{\ApproxSpace} \in (0,\infty)$.
  By Steps 1, 2, and 6, this implies
  \begin{align*}
    \PreNorm\bigl(\tfrac{f + g}{C (A + B)}\bigr)
    & \leq \tfrac{1}{C}
           \PreNorm\bigl(\tfrac{f}{A + B} + \tfrac{g}{A + B}\bigr) \\
    & \leq \PreNorm\bigl(\tfrac{A}{A+B} \tfrac{f}{A}\bigr)
           + \PreNorm\bigl(\tfrac{B}{A+B} \tfrac{g}{B}\bigr) \\
    & \leq \tfrac{A}{A+B} \PreNorm\bigl(\tfrac{f}{A}\bigr)
           + \tfrac{B}{A+B} \PreNorm\bigl(\tfrac{g}{B}\bigr)
      \leq 1,  
  \end{align*}
  and hence
  \(
    \| f+g \|_{\ApproxSpace}
    \leq C \cdot (A + B)
    = C \cdot (\| f \|_{\ApproxSpace} + \| g \|_{\ApproxSpace}),
  \)
  as claimed.

  \medskip{}

  \noindent
  \textbf{Step~8}
  \emph{(Showing $\PreNorm(f) \leq 1 \Longleftrightarrow \| f \|_{\ApproxSpace} \leq 1$):}
  ``$\Rightarrow$'' follows by definition of $\| \cdot \|_{\ApproxSpace}$.

  ``$\Leftarrow$'' is trivial if $f = 0$.
  Otherwise, Steps~6 and 2 show for $\theta := \| f \|_{\ApproxSpace} \in (0,1]$ that
  \({
    \PreNorm(f)
    = \PreNorm(\theta \frac{f}{\theta})
    \leq \PreNorm(f / \theta)
    \leq 1 .
  }\)

  \medskip{}

  \noindent
  \textbf{Step~9:} In this step, we prove the last part of \Cref{lem:ApproximationSpaceProperties}.
  First, note that if $\| f \|_{\ApproxSpace(\Omega)} \leq 1$,
  then $\| f \|_{L^p} \leq \PreNorm(f) \leq 1$ thanks to Step~8.
  This proves $\ApproxSpace(\Omega) \hookrightarrow L^p(\Omega)$.

  Next, if $\Omega \subset \overline{\Omega^\circ}$, then it is easy to see for $f \in C_b(\Omega)$
  that $\| f \|_{\sup,\Omega} := \sup_{x \in \Omega} |f(x)| = \| f \|_{L^\infty(\Omega)}$,
  and this implies that $C_b(\Omega) \subset L^\infty(\Omega)$ is closed.
  Therefore, it suffices to show $\ApproxSpace[\infty](\Omega) \subset \overline{C_b(\Omega)}$.
  To see this, let $f \in \ApproxSpace[\infty](\Omega)$; by Step~3, this implies
  $\theta := \PreNorm[\infty](f) < \infty$.
  Furthermore, $\| f \|_{L^\infty} < \infty$.
  By definition of $\PreNorm[\infty]$, for each $n \in \N$ there exists
  $F_n \in \NNSigma$ satisfying $\| F_n - f \|_{L^\infty} \leq 2 C n^{-\alpha} \to 0$
  as $n \to \infty$; in particular, $\| F_n \|_{\sup,\Omega} = \| F_n \|_{L^\infty} < \infty$.
  Finally, since $F_n$ can be extended to a continuous function on all of $\R^d$,
  we see $F_n \in C_b(\Omega)$ and hence $f \in \overline{C_b(\Omega)} = C_b(\Omega)$.
\end{proof}

\subsection{A technical result used in Section~\ref{sec:HatFunctionConstruction}}%
\label{sub:HatConstructionProofs}

\begin{lemma}\label{lem:CubeIntersectionMeasure}
  For each $d \in \N$, $T \in (0,1]$, and $x \in [0,1]^d$, we have
  \[
    \LebesgueMeasure \big( [0,1]^d \cap (x + [-T,T]^d) \big)
    \geq 2^{-d} \, T^d .
  \]
\end{lemma}

\begin{proof}
  For brevity, set $Q := [0,1]^d$.
  Below, we show 
  \begin{equation}
    \LebesgueMeasure \big( Q \cap (x + [-T,T]^d) \big)
    \geq T^d
    \qquad \forall x \in Q \text{ and } T \in (0,\tfrac{1}{2}]
    ,
    \label{eq:MeasureLowerBound}
  \end{equation}
  which clearly implies the claim for these $T$.
  Furthermore, for $T \in [\frac{1}{2},1]$, the above estimate shows
  \(
    \LebesgueMeasure \big( Q \cap (x + [-T,T]^d) \big)
    \geq \LebesgueMeasure \big( Q \cap (x + [-\frac{1}{2}, \frac{1}{2}]^d ) \big)
    \geq 2^{-d}
    \geq 2^{-d} T^d
    ,
  \)
  which proves the claim for general $T \in (0,1]$.

  Thus, let $x \in Q$ and $T \in (0,\frac{1}{2}]$.
  For each $j \in \FirstN{d}$, define $\eps_j := -1$ if $x_j \geq \frac{1}{2}$
  and $\eps_j := 1$ otherwise.
  Let $P := \prod_{j=1}^d \big( \eps_j \, [0,T] \big) \subset [-T,T]^d$.
  We claim that $x + P \subset Q$.
  Once this is shown, it follows that
  \(
    \LebesgueMeasure \bigl( Q \cap (x + [-T,T]^d) \bigr)
    \geq \LebesgueMeasure (x + P)
    =    T^d ,
  \)
  proving \Cref{eq:MeasureLowerBound}.

  To see that indeed $x + P \subset Q$, let $y \in P$ be arbitrary.
  For each $j \in \FirstN{d}$, there are then two cases:
  \begin{enumerate}
    \item If $x_j \geq \frac{1}{2}$, then $\eps_j = -1$
          and $-\frac{1}{2} \leq -T \leq y_j \leq 0$.
          Thus, $0 \leq x_j - \frac{1}{2} \leq x_j + y_j \leq x_j \leq 1$,
          meaning $(x + y)_j \in [0,1]$.

    \item If $x_j < \frac{1}{2}$, then $\eps_j = 1$ and $0 \leq y_j \leq T \leq \frac{1}{2}$.
          Thus, $0 \leq x_j \leq x_j + y_j \leq \frac{1}{2} + \frac{1}{2} = 1$,
          so that we see again $(x + y)_j \in [0,1]$.
  \end{enumerate}
  Overall, this shows in both cases that $x + y \in [0,1]^d = Q$.
\end{proof}

\subsection{A technical result regarding measurability}%
\label{sub:IntegrationErrorTechnicalProofs}

\begin{lemma}\label{lem:MeasurableERM}
  Let $\emptyset \neq \Omega \subset \R^d$ be compact
  and let $\emptyset \neq \CalH \subset C(\Omega)$ be compact.
  Then, equipping $\CalH$ with the Borel $\sigma$-algebra induced from $C(\Omega)$,
  the following hold:
  \begin{enumerate}
    \item The map
          \[
            M : \quad
            \Omega^m \times \CalH \to \Omega^m \times \R^m, \quad
            (\x, f) = \bigl( (x_1,\dots,x_m), f\bigr)
            \mapsto \Big(\x, \bigl( f(x_1),\dots,f(x_m) \bigr) \Big)
          \]
          is continuous and hence measurable;

    \item there is a measurable map $B : \Omega^m \times \R^m \to \CalH$ satisfying
          \[
            B(\x, \y)
            \in \argmin_{g \in \CalH}
                  \sum_{i=1}^m
                    \big( g(x_i) - y_i \big)^2
            \quad \forall \, \x = (x_1,\dots,x_m) \in \Omega^m
                             \text{ and } \y = (y_1,\dots,y_m) \in \R^m.
          \]
  \end{enumerate}
\end{lemma}

\begin{proof}
\textbf{Part~1:}
It is enough to prove continuity of each of the components of $M$.
For the component $(\x,f) \mapsto \x$ this is trivial.
For the component $(\x,f) \mapsto f(x_j)$ note that if $\Omega \ni \x^{(n)} \to \x \in \Omega$
and $\CalH \ni f_n \to f \in \CalH$ (with convergence in $C(\Omega)$), then
\begin{equation}
  \begin{split}
    \bigl|f(\x_j) - f_n\bigl(\x^{(n)}_j\bigr)\bigr|
    & \leq \bigl|f(\x_j) - f\bigl(\x^{(n)}_j\bigr)\bigr|
           + \bigl|f\bigl(\x^{(n)}_j\bigr) - f_n\bigl(\x^{(n)}_j\bigr)\bigr| \\
    & \leq \bigl|f(\x_j) - f\bigl(\x^{(n)}_j\bigr)\bigr| + \| f - f_n \|_{C(\Omega)}
      \xrightarrow[n\to\infty]{} 0 ,
  \end{split}
  \label{eq:PointEvaluationJointlyContinuous}
\end{equation}
since $f$ is continuous.
Thus, $M$ is continuous.
To see that this implies that $M$ is measurable, note that both $\Omega$ and $\CalH$
are separable metric spaces (and hence second countable), so that
the product $\sigma$-algebra on $\Omega \times \CalH$ coincides with the Borel $\sigma$-algebra
on $\Omega \times \CalH$; see for instance \cite[Theorem~7.20]{FollandRA}.

\medskip{}

\noindent
\textbf{Part~2:}
For this part, we use the ``Measurable Maximum Theorem,''
\cite[Theorem~18.19]{AliprantisBorderHitchhiker}.
Thanks to this theorem, setting $S := \Omega \times \R^m$, it is enough to show that
\begin{enumerate}
  \item the set-valued map%
        \footnote{A set-valued map $f : X \twoheadrightarrow Y$ is a map $f : X \to 2^Y$.}
        $\varphi : S \twoheadrightarrow C(\Omega), (\x,f) \mapsto \CalH$
        is weakly measurable with nonempty, compact values;

  \item the map
        \(
          F :
          S \times C(\Omega) \to \R,
          \bigl( (\x,\y), g \bigr) \mapsto -\sum_{i=1}^m \bigl(g(x_i) - y_i\bigr)^2
        \)
        is a Carathéodory function (see \cite[Definition~4.50]{AliprantisBorderHitchhiker}).
\end{enumerate}
By our assumptions on $\CalH$, it is clear that $\varphi$ has nonempty, compact values.
The weak measurability of $\varphi$ follows directly from the definition,
see \cite[Definition~18.1]{AliprantisBorderHitchhiker}.
For the second property, it is enough to show that $F$ is continuous.
This follows as in \Cref{eq:PointEvaluationJointlyContinuous}.
\end{proof}

\subsection{A technical result regarding random subsets of \texorpdfstring{$\{ 1,\dots,m \}$}{\{1,...,m\}}}%
\label{sub:RandomSubsetResult}

\begin{lemma}\label{lem:RandomSubsetExpectationBound}
  Let $m \in \N$ and $1 \leq k \leq 2 m$.
  Write $\CalP_k (\FirstN{2m}) := \{ J \subset \FirstN{2m} \colon |J| = k \}$.
  Then, for each subset $I \subset \FirstN{2m}$ with $|I| \geq m$, we have
  \[
    \avsum_{j \in \CalP_k (\FirstN{2m})}
      |J \cap I|^{1/2}
    \geq \frac{1}{4} \cdot k^\beta .
  \]
\end{lemma}

\begin{proof}
  Let $I^c := \FirstN{2m} \setminus I$.
  We note for any $T \subset \FirstN{2m}$ that the quantity
  \(
    \psi(T)
    := \avsum_{J \in \SpecialSet}
         |J \cap T|^{1/2}
  \)
  only depends on the cardinality $|T|$ and that $\psi(T) \leq \psi(S)$ if $|T| \leq |S|$.
  Since $|I| \geq m \geq |I^c|$, this implies $\psi(I) \geq \psi(I^c)$.
  Combined with the estimate
  \[
    |J \cap I|^{\frac{1}{2}} + |J \cap I^c|^{\frac{1}{2}}
    \geq \Big[ \max \big\{ |J \cap I| , |J \cap I^c| \big\} \Big]^{\frac{1}{2}}
    \geq \Big[ \tfrac{1}{2} \bigl(|J \cap I| + |J \cap I^c|\bigr) \Big]^{\frac{1}{2}}
    \geq \big( \tfrac{1}{2} |J| \big)^{\frac{1}{2}}
    \geq \tfrac{1}{2} |J|^{\frac{1}{2}}
    =    \tfrac{1}{2} k^{\frac{1}{2}}
  \]
  which holds for all $J \in \SpecialSet$, we finally see
  \[
    \avsum_{J \in \SpecialSet}
      |J \cap I|^{1/2}
    = \psi(I)
    \geq \frac{\psi(I) + \psi(I^c)}{2}
    =    \frac{1}{2}
         \avsum_{J \in \SpecialSet}
           \big(
             |J \cap I|^{\frac{1}{2}} + |J \cap I^{c}|^{\frac{1}{2}}
           \big)
    \geq \frac{1}{4} k^{\frac{1}{2}} .
    \qedhere
  \]
\end{proof}

\subsection{Two technical optimization results}%
\label{sub:OptimizationResults}

\begin{lemma}\label{lem:OptimizationLemma1}
  Let $\gamma^\flat \in [1,\infty]$ and $\alpha > 0$.
  Let
  \begin{equation}
    \Psi
    := \big\{
         (\gamma,\theta,\lambda) \in (0,\infty) \times (0,\infty) \times [0,1]
         \colon
         \gamma < \gamma^\flat \text{ and } \theta \lambda \leq 1
       \big\} .
    \label{eq:PsiDefinition}
  \end{equation}
  Then
  \[
    \inf_{(\gamma,\theta,\lambda) \in \Psi}
      \max
      \big\{
        \theta \cdot (\alpha - \tfrac{\lambda}{2}), \,\,\,
        1 + \theta \cdot (\tfrac{\lambda}{2} - \gamma)
      \big\}
    \leq \begin{cases}
           \min
           \big\{
             \frac{\alpha}{\alpha + \gamma^\flat}, \,\,
             \frac{2 \alpha}{\alpha + \gamma^\flat} - \frac{1}{2}
           \big\} ,
           & \text{if } \alpha + \gamma^\flat <    2, \\
           \min
           \big\{
             \frac{\alpha}{\alpha + \gamma^\flat}, \,\,
             \alpha - \frac{1}{2}, \,\,
             \frac{\alpha - \frac{1}{2}}{\alpha + \gamma^\flat - 1}
           \big\} ,
           & \text{if } \alpha + \gamma^\flat \geq 2.
         \end{cases}
  \]
\end{lemma}

\begin{rem*}
  In fact, one has equality.
  But since we do not need this, we omit the proof (and the explicit statement) of this fact.
\end{rem*}

\begin{proof}
  \textbf{Step~1 (Preparations):}
  Define $f_1(\gamma,\theta,\lambda) := \theta \cdot (\alpha - \frac{\lambda}{2})$
  and $f_2 (\gamma,\theta,\lambda) := 1 + \theta \cdot (\frac{\lambda}{2} - \gamma)$
  as well as $f := \max \{ f_1, f_2 \}$ and
  $\beta_\ast := \inf_{(\gamma,\theta,\lambda) \in \Psi} f(\gamma,\theta,\lambda)$.
  For arbitrary $0 < \gamma < \gamma^\flat$, we have $(\gamma,\frac{1}{\alpha+\gamma},0) \in \Psi$
  and hence
  \(
    \beta_\ast
    \leq f(\gamma,\frac{1}{\alpha+\gamma},0)
    = \max
      \big\{
        \frac{\alpha}{\alpha+\gamma}, 1 - \frac{\gamma}{\alpha + \gamma}
      \big\}
    = \frac{\alpha}{\alpha + \gamma}
    .
  \)
  Letting $\gamma \uparrow \gamma^\flat$, this implies
  \begin{equation}
    \beta_\ast
    \leq \frac{\alpha}{\alpha + \gamma^\flat}
    .
    \label{eq:OptimizationLemma1Step1}
  \end{equation}

  \medskip{}

  \noindent
  \textbf{Step~2 (The case $\gamma^\flat = \infty$):}
  Let us first consider the case $\gamma^\flat = \infty$.
  In this case, \Cref{eq:OptimizationLemma1Step1} shows $\beta_\ast \leq 0$.
  Furthermore, given $0 < \gamma < \gamma^\flat = \infty$, we have $(\gamma,1,1) \in \Psi$,
  which shows that
  $\beta_\ast \leq f(\gamma,1,1) = \max \bigl\{ \alpha - \frac{1}{2}, \frac{3}{2} - \gamma \bigr\}$.
  Letting $\gamma \to \infty$, we thus see $\beta_\ast \leq \alpha - \frac{1}{2}$
  and hence $\beta_\ast \leq \min \{ 0, \alpha - \frac{1}{2} \}$.
  It is easy to see that this implies the claim for $\gamma^\flat = \infty$.

  Hence, we can assume from now on that $\gamma^\flat$ is finite.
  Then, we easily see for $g_1(\theta,\lambda) := \theta \cdot (\alpha - \frac{\lambda}{2})$
  and $g_2(\theta,\lambda) := 1 + \theta \cdot (\frac{\lambda}{2} - \gamma^\flat)$
  as well as $g := \max \{ g_1, g_2 \}$ and
  $\Omega := \{ (\theta, \lambda) \in (0,\infty) \times [0,1] \colon \theta \lambda \leq 1  \}$
  that $\beta_\ast \leq \inf_{(\theta,\lambda) \in \Omega} g(\theta,\lambda)$.

  \medskip{}

  \noindent
  \textbf{Step~3 (The case $\alpha + \gamma^\flat < 2$):}
  In this case, we have $\frac{2}{\alpha + \gamma^\flat} \in (1,\infty)$
  and hence $(\frac{2}{\alpha+\gamma^\flat}, \frac{\alpha + \gamma^\flat}{2}) \in \Omega$.
  Furthermore,
  \(
    g_1(\frac{2}{\alpha+\gamma^\flat}, \frac{\alpha + \gamma^\flat}{2})
    = g_2(\frac{2}{\alpha+\gamma^\flat}, \frac{\alpha + \gamma^\flat}{2})
    = \frac{2 \alpha}{\alpha + \gamma^\flat} - \frac{1}{2}
  \)
  and hence $\beta_\ast \leq \frac{2 \alpha}{\alpha + \gamma^\flat} - \frac{1}{2}$.
  Together with \Cref{eq:OptimizationLemma1Step1},
  this proves the claim for $\alpha + \gamma^\flat < 2$.

  \medskip{}

  \noindent
  \textbf{Step~4 (The case $\alpha + \gamma^\flat \geq 2$):}
  Note $g_1(1,1) = \alpha - \frac{1}{2}$ and
  ${g_2(1,1) = \frac{3}{2} - \gamma^\flat \leq \frac{3}{2} - (2 - \alpha) = \alpha - \frac{1}{2}}$.
  Since $(1,1) \in \Omega$, this implies $\beta_\ast \leq g(1,1) = \alpha - \frac{1}{2}$.
  Furthermore, $\theta_0 := \frac{1}{\alpha + \gamma^\flat - 1} \in (0,1]$
  and hence $(\theta_0,1) \in \Omega$.
  It is easy to see
  $g_1(\theta_0,1) = g_2(\theta_0,1) = \frac{\alpha - \frac{1}{2}}{\alpha + \gamma^\flat - 1}$
  and hence $\beta_\ast \leq g(\theta_0,1) = \frac{\alpha - \frac{1}{2}}{\alpha + \gamma^\flat - 1}$.
  Combining these two estimates with \Cref{eq:OptimizationLemma1Step1} completes the proof
  for the case $\alpha + \gamma^\flat \geq 2$.
\end{proof}

\begin{lemma}\label{lem:OptimizationLemma2}
  Let $\gamma^\flat \in [1,\infty]$ and $\alpha > 0$.
  Let $\Psi$ be as in \Cref{eq:PsiDefinition}.
  Then
  \begin{equation}
    \inf_{(\gamma,\theta,\lambda) \in \Psi}
      \max \big\{ \theta \cdot (\alpha - \lambda), \quad 1 - \theta \gamma \big\}
    \leq \begin{cases}
           \frac{2 \alpha}{\alpha + \gamma^\flat} - 1,
           & \text{if } \alpha + \gamma^\flat \leq 2, \\
           \min \big\{ \alpha - 1, \frac{\alpha - 1}{\alpha + \gamma^\flat - 1} \big\}
           , & \text{if } \alpha + \gamma^\flat >    2.
         \end{cases}
    \label{eq:OptimizationLemma2}
  \end{equation}
\end{lemma}

\begin{proof}
  For brevity, denote the left-hand side of \Cref{eq:OptimizationLemma2} by $\beta_\ast$.

  We first consider the special case $\gamma^{\flat} = \infty$.
  Define $g := \max \{ g_1, g_2 \}$, where
  $g_1(\gamma,\theta,\lambda) := \theta \cdot (\alpha - \lambda)$
  and $g_2(\gamma,\theta,\lambda) := 1 - \theta \gamma$.
  For any $\gamma > 0$, we have $g_1(\gamma,1,1) = \alpha - 1$ and $g_2(\gamma,1,1) = 1 - \gamma$
  and furthermore $(\gamma,1,1) \in \Psi$.
  Therefore,
  \(
    \beta_\ast
    \leq g(\gamma,1,1)
    = \max
      \{
        \alpha - 1, \,\,
        1 - \gamma
      \}
    \xrightarrow[\gamma\to\infty]{} \alpha - 1.
  \)
  Furthermore, for arbitrary $\gamma > 0$ we have $(\gamma,\frac{1}{\gamma},0) \in \Psi$
  and $g_1(\gamma,\frac{1}{\gamma},0) = \frac{\alpha}{\gamma}$
  and $g_2(\gamma,\frac{1}{\gamma},0) = 0$, so that
  $\beta_\ast \leq \min \{ 0, \frac{\alpha}{\gamma} \} \xrightarrow[\gamma\to\infty]{} 0$.
  Overall, we have thus shown $\beta_\ast \leq \min \{ \alpha - 1, 0 \}$, which easily
  implies that \Cref{eq:OptimizationLemma2} holds in case of $\gamma^\flat = \infty$.

  \medskip{}

  Hence, we can assume that $\gamma^\flat < \infty$.
  Then, setting
  $\Omega := \{ (\theta,\lambda) \in (0,\infty) \times [0,1] \colon \theta \lambda \leq 1 \}$
  and furthermore $f := \max \{ f_1, f_2 \}$ for $f_1(\theta,\lambda) := \theta (\alpha - \lambda)$
  and $f_2(\theta,\lambda) := 1 - \theta \gamma^\flat$, it is easy to see
  by continuity that $\beta_\ast \leq \inf_{(\theta,\lambda) \in \Omega} f(\theta,\lambda)$.
  We now distinguish two cases:

  \smallskip{}

  \textbf{Case~1 ($\alpha + \gamma^\flat \leq 2$):}
  In this case, $\theta_0 := \frac{2}{\alpha + \gamma^\flat} \in [1,\infty)$ and
  $\lambda_0 := \frac{1}{\theta} \in (0,1]$ satisfy $(\theta,\lambda) \in \Omega$.
  Furthermore, it is easy to see
  $f_1(\theta_0,\lambda_0) = \frac{2 \alpha}{\alpha + \gamma^\flat} - 1 = f_2(\theta_0,\lambda_0)$.
  Thus, $\beta_\ast \leq f(\theta_0,\lambda_0) = \frac{2 \alpha}{\alpha + \gamma^\flat} - 1$,
  which proves \Cref{eq:OptimizationLemma2} in this case.

  \medskip{}

  \textbf{Case~2 ($\alpha + \gamma^\flat >    2$):}
  First note because of $\alpha + \gamma^\flat > 2$ that
  $f_1(1,1) = \alpha - 1 > 1 - \gamma^\flat = f_2(1,1)$ and hence
  $\beta_\ast \leq f(1,1) = \alpha - 1$.
  Furthermore, we have $\theta^\ast := \frac{1}{\alpha + \gamma^\flat - 1} \in (0,1)$
  and hence $(\theta^\ast, 1) \in \Omega$.
  Furthermore, it is easy to see
  \(
    f_1(\theta^\ast,1)
    = \frac{\alpha - 1}{\alpha + \gamma^\flat - 1}
    f_2(\theta^\ast,1) ,
  \)
  which implies $\beta_\ast \leq f(\theta^\ast,1) = \frac{\alpha - 1}{\alpha + \gamma^\flat - 1}$.
  Overall, we see
  $\beta_\ast \leq \min \big\{ \alpha - 1, \frac{\alpha - 1}{\alpha + \gamma^\flat - 1} \big\}$,
  which shows that \Cref{eq:OptimizationLemma2} holds for $\alpha + \gamma^\flat > 2$.
\end{proof}

\bibliographystyle{myplain}
\footnotesize
\bibliography{references,refs}

\end{document}